\documentclass[journal]{IEEEtran}
\usepackage{cite}
\usepackage{amsmath,amssymb,amsfonts}
\usepackage{caption}
\usepackage{floatrow}   
\usepackage{subcaption}
\usepackage{algorithmic}
\usepackage{graphicx}
\usepackage{textcomp}
\usepackage{xcolor}
\usepackage{svg}
\usepackage{multicol}
\usepackage{multirow}
\ifCLASSINFOpdf
\usepackage{ragged2e}
\usepackage{booktabs}
\usepackage{fancyhdr}
\usepackage[font=footnotesize]{caption}

\begin{document}

\title{An Empirical Review of Deep Learning Frameworks for Change Detection: Model Design, Experimental Frameworks, Challenges and Research Needs}

\author{Murari~Mandal, Santosh~Kumar~Vipparthi
\thanks{This work is in part supported by the DST-SERB project under Grant EEQ/2017/000673. The authors are with the Vision Intelligence Lab, Department of Computer Science and Engineering, Malaviya National Institute of Technology Jaipur, 302017, India (E-mail: murarimandal.cv@gmail.com; skvipparthi@mnit.ac.in)}
}
\markboth{IEEE Transactions on Intelligent Transportation Systems, Accepted}{}

\maketitle
\begin{abstract}
Visual change detection, aiming at segmentation of video frames into foreground and background regions, is one of the elementary tasks in computer vision and video analytics. The applications of change detection include anomaly detection, object tracking, traffic monitoring, human machine interaction, behavior analysis, action recognition, and visual surveillance. Some of the challenges in change detection include background fluctuations, illumination variation, weather changes, intermittent object motion, shadow, fast/slow object motion, camera motion, heterogeneous object shapes and real-time processing. Traditionally, this problem has been solved using hand-crafted features and background modelling techniques. In recent years, deep learning frameworks have been successfully adopted for robust change detection. This article aims to provide an empirical review of the state-of-the-art deep learning methods for change detection. More specifically, we present a detailed analysis of the technical characteristics of different model designs and experimental frameworks. We provide model design based categorization of the existing approaches, including the 2D-CNN, 3D-CNN, ConvLSTM, multi-scale features, residual connections, autoencoders and GAN based methods. Moreover, an empirical analysis of the evaluation settings adopted by the existing deep learning methods is presented. To the best of our knowledge, this is a first attempt to comparatively analyze the different evaluation frameworks used in the existing deep change detection methods. Finally, we point out the research needs, future directions and draw our own conclusions.

\end{abstract}

\IEEEpeerreviewmaketitle

\section{Introduction}
\IEEEPARstart{C}hange detection (CD) in video streams is an essential task in computer vision with numerous applications in video synopsis generation~\cite{li2017general,nie2019collision}, anomaly detection~\cite{marotirao2019challenges}, object tracking~\cite{dong2019quadruplet,xu2019learning}, traffic monitoring~\cite{zhang2020rapnet}, human machine interaction~\cite{motiian2016online}, behavior analysis~\cite{li2020quantifying}, action recognition~\cite{zhang2018real}, and visual surveillance~\cite{yi2016pedestrian,fu2019foreground}. The aim of a CD algorithm is to segment a video frame into foreground and background regions. Such pre-processed video frames are frequently used in higher-level tasks as discussed above. Since the outcome of the CD algorithm has a great impact on the overall performance of subsequent steps in high-level applications. Therefore, it is crucial that the method produce as robust foreground/background segmentation as possible. One of the strengths of CD algorithms 
is that it is completely free from the requirement of manual target or object mask initialization by the user. The onus of background initialization and maintenance for identifying the foreground objects is also put on the CD method\cite{Mandal_2020_WACV}. Thus, the CD algorithms can also assist the visual object tracking methods in assigning the target objects for further processing. However, designing a robust CD method is a very challenging task due to numerous real world challenges discussed earlier.\par

\textbf{Application of CD in intelligent transportation systems:} The tremendous advancement in deep learning has fueled breakthrough in several computer vision applications for intelligent transportation systems (ITS). The video-based analytics are often preferred in developing ITS over other modalities (such a LIDAR) due to its lower cost and ease in accessibility. As a low-level video task, change detection or moving object detection is commonly used in autonomous driving~\cite{rasouli2019autonomous}, anomaly detection~\cite{kim2020anomaly,roy2020detection}, traffic analysis~\cite{tian2019online}, and intelligent surveillance~\cite{wan2020intelligent,ahmed2019query}. However, various real-world scenarios such as fluctuation in background regions, illumination variation, shadow, heterogeneous object shapes, variable frame rate of different cameras, weather changes, intermittent object motion, camera jitter and variable object motion make change detection a very challenging task~\cite{patil2018msfgnet,akilan20193d,mandal20203dcd,akilan20193d}. Furthermore, for real-time applications in various mobile devices, it is imperative that the CD methods function at very high speed with minimal resource requirements~\cite{murari-tits,roy2017real,zhao2016real}. These challenges have been partially addressed (independently or collectively) in the literature. Our detailed review and analysis of the existing deep CD methods is an important contribution for ITS applications.\par

\begin{figure}[t]
\setlength\abovecaptionskip{-0.5\baselineskip}
\centering
	\includegraphics[width=1\textwidth ]{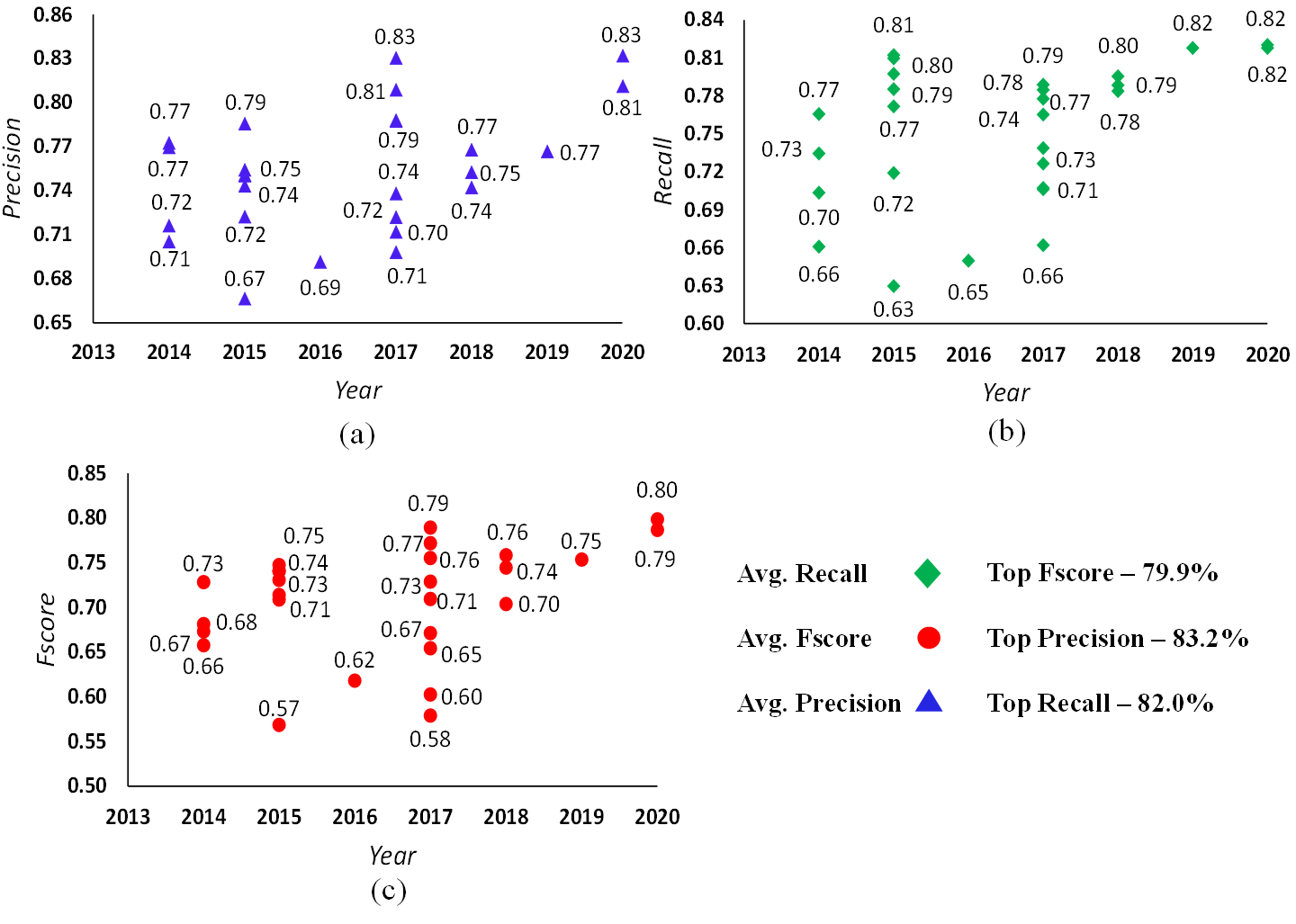}
	\caption{The evolution of change detection performance with traditional approaches in CDnet 2014 dataset. We observe the topmost performances in the range of 79\%-84\%.There has been steady rise in the performance over the years. However, significant improvements have been obtained with the adoption of deep learning as shown in Fig. \ref{cdsurvey_fig2}
	}
	\label{cdsurvey_fig1}
\end{figure}

\begin{figure}[t]
\centering
\begin{subfigure}{0.8\linewidth}
    \centering
    \includegraphics[width=0.9\textwidth]{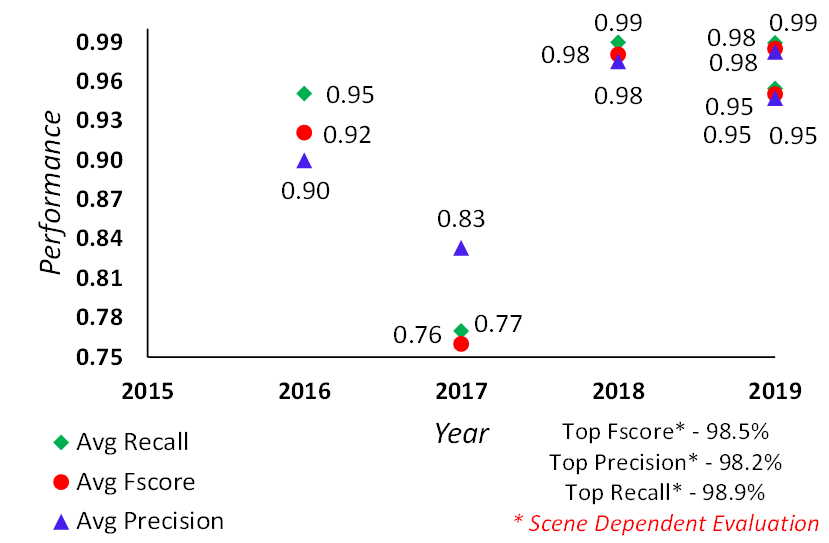}
    \caption{}\label{cdsurvey_fig2a}
\end{subfigure}
    \hfill
\begin{subfigure}{0.9\linewidth}
    \centering
    \includegraphics[width=0.8\textwidth]{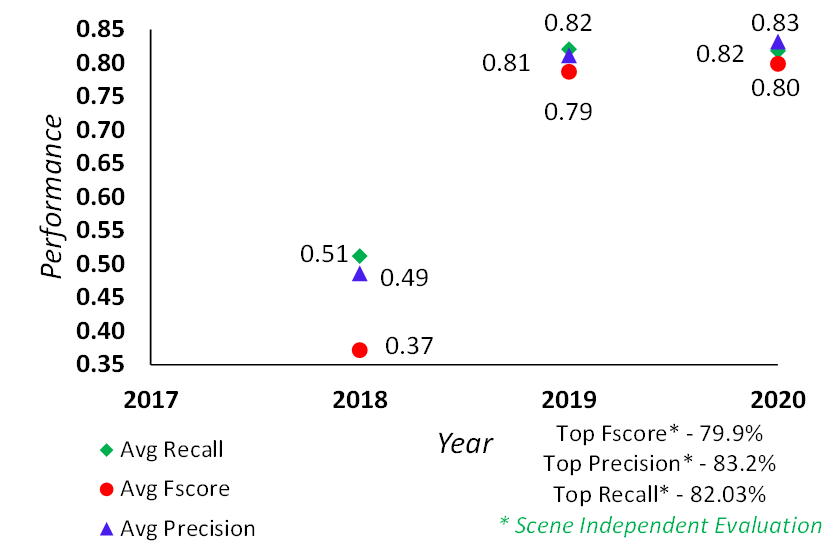}
    \caption{}\label{cdsurvey_fig2b}
\end{subfigure}
	\caption{The recent evolution of change detection performance with deep learning approaches in CDnet 2014 dataset~\cite{wang2014cdnet}. We observe the top performances in the range of 98\%-99\% in scene dependent evaluation (SDE) setup. Whereas, in scene independent evaluation (SIE) setup, the top performances are in the range of 79\%-84\%. (a) Results in SDE setup. (training and testing frames are collected from the same videos). (b) Results in SIE setup. (training and testing frames are collected from completely different videos). The SDE results have significantly outperformed all previous results. However, the well-generalized SIE results are improving gradually with recent focus in this direction among researchers}
	\label{cdsurvey_fig2}
\end{figure}

Overall, the CD methods can be categorized into traditional and learning-based approaches. The traditional methods could be further grouped into parametric~\cite{zivkovic2004improved,stauffer1999adaptive,varadarajan2013spatial}, non-parametric~\cite{jiang2017wesambe,wang2014fast,ramirez2016auto,mandal2018antic,st2014subsense,hofmann2012background} and hybrid/other methods based on the background subtraction techniques used to model background behavior and identify foreground region using various thresholding techniques. The growth in the performance of the traditional approaches over the benchmark CDnet 2014 dataset is shown in Fig. \ref{cdsurvey_fig1}. The learning-based methods are further divided into supervised and unsupervised approaches. The most significant improvement in performance is led by the recent convolutional neural networks (CNN) designed for supervised change detection. In this paper, we primarily focus on the characteristics of these deep learning based methods and their generalization capabilities to real-world unseen videos. In Fig.~\ref{cdsurvey_fig2a} and Fig.~\ref{cdsurvey_fig2b}, we depict the evolution in the performance of the deep learning methods over benchmark CDnet 2014 dataset.\par

\begin{figure}[t]
\centering
	\includegraphics[width=0.9\textwidth ]{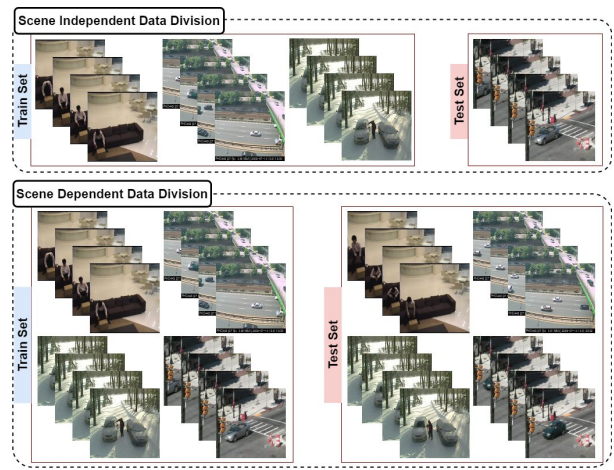}
	\caption{Difference between scene dependent (SDE) and scene independent (SIE) data division schemes. In SDE setup, training and testing frames are collected from the same video. Whereas, in SIE setup, only completely unseen videos are used for evaluation}
	\label{cdsurvey_fig3}
\end{figure}

The deep learning CD methods are further grouped into different types based on the characteristics such as dependence on hand-crafted background models, single/multiple frame-based segmentation, patch-based analysis, dependence on pretrained weights and finetuning. The generative adversarial networks (GANs) and autoencoders are another class of models used for CD. In terms of the experimental setups, the traditional and supervised methods require different considerations. The traditional methods for CD usually do not have prior requirement of labelled data for algorithms development. Thus, there is no need to define train-test splits. However, it is a crucial decision in supervised change detection techniques. The benchmark CDnet 2014~\cite{wang2014cdnet} and other datasets such as PTIS \cite{li2004statistical}, and LASIESTA do not define the train-test division. Thus, researchers have used different data division strategies for network training and evaluation. We further categorize the supervised methods into scene dependent (SDE) and scene independent evaluation (SIE) setup. In ‘scene dependent’ setup, train and test sets consist of frames from the same video sequences. Whereas in ‘scene independent’ setup, completely unseen videos are used for testing. A sample data-division strategy for SDE and SIE is shown in Fig. \ref{cdsurvey_fig3}.\par

\begin{table*}[]
\footnotesize
\centering
\caption{Summarization of a number of related surveys in the last decade}  \resizebox{\columnwidth}{!}{
\begin{tabular}{c c c c}
\hline\hline
\textbf{No.}	&\textbf{Pub-Year}	&\textbf{Title}	&\textbf{Content} \\
\hline\hline
\multirow{2}{*}{1}	&\multirow{2}{*}{CSR-2014~\cite{bouwmans2014traditional}}
 &Traditional and Recent Approaches in Background 	
 &A survey of the traditional approaches \\
 & &Modeling for Foreground Detection: An Overview & for background subtraction \\
 \hline
\multirow{2}{*}{2}	&\multirow{2}{*}{CVIU-2014~\cite{sobral2014comprehensive}}	
&A Comprehensive Review of Background Subtraction 
&A detailed comparison of 29 background subtraction methods  \\
& &Algorithms Evaluated with Synthetic and Real Videos	 &in terms of quantitative performance and computational complexity \\
\hline
\multirow{2}{*}{3}	&\multirow{2}{*}{TITS-2016~\cite{datondji2016survey}}	
&A Survey of Vision-Based Traffic Monitoring 
&A review of studies related to vehicle detection  \\
& &of Road Intersections	 &and tracking in intersection-like scenarios \\
\hline
\multirow{2}{*}{4}	&\multirow{2}{*}{WACV-2016~\cite{sommer2016survey}}	
&A Survey on Moving Object Detection 	
&An overview of the existing methods for moving  \\
& &for Wide Area Motion Imagery &object detection in WAMI data \\
\hline
\multirow{2}{*}{5}	&\multirow{2}{*}{CVIU-2016~\cite{cuevas2016detection}}	
&Detection of Stationary Foreground 	&A survey of the most relevant approaches for detecting  \\
& &Objects: A Survey &all kind of stationary foreground objects \\
\hline
\multirow{2}{*}{6}	&\multirow{2}{*}{TITS-2017~\cite{prasad2017video}}	
&Video Processing From Electro-Optical Sensors for Object	
&An overview of various approaches of video processing \\
& & Detection and Tracking in a Maritime Environment: A Survey &for object detection and tracking in the maritime environment  \\
\hline
\multirow{2}{*}{7}	&\multirow{2}{*}{AIR-2017~\cite{goyal2018review}}	
&Review of Background Subtraction Methods using Gaussian 	&A review of various background  \\
& &Mixture Model for Mideo Surveillance Systems &subtraction algorithms based on GMM \\
\hline
\multirow{2}{*}{8}	&\multirow{2}{*}{PRL-2017~\cite{bouwmans2017scene}}	
&Scene Background 	&A taxonomy study for background initialization  \\
& &Initialization: A Taxonomy &methods for background subtraction \\
\hline
\multirow{2}{*}{9} &\multirow{2}{*}{TITS-2018~\cite{prasad2018object}}	
&Object Detection in a Maritime Environment: Performance 	
&A benchmark of the performance of 23 existing   \\
& &Evaluation of Background Subtraction Methods &background subtraction algorithms for maritime vision\\
\hline
\multirow{2}{*}{10}	&\multirow{2}{*}{JoI-2018~\cite{maddalena2018background}}	
&Background Subtraction for Moving Object &A review on the background subtraction methods  \\
& &Detection in RGBD data: A Survey	 &for moving object detection in RGB-D data \\
\hline
\multirow{2}{*}{11}	&\multirow{2}{*}{CSR-2018~\cite{yazdi2018new}}	
&New Trends on Moving Object Detection in Video Images &A survey on the existing moving object detection methods \\
& &Captured by a Moving Camera: A Survey &in video sequences captured by a moving camera \\
\hline
\multirow{2}{*}{12}	&\multirow{2}{*}{IA-2019~\cite{kalsotra2019comprehensive}}	
&A Comprehensive Survey of Video Datasets &A comprehensive account of the available\\
& &for Background Subtraction	 &public datasets for change detection \\
\hline
\multirow{2}{*}{13}	&\multirow{2}{*}{NN-2020~\cite{bouwmans2019deep}}	
&Deep Neural Network Concepts for Background 	
&A survey of the background initialization and \\
& &Subtraction: A Systematic Review and Comparative Evaluation &subtraction methods based on deep neural networks  \\
\hline
\multirow{2}{*}{14}	&\multirow{2}{*}{Arxiv-2020~\cite{chapel2020moving}}
&Moving Objects Detection with a Moving 
&A categorization of existing methods for moving  \\
& &Camera: A Comprehensive Review &object detection with a moving camera \\
\hline
\multirow{2}{*}{15}	&\multirow{2}{*}{TITS-2020~\cite{espinosa2020detection}}	
&Detection of Motorcycles in Urban Traffic 
&A review of the algorithms used for detection\\
& &Using Video Analysis: A Review & and tracking of motorcycles in CCTV cameras\\
\hline
\multirow{3}{*}{\textbf{16}}	&\multirow{3}{*}{\textbf{Ours}}
&\textbf{An Empirical Review of Deep Learning Frameworks}
&\textbf{A comprehensive empirical review of the recent}  \\
& & \textbf{for Change Detection: Model Design, Experimental} 
&\textbf{deep learning model designs (technical characteristics)}\\
& &\textbf{Frameworks, Challenges and Research Needs} & \textbf{and evaluation frameworks for change detection}\\
\hline\hline
    \end{tabular}
  }
    \label{tab_prev_surveys}
\end{table*}

\begin{figure*}[t]
\centering
	\includegraphics[width=0.7\textwidth ]{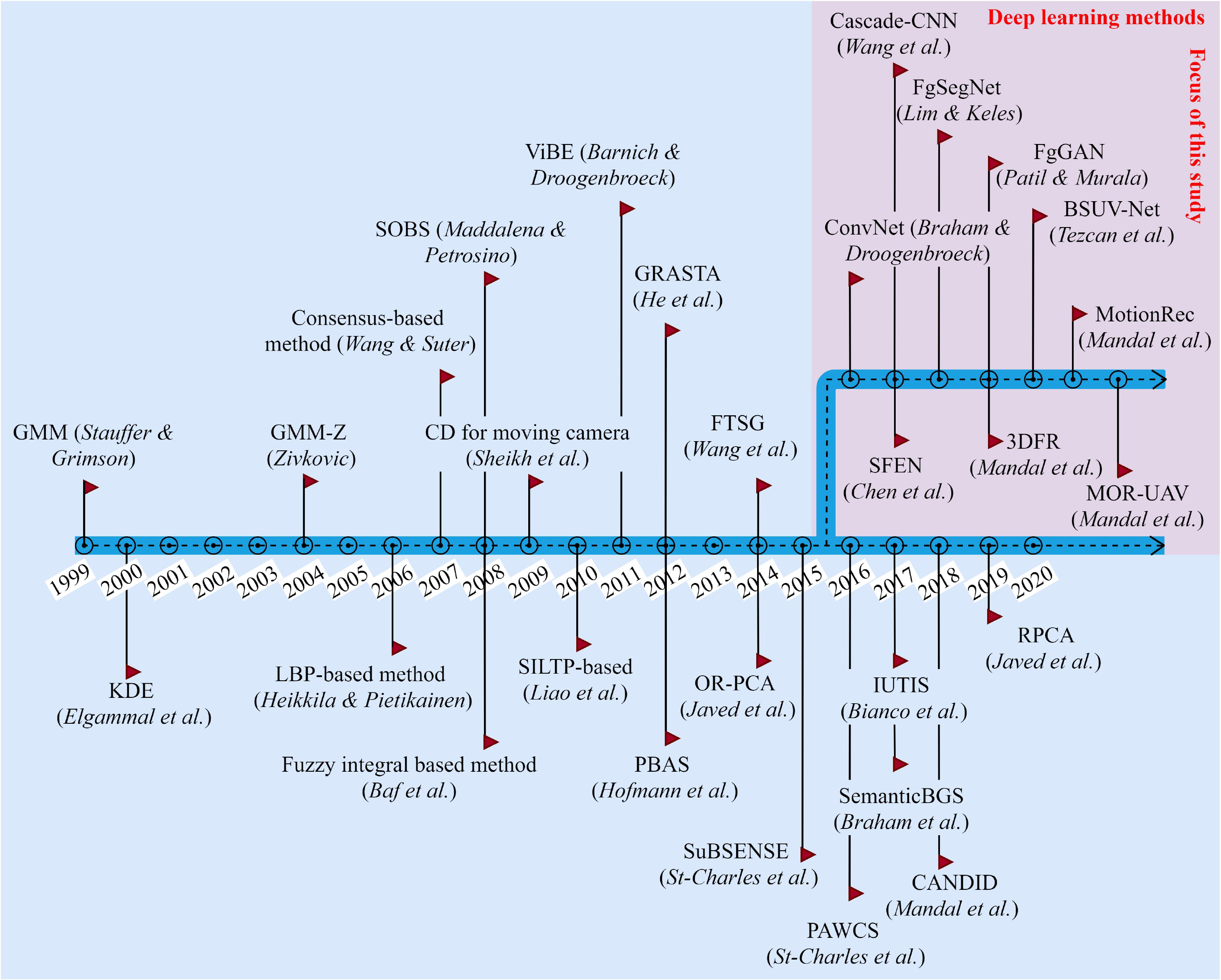}
	\caption{Milestones in change detection, including traditional methods~\cite{stauffer1999adaptive,zivkovic2004improved,wang2007consensus,heikkila2006texture,el2008fuzzy,maddalena2008self,sheikh2009background,liao2010modeling,barnich2010vibe,hofmann2012background,he2011online,wang2014static,javed2014or,st2014subsense,st2016universal,bianco2017combination,braham2017semantic,mandal2018candid,javed2018moving} and the recent deep learning frameworks~\cite{braham2016deep,wang2017interactive,babaee2018deep,chen2017pixel,bakkay2018bscgan,lim2018foreground,mandal20193dfr,patil2019fggan,tezcan2020bsuv}. The time period up to 2016 is dominated by handcrafted features. We see a turning point in 2016 with the development of DCNNs for change detection by Braham and Droogenbroeck~\cite{braham2016deep}. Most listed methods are highly cited and published in top journals or conferences. The focus of this study is highlighted and elaborated further in Fig. \ref{cdsurvey_fig5}
	}
	\label{cdsurvey_fig4}
\end{figure*}

\begin{figure*}[t]
\centering
	\includegraphics[width=0.7\textwidth ]{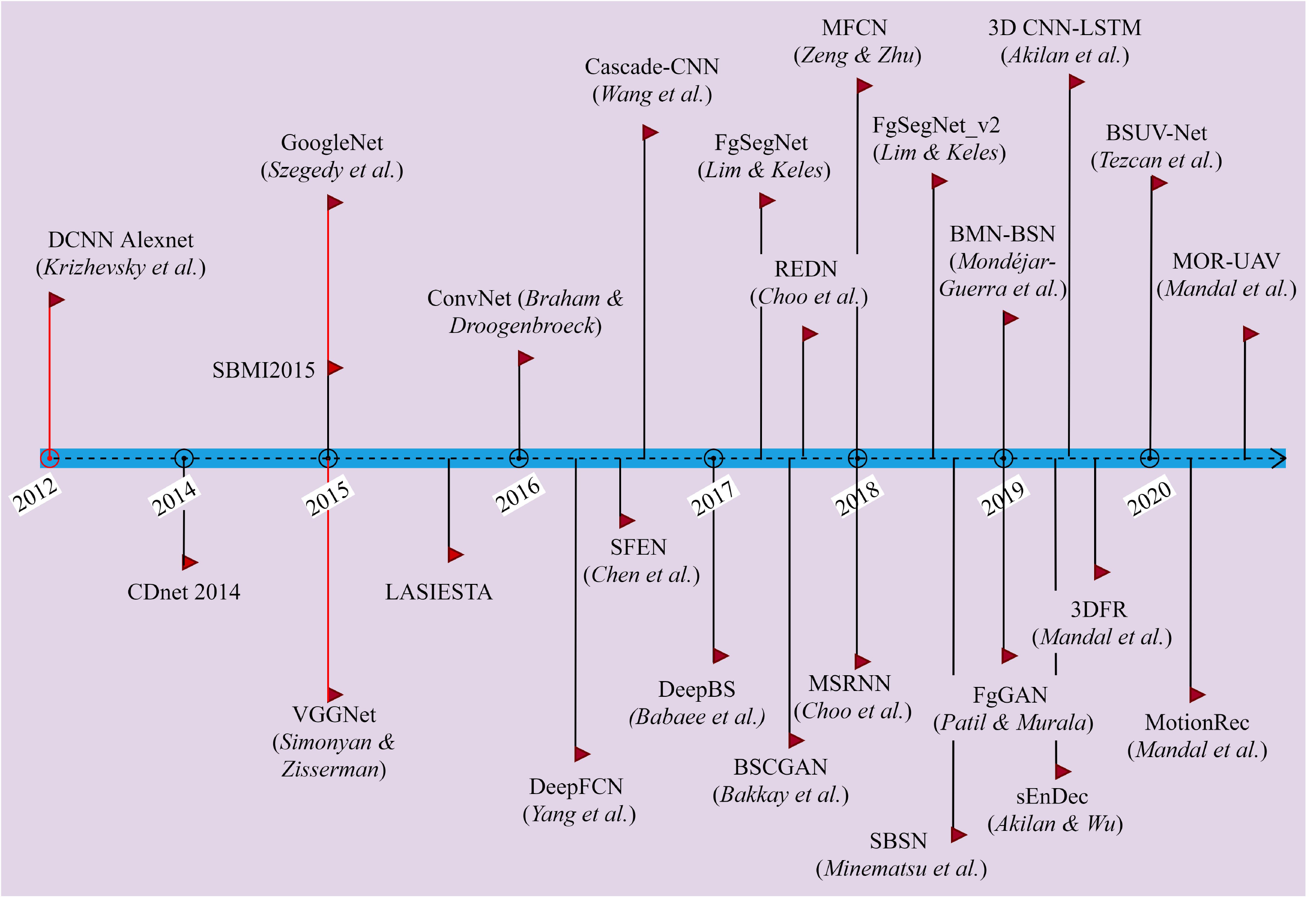}
	\caption{The focus of this study is the recent deep learning CD methods. The milestones of the most representative deep learning frameworks and datasets~\cite{krizhevsky2012imagenet,wang2014cdnet,szegedy2015going,simonyan2014very,maddalena2015towards,braham2016deep,cuevas2016labeled,yang2017deep,chen2017pixel,wang2017interactive,babaee2018deep,lim2018foreground,choo2018learning,choo2018multi,zeng2018background,lim2019learning,minematsu2019,mondejar2019end,patil2019fggan,akilan2019sendec,akilan20193d,mandal20193dfr,tezcan2020bsuv} are shown here}
	\label{cdsurvey_fig5}
\end{figure*}

\subsection{Challenges and Issues}
\begin{figure}[t]
\centering
	\includegraphics[width=0.9\textwidth ]{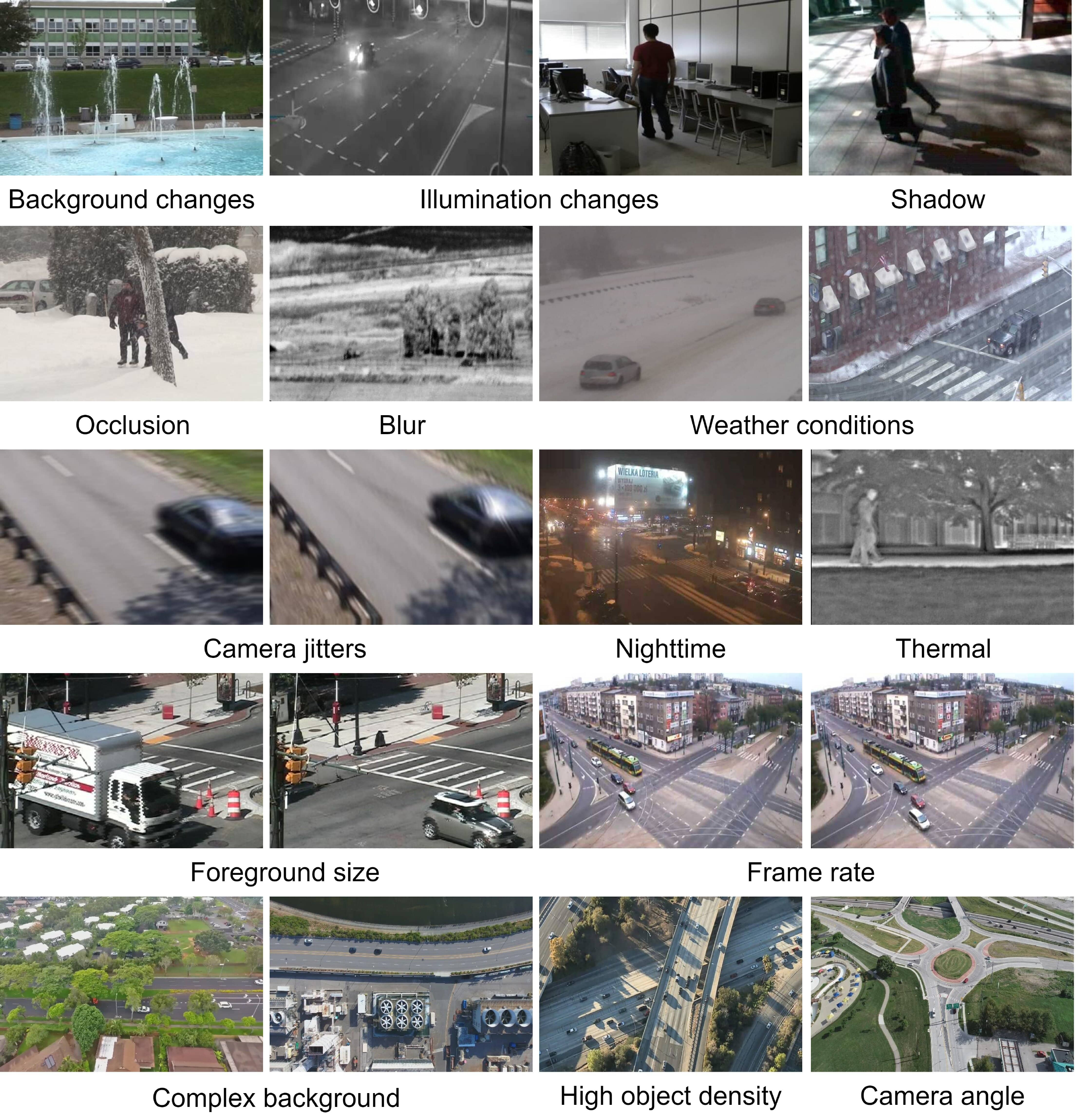}
	\caption{Visual depiction of some of the challenges in change detection}
	\label{cdsurvey_fig6}
\end{figure}

There are many real-world challenges to robust CD such as background noise, intermittent object motion, foreground scale variation, illumination changes, and extreme weather conditions (refer to Fig. \ref{cdsurvey_fig6}). The traditional methods~\cite{liao2010modeling,wang2007consensus,hofmann2012background,barnich2010vibe,jiang2017wesambe,mandal2018antic,st2014subsense,st2014improving,jiang2017wesambe,mandal2018antic,st2014subsense,mandal2018candid} address these diverse challenges by feature extraction, background modelling and thresholding to identify the changes accurately. These CD techniques do not require any labeled samples for algorithm development. These methods naturally follow scene independent strategy for performance evaluation. Thus, the evaluations across different CD datasets are uniform as all the labeled frames in the video are used for evaluation and none of them are used as priors in the traditional algorithms.\par

On the other hand, the supervised approaches require certain amount of labelled training data/samples in order to learn optimal model parameters. However, most of the existing CD datasets \cite{wang2014cdnet,cuevas2016labeled} do not provide a clear demarcation for training and testing samples. This simple observation gives rise to the question, “what should be the data division strategy for supervised change detection?”. The researchers~\cite{lin2018foreground,lim2018foreground,zeng2018multiscale,lim2017background,nguyen2018change,babaee2018deep,braham2016deep,wang2017interactive,chen2017pixel,yang2017deep,patil2018msfgnet,akilan20193d,bakkay2018bscgan} have adopted different data division schemes to evaluate and compare their supervised models with existing methods. The supervised deep learning methods have dominantly shown very high performances over these datasets. However, these approaches most prominently used SDE strategy is to select training data from certain temporal proportions of video sequence. Since the background remains more or less similar in the entire video sequence. The train and test data are highly similar. In other words, most of the deep learning models~\cite{lin2018foreground,lim2018foreground,zeng2018multiscale,lim2017background,nguyen2018change,babaee2018deep,braham2016deep,wang2017interactive,chen2017pixel,yang2017deep,patil2018msfgnet,akilan20193d,bakkay2018bscgan} have been either optimized for one specific video or a group of similar videos. We denote such training and evaluation scheme as scene dependent evaluation (SDE). In SDE, some frames from the test videos are used for training the model. This would give an unfair advantage to the CNN model in evaluation in comparison to traditional unsupervised methods.  This has led to bloated results over the benchmark datasets~\cite{wang2014cdnet,cuevas2016labeled}. The performance of the trained models has not been evaluated on unseen videos. The same warning is clearly mentioned in changedetection.net leaderboard page: “Methods with the “supervised method” tag involve a supervised machine learning algorithm potentially trained on the ground truth data used to produce the metrics reported in this page”.\par

Moreover, even for SDE setup, inconsistent data spit strategies in different papers~\cite{lin2018foreground,lim2018foreground,zeng2018multiscale,lim2017background,nguyen2018change,babaee2018deep,braham2016deep,wang2017interactive,chen2017pixel,yang2017deep,patil2018msfgnet,akilan20193d,bakkay2018bscgan} has also led to documentation of incomparable results. There is a need for clearly defined data-division schemes for training and evaluation in SDE setup. Furthermore, to evaluate the robustness of deep learning models in completely unseen videos, it is important to ensure scene independence in the evaluation of supervised CD methods. We discuss and compare the training and evaluation frameworks of different deep CD methods in detail in Section \ref{framework_compare}. \par

Another very challenging aspect is the model design of deep CD methods. Some important considerations are (i) how to initialize the background? (ii) how to model the background and/or the motion information? (iii) is support of traditional background subtraction algorithms needed?  (iv) how to pre-process and post-process the input and output data respectively? (v) which datasets or a subset of the datasets are suitable for evaluation and what are their characteristics? A number of deep learning models have been proposed which attempt to answer some of these issues for various scenarios. Different methods handle the challenges differently. Most of the models use encoder-decoder blocks to generate the binary segmentation map representing the pixel-wise changes in the current frame. Furthermore, several post-processing steps are also used for computing the final response of the CD methods. We discuss and compare the model design and characteristics of different deep CD methods in detail in Section \ref{model_compare}.

\subsection{Comparison with Previous Reviews}
Many notable surveys related to change detection have been published, as summarized in Table~\ref{tab_prev_surveys}. These include many excellent surveys of the traditional background subtraction methods~\cite{bouwmans2014traditional,sobral2014comprehensive,goyal2018review,maddalena2018background,kalsotra2019comprehensive}, deep neural network methods for background subtraction~\cite{bouwmans2019deep}, traffic monitoring~\cite{datondji2016survey,espinosa2020detection}, background initialization~\cite{bouwmans2017scene}, foreground detection~\cite{cuevas2016detection}, wide area motion detection~\cite{sommer2016survey}, maritime surveillance~\cite{prasad2017video,prasad2018object}, and moving object detection with a moving camera~\cite{yazdi2018new,chapel2020moving}.\par

There are comparatively very few surveys focusing directly on the deep learning based methods for change detection. Bouwmans et al.~\cite{bouwmans2019deep} conducted a survey of the existing deep neural network based methods. The research only discusses about the categorization of different types of networks. Moreover, the authors \textit{assume a uniform evaluation setup} among the existing methods while introducing the \textit{comparative performance evaluation tables}. Specifically, it fails to address the two important issues related to the evaluation frameworks in the literature:
\begin{itemize}
    \item The training-testing divisions in the existing deep change detection methods are different to each other. The inconsistent data-division strategies make the results in claimed by different papers incomparable to other approaches.
    \item The frames from the same video are used in both training and testing set, giving the models unfair advantage while testing. Recently, few researchers~\cite{mandal20193dfr,tezcan2020bsuv,mondejar2019end} have addressed this issue by presenting scene independent evaluation (SIE) in completely unseen videos.
\end{itemize}

The survey in~\cite{bouwmans2019deep} does not present the comparative view of the different type of evaluation strategies adopted in the existing deep learning methods. In contrast, a thorough empirical review of the existing deep learning \textit{model designs (technical characteristics)} along with the \textit{evaluation frameworks} is presented in our survey. To the best of our knowledge, this is a first attempt to comparatively analyze the different evaluation frameworks adopted in the existing deep change detection methods.

\subsection{Contributions and Organization of this Review}
\begin{figure}[t]
\centering
	\includegraphics[width=0.9\textwidth ]{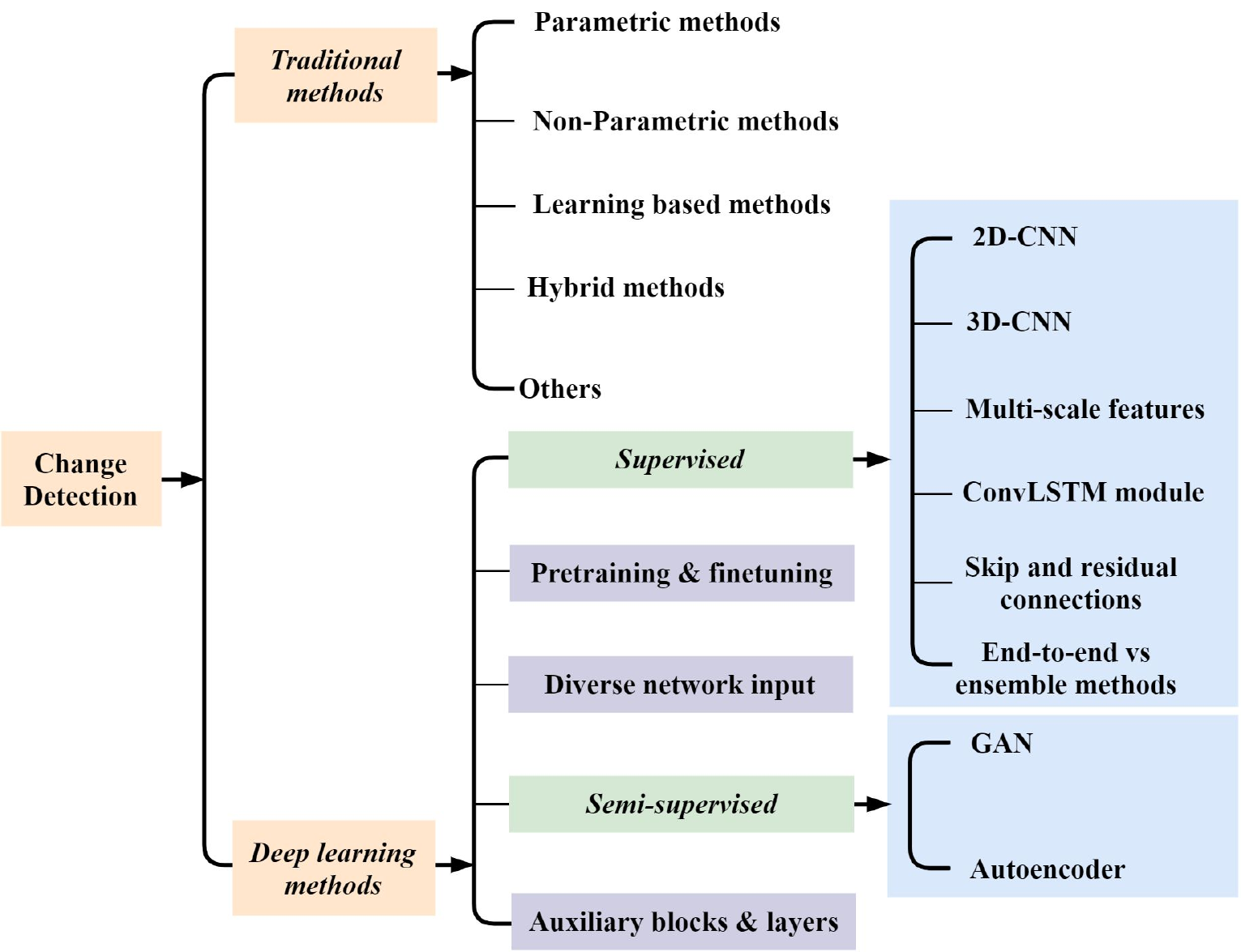}
	\caption{Flowchart of the different model designs, finetuning combinations, learning frameworks, and other architectural structures}
	\label{cdsurvey_flowchart}
\end{figure}

Motivated by the objectives discussed in the previous sections, this paper divides the CD methods into broad categories and respective subcategories to provides a comprehensive review of most representative deep learning-based CD approaches. 
Our paper presents a empirical review and analysis of the existing deep CD methods in term of design decisions, effects and best practices. Moreover, we point out some of the glaring oversight in most of the deep CD methods in terms of training and testing data division. We analyze these factors in detail and also discuss some of the possible solutions. The summary description of the large-scale CD datasets is provided. In addition, we discuss new trends in the community, and provide several interesting ideas for new methods. We hope to help readers gain valuable knowledge in deep CD algorithms and choose the most appropriate approach for their specific applications.  
We first group the CD methods in terms of algorithm characteristics. We also group the existing methods in term of the evaluation frameworks. We summarize our contributions as follows: (i) As shown in Fig. \ref{cdsurvey_flowchart}, a detailed categorization of existing approaches is provided in change detection. We classify the methods into two categories. Then, for each category, different subcategories are further defined. (ii) We provide a detailed discussion and overview of the technical characteristics of the different methods in supervised deep CD methods (refer to Table \ref{tab_modeldesign1} and Table Table \ref{tab_modeldesign2}). (iii) We categorize the training and evaluation frameworks adopted by the supervised methods in related video datasets (refer to Table \ref{tab_framework1} and Table Table \ref{tab_framework2}). After careful analysis, we identify the shortcomings in the widely adopted setup and provide directions for fair comparative analysis. We also discuss the future direction and research opportunities in Section \ref{research_needs}. We conclude our work in Section~\ref{sec_conclusion}.

\section{Deep Learning Models for Change Detection}
\label{model_compare}
The chronological advancement in the change detection algorithms is depicted in Fig. \ref{cdsurvey_fig4}. The focus of this study is highlighted in colored background. Further, in Fig. \ref{cdsurvey_fig5}, we present the milestone deep learning algorithms and datasets for change detection. We present an empirical study of these deep learning methods. As shown in Fig. \ref{cdsurvey_flowchart}, the CD methods can be grouped into traditional and deep learning-based methods. We first give a brief overview of the traditional methods. We then characterize the deep learning methods with properties such as pretrained Weights and finetuning, diverse network input, auxiliary blocks \& Layers, supervised and semi-supervised methods. The different methods are discussed in terms of the model design and other technical characteristics.

\subsection{Traditional CD Methods}
The general framework for traditional change detection techniques can be roughly divided into three stages: feature extraction, background model initialization and maintenance, and foreground detection.

\subsubsection{Feature extraction}
The low-level image features, i.e., grayscale/color intensity~\cite{zivkovic2004improved,varadarajan2013spatial,stauffer1999adaptive,wang2014static,hofmann2012background,barnich2010vibe,jiang2017wesambe,mandal2018candid,romero2017background,sajid2017universal} and edge magnitudes~\cite{roy2017adaptive,rivera2013background} are commonly used in change detection algorithms. Superpixel based features have also been used in~\cite{chen2017spatiotemporal,chen2019effective,javed2015background}. Moreover, specific spatial and spatiotemporal feature descriptors~\cite{st2014improving,st2014subsense,mandal2018antic} have been designed for enhanced performance.

\subsubsection{Background model initialization and maintenance}
The background modelling techniques can be loosely categorized into parametric~\cite{stauffer1999adaptive,zivkovic2004improved,varadarajan2013spatial,haines2013background,lee2005effective,lin2010regularized,lanza2010accurate}, non-parametric~\cite{liao2010modeling,wang2007consensus,hofmann2012background,barnich2010vibe,jiang2017wesambe,mandal2018antic,st2014subsense,st2014improving,jiang2017wesambe,mandal2018antic,st2014subsense,mandal2018candid}, traditional learning-based~\cite{maddalena2007self,maddalena2008self2,maddalena2012sobs,chacon2013improvement,gemignani2015novel,ramirez2016auto,cheng2009realtime,han2011density,yu2013online,farcas2012background,marghes2010background,marghes2012background,javed2014or,he2011online,rodriguez2016incremental,narayanamurthy2018fast}, hybrid~\cite{bianco2017combination,cheng2015hybrid,sajid2017universal,braham2017semantic} and other~\cite{st2015self,sedky2014spectral,chiranjeevi2016interval} approaches.\par

\textbf{Parametric approaches.} In parametric approaches, the distribution at each location is modeled and updated through statistical models such as mixture of Gaussians (MOG)~\cite{stauffer1999adaptive} and Expectation Maximization (EM) algorithms. Zivkovic~\cite{zivkovic2004improved} and Varadarajan et al.~\cite{varadarajan2013spatial} improved upon the MOG with variable parameter selection, spatial mixture of Gaussians and fast initialization. Similarly, extension of these models~\cite{lee2005effective,lin2010regularized} and other statistical models were also developed in literature such as Poisson distribution~\cite{faro2011adaptive}, Dirichlet distribution~\cite{haines2013background} and regression models~\cite{lanza2010accurate}. 

\textbf{Non parametric approaches.} The non-parametric methods are primarily inspired by the strategies based on kernel density estimation~\cite{liao2010modeling} and the consensus-based method~\cite{wang2007consensus}. In a seminal work ViBe~\cite{barnich2010vibe}, three significant background model maintenance policies were proposed: random background sample replacement to represent short and long-term history memoryless update policy and spatial diffusion via background
sample propagation. These strategies have been widely adopted
in recent state-of-the-art change detection techniques~\cite{hofmann2012background,jiang2017wesambe,mandal2018antic,st2014subsense,st2014improving}. Adaptive update policies for decision thresholds (for foreground segmentation) and learning rates (for model update) were introduced in~\cite{hofmann2012background}. Furthermore, adaptive feedback mechanism to continuously monitor background model fidelity and segmentation entropy to update these parameters were presented in~\cite{jiang2017wesambe,mandal2018antic,st2014subsense}. Mandal et al.~\cite{mandal2018candid} proposed a deterministic policy to update background models.\par

\textbf{Traditional learning based approaches.} Numerous learning-based techniques such as neural networks (NN), support vector machines (SVM) and principal component analysis (PCA) have also been presented in the literature. The seminal work in NN called self-organizing background subtraction (SOBS)~\cite{maddalena2007self,maddalena2008self2} was based on a 2d self-organizing neural network architecture. The network builds the image sequence neural background model by learning in a self-organizing manner, preserving pixel spatial relations. It behaves as a competitive neural network by implementing a winner-take-all function and a mechanism that updates the local weights of neurons, allowing learning to be spatially restricted to the local neighborhood of the most active neurons. Several improvements over these models have also been documented in~\cite{maddalena2012sobs,chacon2013improvement,ramirez2016auto}.\par

SVM models~\cite{cheng2009realtime,han2011density,yu2013online} have been used at different stages in background subtraction. Cheng and Gong~\cite{cheng2009realtime} generalized the one class support vector machines (1-SVMs) to accommodate spatial interactions to support online learning framework to track temporal changes over time. Similarly, Han and Davis~\cite{han2011density} estimated the background likelihood vectors for a set of features and performed background subtraction using an SVM. Similarly, others~\cite{yu2013online} have also explored the SVM models for change detection.\par

The PCA has been exploited for subspace learning to handle illumination changes in the video sequences. Earlier methods used the discriminative~\cite{farcas2012background,marghes2010background} and mixed~\cite{marghes2012background} subspace learning models. However, the regular subspace models suffer from high sensitivity to noise, outliers and missing data. To alleviate these issues, robust principal component analysis (RPCA) based models~\cite{javed2018moving,javed2016spatiotemporal,javed2017background,javed2014or} have been designed to estimate background as a low-rank component and foreground as a sparse matrix. Robust spatiotemporal subspace modelling for dynamic videos were presented in~\cite{javed2016spatiotemporal,javed2017background}. Many other incremental works are also presented to improve performance with PCA models in~\cite{he2011online,lam2020statistical,rodriguez2016incremental,narayanamurthy2018fast}.\par

\begin{figure}[t]
\centering
	\includegraphics[width=0.8\textwidth ]{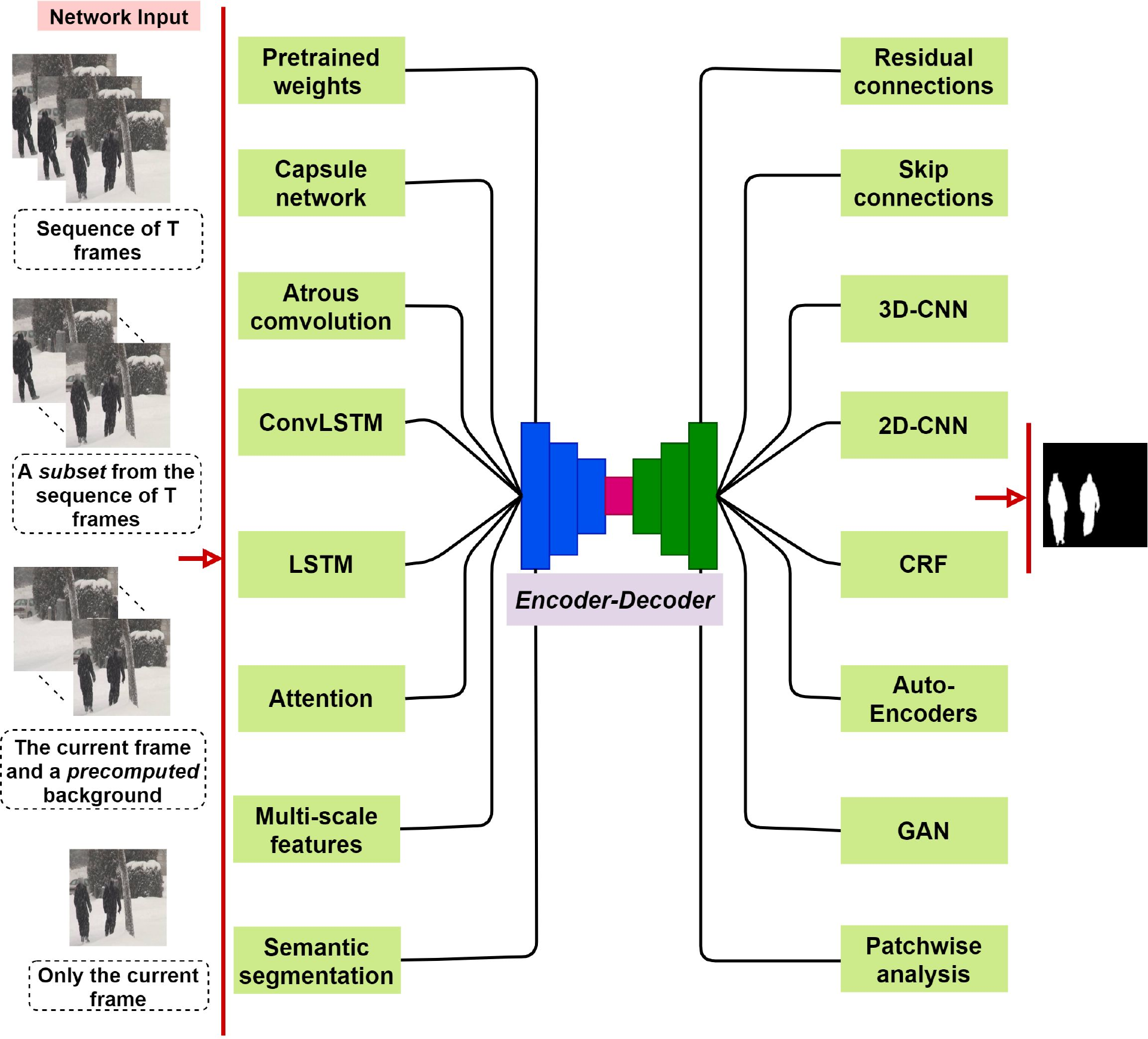}
	\caption{High level diagram of the deep learning frameworks for generic change detection. The properties of these methods are summarized in Table \ref{tab_modeldesign1} and Table \ref{tab_modeldesign2}}
	\label{cdsurvey_fig7}
\end{figure}

\textbf{Hybrid approaches.} Many works~\cite{bianco2017combination,cheng2015hybrid,sajid2017universal,braham2017semantic} have combined different modalities of algorithms to improve the performance. Bianco et al.~\cite{bianco2017combination} conducted multiple experiments to combine various change detection techniques through genetic programming. Sajid et al.~\cite{sajid2017universal} proposed to use multiple background models and fusion of RGB and YCbCr color models to estimate the background probability. Similarly, semantic segmentation~\cite{braham2017semantic} inclusion are some other interesting hybrids presented by the researchers.\par

\textbf{Other approaches.} Many other non-conventional approaches have also been successful in performing background subtraction. Local codebook-based models~\cite{st2015self}, motion modeling using graph cut and optical flow~\cite{miron2015change}, edge-based foreground segmentation~\cite{allebosch2015efic} and physics-based change detection~\cite{sedky2014spectral} are some other interesting methods proposed by the researchers to solve the problems in motion detection. Similarly, fuzzy models~\cite{chiranjeevi2016interval} have also been explored in the literature. A more detailed categorization of traditional change detection techniques can be found in~\cite{bouwmans2014traditional}. 

\subsubsection{Foreground detection}
Threshold based segmentation with post-processing techniques~\cite{barnich2010vibe,sajid2017universal,wang2007consensus,braham2017semantic} are commonly used in the existing literature for foreground detection. Numerous policies~\cite{hofmann2012background,jiang2017wesambe,mandal2018candid,st2014subsense,st2015self} have also been proposed to adaptively update the foreground segmentation thresholds. Moreover, fuzzy similarity between background model and current frame has been measured through interval similarity and membership values in~\cite{chiranjeevi2016interval}.

\begin{figure}[t]
\centering
\begin{subfigure}{1\linewidth}
    \centering
	\includegraphics[width=1\textwidth ]{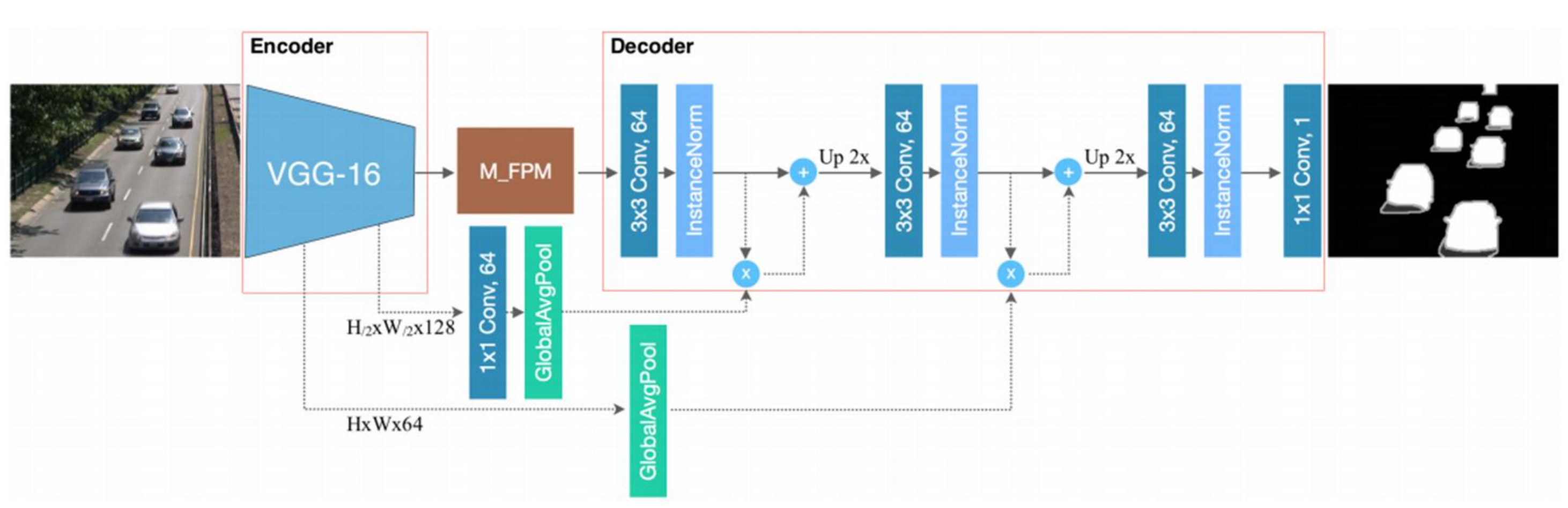}
    \caption{\cite{lim2019learning}}\label{figlit_manya}
\end{subfigure}

\begin{subfigure}{1\linewidth}
    \centering
	\includegraphics[width=1\textwidth ]{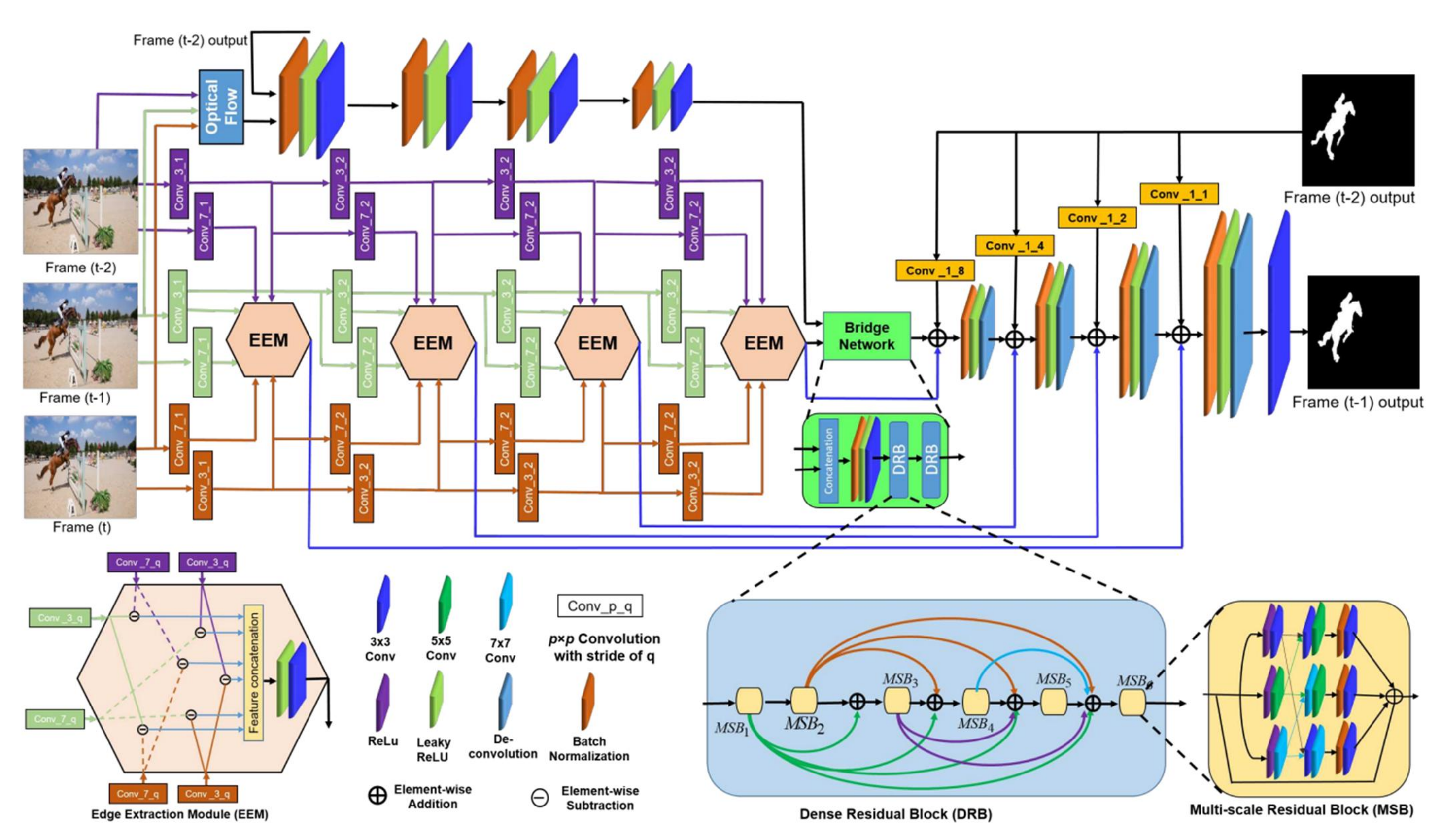}
    \caption{\cite{patil2020end}}\label{figlit_manyb}
\end{subfigure}

\begin{subfigure}{1\linewidth}
    \centering
	\includegraphics[width=1\textwidth ]{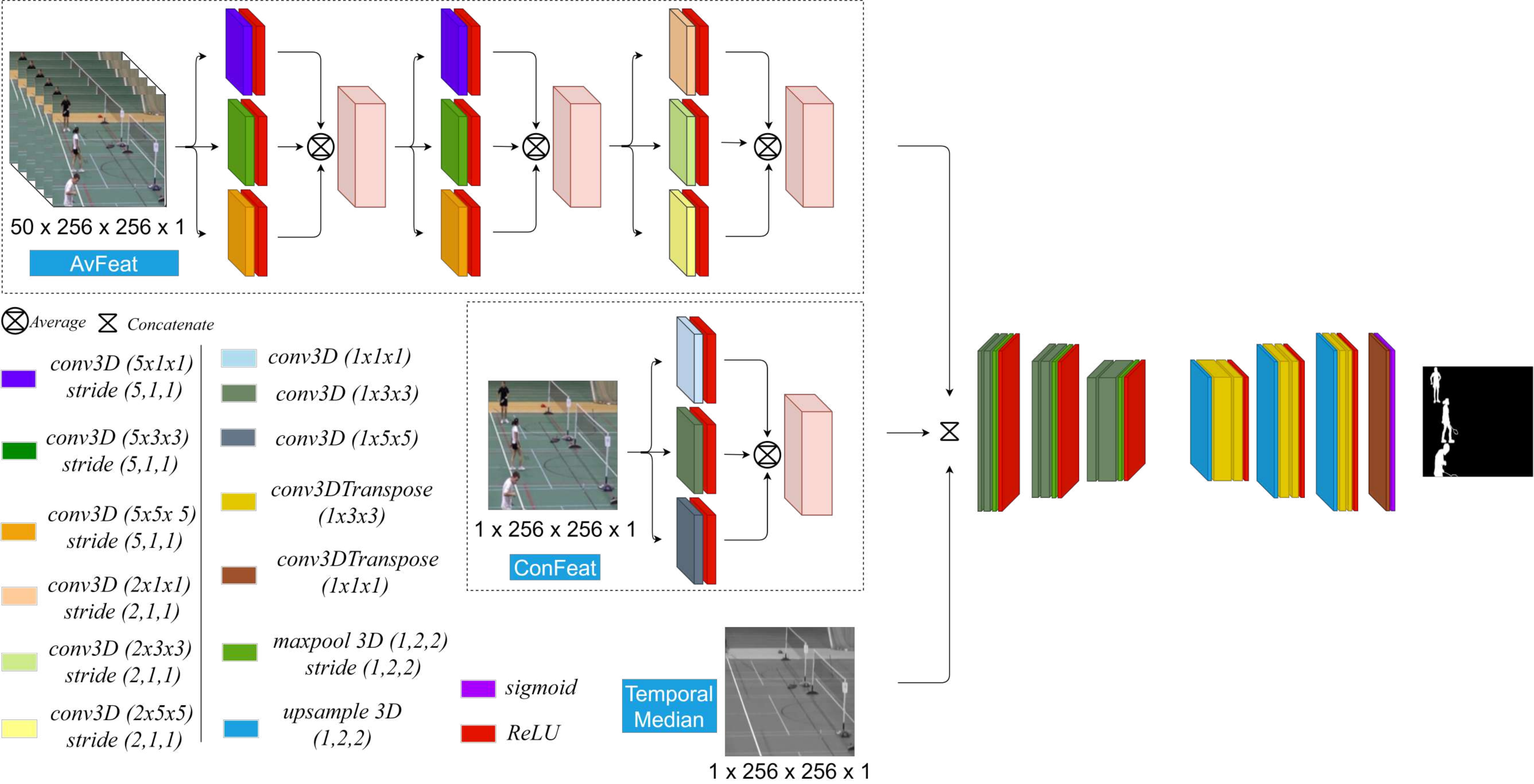}
    \caption{\cite{mandal20193dfr}}\label{figlit_manyc}
\end{subfigure}

    \caption{The change detection methods with (a) single frame~\cite{lim2019learning}, (b) three frames~\cite{patil2020end}, and (c) 50 reference frames~\cite{mandal20193dfr} for analysis. All these methods also use multi-scale features in the respective networks. The model in (b) operate with the skip, residual connections and optical flow. Whereas, the 3DFR (c) presents a 3D-CNN based architecture}
	\label{figlit_many}
\end{figure}

\subsection{Deep Learning based Methods}
Deep learning techniques have achieved superior performance as compared to the traditional hand-crafted approaches in various computer vision tasks including image classification, object detection, semantic segmentation, visual object tracking, action recognition, etc.~\cite{wei2019enhanced,mandal2019avdnet,redmon2017yolo9000,mandal2019sssdet,feng2020deep,chen2017deeplab,ji2019context,marotirao2019challenges,wan2020intelligent,li2020learning,tian2019online}. Recently, many researchers have used the convolutional neural network (CNN) to segment the video frames into foreground and background regions, i.e. change detection. The challenges in designing CNN models for CD is much different from other image and video-based problems. For example, in image classification, object detection and semantic segmentation, the features are learned only is the spatial domain. The features in the spatial dimension are sufficient for such single image-based decision-making problems. These tasks do not require attention to the features in temporal dimension. Therefore, the models designed for these tasks do not work directly in CD. Similarly, in action recognition, the features extracted from both spatial and temporal dimensions lead to prediction of high-level classification labels. However, CD demands spatiotemporal feature learning framework as well as low-level dense pixel-wise labels prediction. All, these factors make design and development of deep learning models for CD a very challenging task. We discuss the recent deep CD methods in terms of the following characteristics.

\subsubsection{Pretraining and Finetuning}
\label{sec_pretrain}
To take advantage of the foundational CNN architectures trained over large-scale image datasets, several studies have proposed the use of pretrained blocks or layers to enhance the representation capability of the CD models. The feature learning capability of off-the-shelf CNN models such as VGG16, ResNet50, GoogleNet, DeepLab and ResNet18 have been successfully adapted for change detection in~\cite{chen2017pixel,lim2017background,zeng2018background,choo2018learning,lin2018foreground,lim2018foreground,zeng2018multiscale,zhang2019x,yang2019end,ou2019moving,zeng2019combining,shahbaz2019deep,lim2019learning,tao2019universal,tezcan2020bsuv,mandal2020motionrec}. Chen et al.~\cite{chen2017pixel} designed an attention ConvLSTM to model pixel-wise changes over time. They conducted experiments with a variety of pretrained models including VGG16, ResNet50, and GoogleNet. Zeng et al.~\cite{zeng2018multiscale} proposed a novel multiscale fully convolutional network architecture which builds upon the VGG16 model. The features from different layers of the VGG16 are fused with the decoder layer of equivalent resolution. Thus, it contains both the category-level semantics and finegrained details. Similarly, Zhang et al.~\cite{zhang2019x} augment a U-Net shaped architecture to detect the pixel-level change. The pretrained weights of VGG16 has been most widely used due to its simplicity and flexibility to alter the intermediate layers for enhance the encoder-decoder for change detection~\cite{lim2017background,zeng2018background,lin2018foreground,lim2018foreground,zeng2018multiscale,zhang2019x,yang2019end,zeng2019combining,shahbaz2019deep,lim2019learning}. Ou et al.~\cite{ou2019moving} designed the CD network with ResNet18 and encoder–decoder structure in order to retain the low-level features through a shallow architecture. Tao et al.~\cite{tao2019universal} present a deep features fusion network based foreground segmentation method. The DFFnetSeg~\cite{tao2019universal} network uses both shallow layers and deep layers of a deep semantic segmentation network PSPNet. In~\cite{choo2018learning}, a multi-branch network consisting of a recurrent brach and semantic branch was presented. the semantic branch uses a DeepLabV3+ model for semantic prediction. Similarly, Tezcan et al.~\cite{tezcan2020bsuv} used the DeepLabv3 to produce semantic segmentation outcome of each frame to obtain semantically accurate foreground detections. Mandal et al.~\cite{Mandal_2020_WACV} utilize the ResNet50 backbone to train a bounding-box based detector to localize and classify only the moving objects in a video. 

\begin{figure}[]
\centering
    \includegraphics[width=0.6\textwidth]{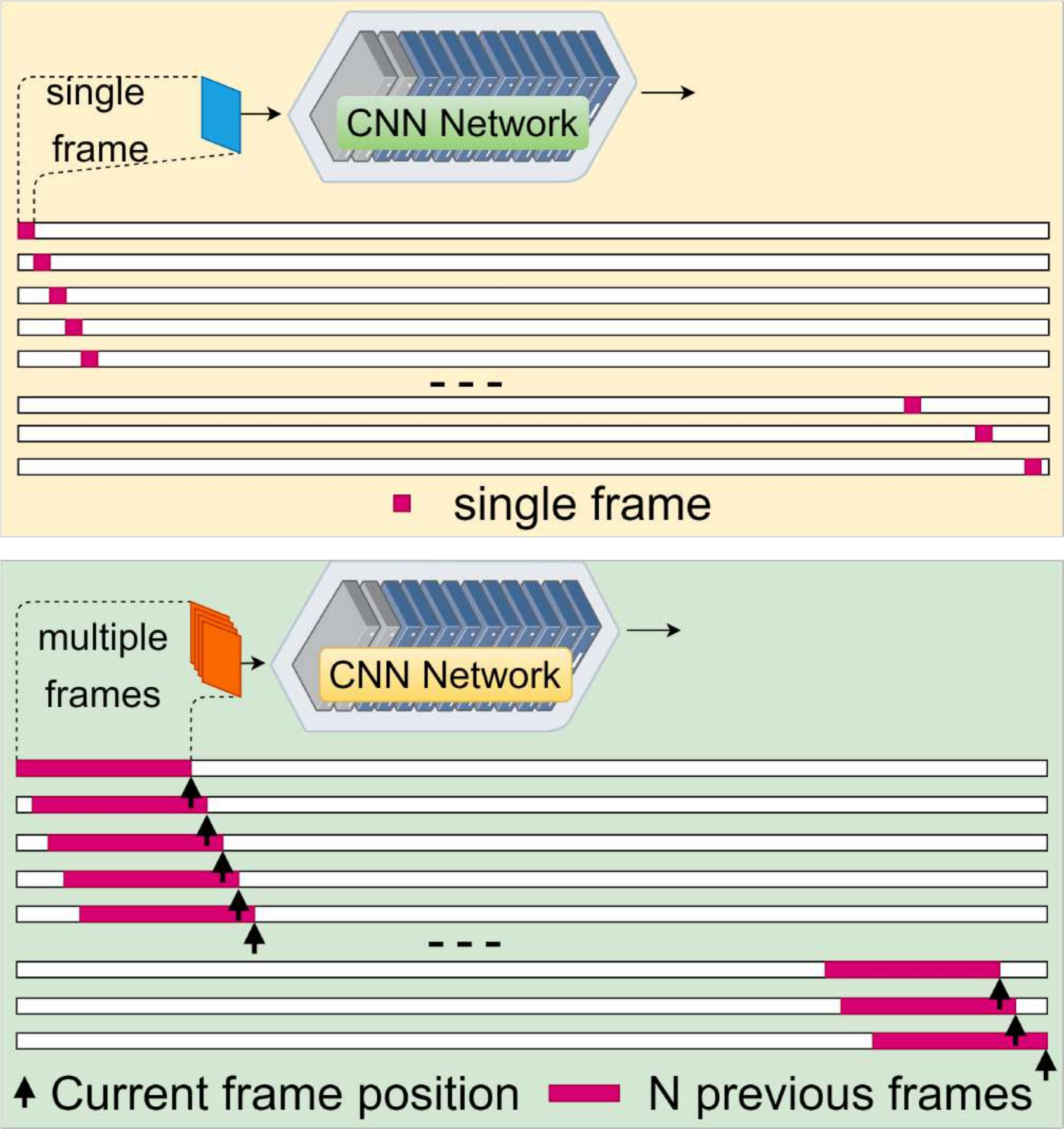}
	\caption{Difference between the training process in a video for change detection. In case of a single-frame input, the method works as a simple image segmentation network. In case of multi-frame input, the $N$ frames chunk is shifted by one frame to the right in order to process every frame}
	\label{cdsurvey_fig9}
\end{figure}

\subsubsection{Diverse Network Input}
The traditional practices either create a parametric model~\cite{stauffer1999adaptive,zivkovic2004improved,varadarajan2013spatial} of the background with few past frames or develop a non-parametric background model~\cite{st2014subsense,mandal2018candid,mandal2018antic} based on the recent 20-30 frames. The input layer to a deep learning model is defined in the network design phase. The trained model expects the input in a particular shape to compute the final output. The existing deep learning CD methods have employed diverse network inputs to train the models. We can categorize them into networks with: single frame~\cite{zhao2017joint,wang2017interactive,chen2017pixel,zeng2018background,liang2018deep,zhao2018background,choo2018multi,lim2018foreground,zeng2018multiscale,patil2018msednet,ou2019moving,zeng2019combining,shahbaz2019deep,lim2019learning,jung2019cosine,gracewell2019dynamic}, 2 frames~\cite{braham2016deep,guo2018learning,chen2018mfcnet,varghese2018changenet,lin2018foreground,nguyen2018change,babaee2018deep,bakkay2018bscgan,liao2018multiscale,zhang2019x,akilan2019sendec,akilan2019video,han2019aerial,minematsu2019,liang2019spatio,zheng2019novel}, 3-10 frames~\cite{lim2017background,akilan2018new,sakkos2018end,yang2017deep,akilan20193d,tao2019universal,tezcan2020bsuv}, 11-30 frames~\cite{choo2018learning,hu20183d,wang2018foreground1,wang2018foreground2,yang2019end,mondejar2019end,mandal2020motionrec}, and 50 frames~\cite{patil2018msfgnet,mandal20193dfr}. The methods with single frame input primarily rely on the availability of certain labeled frames in a video. The model learns the presence of foreground pixels by training a single image based binary segmentation network. Lim et al.~\cite{lim2018foreground,lim2019learning} have designed multi-scale CNN architecture with single-frame input. Patil et al.~\cite{patil2018msednet} feed a pre-processed frame to the encoder-decoder. Similarly, others~\cite{zhao2017joint,wang2017interactive,chen2017pixel,zeng2018background,liang2018deep,zhao2018background,choo2018multi,zeng2018multiscale,ou2019moving,zeng2019combining,shahbaz2019deep,jung2019cosine,gracewell2019dynamic} have also pursued such approaches. We depict the difference between the training process in single frame and multiple frames based input for change detection in Fig.~\ref{cdsurvey_fig9}. The methods with 2 frames as input aims to detect the changing pixels between them. Usually, these methods~\cite{braham2016deep,guo2018learning,chen2018mfcnet,varghese2018changenet,lin2018foreground,nguyen2018change,babaee2018deep,bakkay2018bscgan,liao2018multiscale,zhang2019x,akilan2019sendec,akilan2019video,han2019aerial,minematsu2019,liang2019spatio,zheng2019novel} first compute a background with the help of a traditional background subtraction method. Thereafter, the current frame and the background image is fed to the network to produce the result. Such inputs help the network learn the contrasting features for foreground segmentation. The methods with 3-10 frames combine multiple information from the current frame, previous frames, and the background frame. Yang et al.~\cite{yang2017deep} temporally encoded the motion information by sampling multiple images from previous frames with increasing intervals. They also designed the network using atrous convolutions, skip connections and conditional random field (CRF) layers. Akilan et al.~\cite{akilan2018new,akilan20193d} feed a 3-channel input (2-consecutive frames and a precomputed generic background) to their network. Some researchers have also used 11-30 frames as input to the network. Mondejar et al.~\cite{mondejar2019end} used 16 previous frames from a scene to generate a multidimensional map to represent the background model. Yang et al.~\cite{yang2019end} selected 14 frames to be fed to their spatio-temporal model. Mandal et al.~\cite{Mandal_2020_WACV} have conducted experiments with 10, 20, 30 frames as input to the network. Furthermore, methods~\cite{choo2018learning,hu20183d,wang2018foreground1,wang2018foreground2} have selected 12, 16, and 20 reference frames as well. Certain methods~\cite{mandal20193dfr,patil2018msfgnet} have also used 50 previous frames to model the background in and end-to-end manner for effective change detection.\par

\begin{figure}[t]
\centering
	\includegraphics[width=0.9\textwidth ]{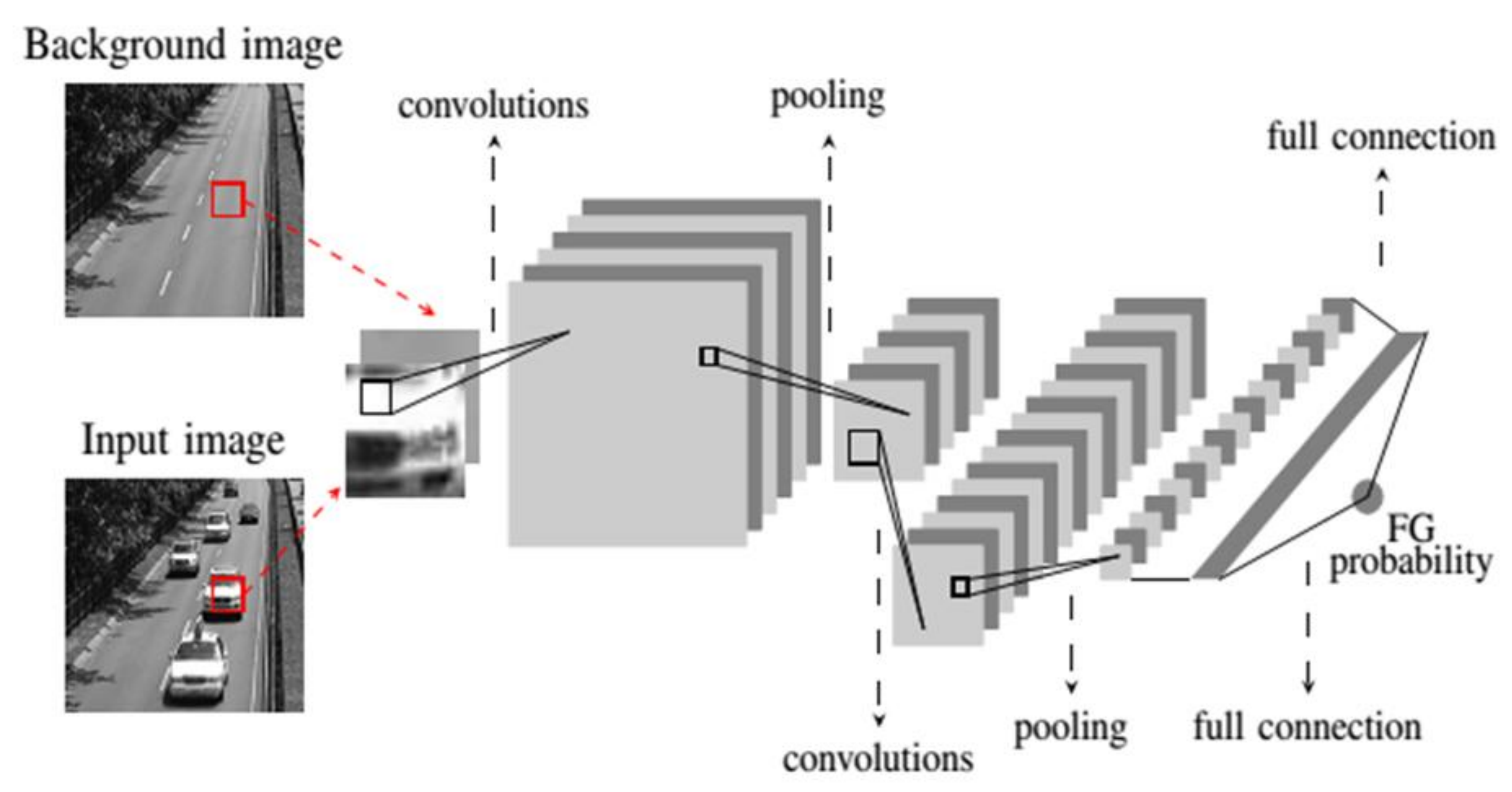}
	\caption{The representative method~\cite{braham2016deep} for patch-level analysis of the video frames for change detection}
	\label{figlit_patch_based}
\end{figure}

Some methods have also partitioned the frames into patches and use it as input layer to the network~\cite{braham2016deep,wang2017interactive,nguyen2018change,babaee2018deep,liao2018multiscale,qiu2019fully}. Babaee et al.~\cite{babaee2018deep} first generate a background image using SubSENSE and partition both the current frame and background into small patches and concatenated together to form the input layer.
The motion features are learned by training a CNN network. The final response is generated by augmenting these segmentation maps. Nguyen et al.~\cite{nguyen2018change} process the smaller patches through a triplet CNN network to extract the relevant features for change detection. Similarly, the methods in~\cite{braham2016deep,wang2017interactive,liao2018multiscale,qiu2019fully} also train the models in patch-based manner. In Fig.~\ref{figlit_patch_based}, we depict a representative existing method~\cite{braham2016deep} that follow patch-level analysis.

\begin{figure}[t]
\centering
\begin{subfigure}{0.8\linewidth}
    \centering
    \includegraphics[width=1\textwidth ]{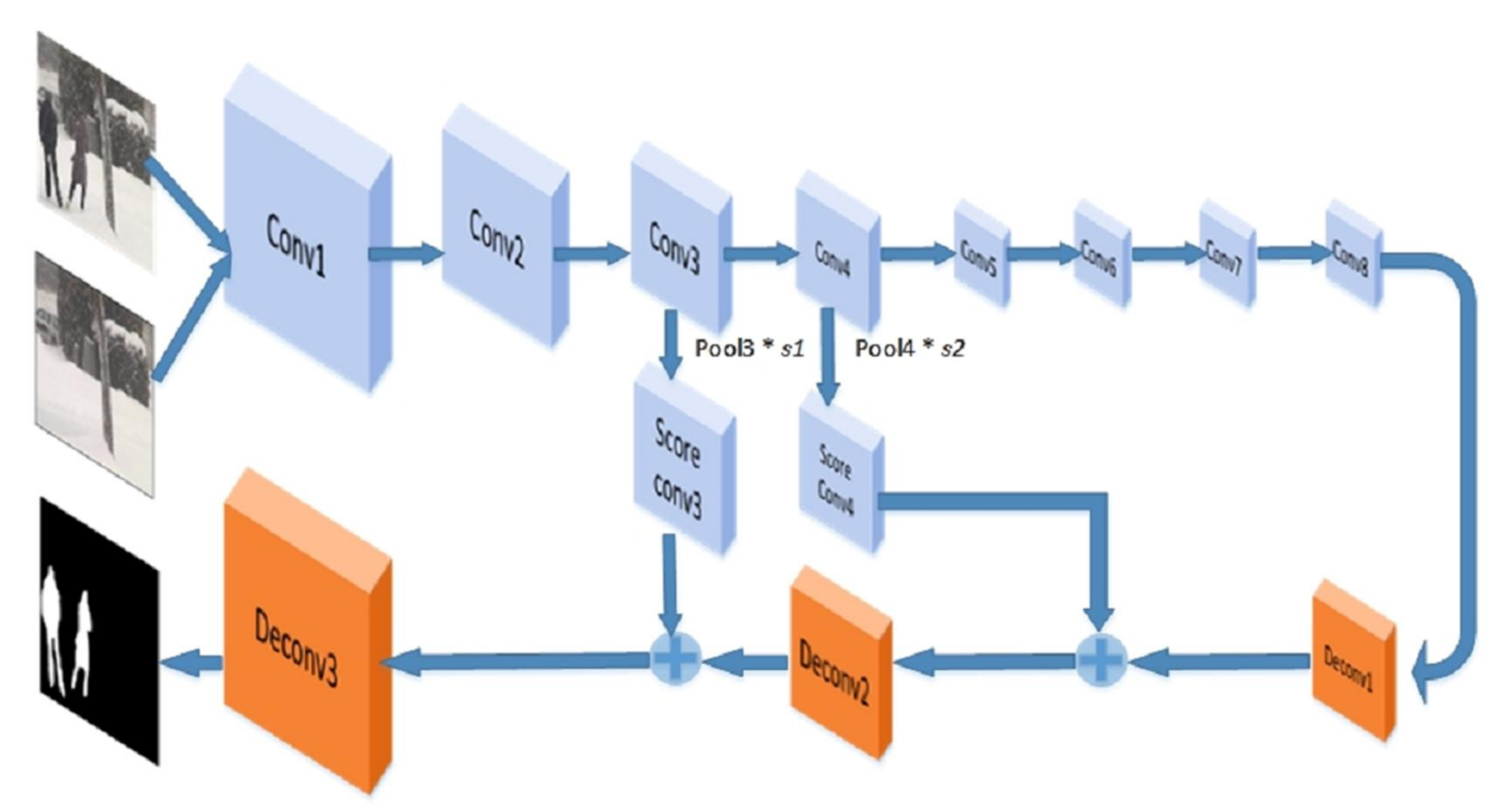}
    \caption{\cite{lin2018foreground}}\label{figlit_subsense}
\end{subfigure}

\begin{subfigure}{1\linewidth}
    \centering
    \includegraphics[width=1\textwidth ]{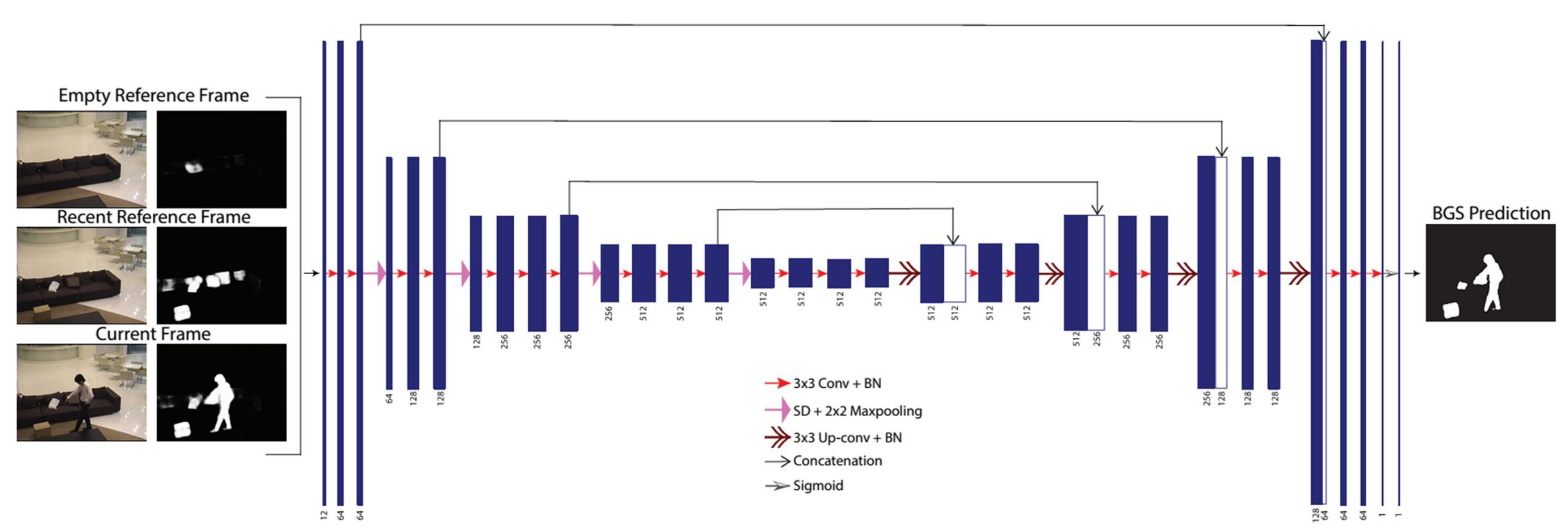}
    \caption{\cite{tezcan2020bsuv}}\label{figlit_with_median}
\end{subfigure}
    \caption{Methods with auxiliary blocks such as SuBSENSE~\cite{lin2018foreground} and simple temporal median~\cite{tezcan2020bsuv} for background estimation}
	\label{figlit_aux}
\end{figure}

\begin{table*}[t]
\footnotesize
    \centering
    \caption{Network design based comparison of the existing deep learning approaches.\textit{Year 2016-2018}}
    \resizebox{\columnwidth}{!}{
    \begin{tabular}{c|c c c c c c c c}
    \hline\hline
\multirow{2}{*}{\textbf{Pub-Yr}} &\textbf{Input}  &\textbf{Handcrafted} 	&\textbf{Network} 	&\multirow{2}{*}{\textbf{End-to-End}} &\textbf{Pretrained} &\textbf{System} &\multirow{2}{*}{\textbf{Speed}}	&\textbf{Evaluation} \\
&\textbf{frames} &\textbf{support} &\textbf{type} & &\textbf{weights} &\textbf{config.} & &\textbf{setup}\\
\hline\hline
\multirow{2}{*}{ICSSIP-16\cite{braham2016deep}}	&2 frames,	&\multirow{2}{*}{IUTIS-5}	&\multirow{2}{*}{CNN}	&\multirow{2}{*}{No}	&\multirow{2}{*}{No}	&\multirow{2}{*}{NA}	&\multirow{2}{*}{NA}	&\multirow{2}{*}{SDE}	\\
&patch based & & & & & & & \\
\hline
ICME-17\cite{zhao2017joint}	&single image &No	&CNN, Multi-task Loss	&	&Yes, DeepLab	&Titan X	&5 fps	&SDE	\\
\hline
\multirow{2}{*}{PRL-17\cite{wang2017interactive}}	&single image &\multirow{2}{*}{No}	&\multirow{2}{*}{CNN}	&\multirow{2}{*}{No}	&\multirow{2}{*}{No}	&\multirow{2}{*}{NA}	&\multirow{2}{*}{NA}	&\multirow{2}{*}{SDE}	\\
& patch based& & & & & & & \\
\hline
\multirow{2}{*}{TCSVT-17\cite{chen2017pixel}}	&\multirow{2}{*}{single image}	&\multirow{2}{*}{No}	&CNN, ConvLSTM,	&\multirow{2}{*}{Yes}	&ResNet50, VGG16,	&\multirow{2}{*}{Titan X}	&\multirow{2}{*}{5 fps}	&\multirow{2}{*}{SDE}	\\
& & &CRF & &GoogleNet & & & \\
\hline
AVSS-17\cite{lim2017background}	&3 frames	&Designed 	&CNN 	&Yes	&VGG16	&NA	&NA	&SIE\\

\hline
\multirow{2}{*}{IA-18\cite{zeng2018background}}	&\multirow{2}{*}{single image} &\multirow{2}{*}{PAWCS}	&CNN,	&\multirow{2}{*}{No}	&\multirow{2}{*}{VGG16}	&\multirow{2}{*}{Titan Xp}	&\multirow{2}{*}{20 fps}	&\multirow{2}{*}{SDE}	\\
& & &Multi-scale & & & & & \\
\hline
\multirow{2}{*}{Arxiv-18 \cite{guo2018learning}}	&\multirow{2}{*}{2 frames}	&\multirow{2}{*}{No}	&Siamese Network, &\multirow{2}{*}{Yes}	&\multirow{2}{*}{No}&\multirow{2}{*}{NA}	&\multirow{2}{*}{NA}	&SIE,\\
& & &CNN & & & & &SDE \\
\hline
ICIP-18\cite{chen2018mfcnet}	&2 frames	&No	&CNN, MatchNet	&Yes	&No	&NA	&NA	&SDE	\\
\hline
\multirow{2}{*}{ICME-18 \cite{liang2018deep}}	&\multirow{2}{*}{single image} &\multirow{2}{*}{SuBSENSE}	&CNN,	&\multirow{2}{*}{No}	&\multirow{2}{*}{No}	&\multirow{2}{*}{GTX 1080Ti}	&\multirow{2}{*}{28 fps}	&\multirow{2}{*}{SDE}	\\
& & &Multi-scale & & & & & \\
\hline
\multirow{2}{*}{ICME-18 \cite{zhao2018background}}	&single image &Random per.	&\multirow{2}{*}{CNN}	&\multirow{2}{*}{No}	&\multirow{2}{*}{No}	&\multirow{2}{*}{NA}	&\multirow{2}{*}{NA}	&\multirow{2}{*}{SDE}	\\
& &of temporal pixels & & & & & & \\
\hline
\multirow{2}{*}{ECCVW-18\cite{varghese2018changenet}}	&\multirow{2}{*}{2 frames}	&\multirow{2}{*}{No}	&CNN,	&\multirow{2}{*}{Yes}	&\multirow{2}{*}{No}	&\multirow{2}{*}{NA}	&\multirow{2}{*}{NA}	&\multirow{2}{*}{SDE}\\
& & &Siamese Network & & & & & \\

\hline
\multirow{2}{*}{ICPR-18\cite{choo2018multi}}	&\multirow{2}{*}{single image} &\multirow{2}{*}{No}	& LSTM,	&\multirow{2}{*}{Yes}	&\multirow{2}{*}{No}	&\multirow{2}{*}{NA}	&\multirow{2}{*}{NA}	&\multirow{2}{*}{SDE}	\\
& & &CNN & & & & & \\
\hline
\multirow{2}{*}{ACCV-18\cite{choo2018learning}}	&\multirow{2}{*}{20 frames}	&\multirow{2}{*}{No}	&LSTM, &\multirow{2}{*}{No}	&\multirow{2}{*}{DeepLabv3+}	&\multirow{2}{*}{NA}	&\multirow{2}{*}{NA}	&SIE,\\
& & &CNN & & & & &SDE \\
\hline
\multirow{2}{*}{MWSCAS-18\cite{akilan2018new}}	&\multirow{2}{*}{3 frames}	&temporal	&CNN, LSTM,	&\multirow{2}{*}{No}	&\multirow{2}{*}{No}	&\multirow{2}{*}{GTX 1080Ti}	&45 fps, &\multirow{2}{*}{SDE}\\
& &median & skip connection & & & &15fps & \\
\hline
\multirow{2}{*}{IA-18\cite{hu20183d}}	&\multirow{2}{*}{12 frames}	&\multirow{2}{*}{No}	&3d CNN, ConvLSTM, &\multirow{2}{*}{No}	&\multirow{2}{*}{No}	&\multirow{2}{*}{NA}	&\multirow{2}{*}{NA}	&\multirow{2}{*}{SDE}	\\
& & & atrous convolution & & & & & \\
\hline
MTAP-18\cite{sakkos2018end}	&10 frames	&No	&3d CNN	&Yes	&No	&NA	&NA	&SDE	\\
\hline
\multirow{2}{*}{Sensors-18\cite{wang2018foreground1}}	&\multirow{2}{*}{16 frames}	&\multirow{2}{*}{No}	&3D CNN,	&\multirow{2}{*}{Yes}	&\multirow{2}{*}{Sports-1M}	&\multirow{2}{*}{Titan Xp}	&\multirow{2}{*}{12 fps}	&\multirow{2}{*}{SDE}	\\
& & & atrous convolution & & & & & \\
\hline
IGRSL-18 \cite{wang2018foreground2}	&16 frames	&No	&3d CNN	&Yes	&Sports-1M	&NA	&NA	&SDE	\\
\hline
\multirow{2}{*}{ICIP-18\cite{lin2018foreground}}	&\multirow{2}{*}{2 frames}	&\multirow{2}{*}{SuBSENSE}	&CNN,	&\multirow{2}{*}{No}	&\multirow{2}{*}{VGG16}	&\multirow{2}{*}{GTX 1080Ti}	&\multirow{2}{*}{48 fps}	&\multirow{2}{*}{SIE}	\\
& & & skip connection & & & & & \\
\hline
\multirow{2}{*}{PRL-18\cite{lim2018foreground}}	&\multirow{2}{*}{single image} &\multirow{2}{*}{No}	&CNN, &\multirow{2}{*}{Yes}	&\multirow{2}{*}{VGG16}	&\multirow{2}{*}{GTX 970}	&\multirow{2}{*}{18 fps}	&\multirow{2}{*}{SDE}	\\
& & & multi-scale & & & & & \\
\hline
\multirow{2}{*}{IGRSL-18\cite{zeng2018multiscale}} &\multirow{2}{*}{single image}	&\multirow{2}{*}{No}	&CNN,	&\multirow{2}{*}{Yes}	&\multirow{2}{*}{VGG16}	&\multirow{2}{*}{GTX 1060}	&\multirow{2}{*}{27 fps}	&\multirow{2}{*}{SDE}	\\
& & & skip connection & & & & & \\
\hline
\multirow{2}{*}{TCSVT-18\cite{nguyen2018change}}	&2 frames,	&\multirow{2}{*}{SuBSENSE}	&\multirow{2}{*}{CNN}	&\multirow{2}{*}{No}	&\multirow{2}{*}{No}	&\multirow{2}{*}{NA}	&\multirow{2}{*}{NA}	&\multirow{2}{*}{SDE}	\\
&patch based & & & & & & & \\
\hline
\multirow{2}{*}{PR-18\cite{babaee2018deep}}	&2 frames,	&\multirow{2}{*}{SuBSENSE}	&\multirow{2}{*}{CNN}	&\multirow{2}{*}{No}	&\multirow{2}{*}{No}	&\multirow{2}{*}{NA}	&\multirow{2}{*}{NA}	&\multirow{2}{*}{SDE}	\\
&patch based & & & & & & & \\
\hline
\multirow{2}{*}{SMC-18\cite{patil2018msednet}}	&\multirow{2}{*}{single image}	&temporal	&CNN,	&\multirow{2}{*}{No}	&\multirow{2}{*}{No}	&\multirow{2}{*}{GTX 1080}	&\multirow{2}{*}{10 fps}	&\multirow{2}{*}{SDE}	\\
& &saliency map &multi-scale,  & & & & & \\
\hline
\multirow{3}{*}{TITS-18\cite{yang2017deep}}	&\multirow{3}{*}{6 frames}	&\multirow{3}{*}{No}	&CNN, CRF, &\multirow{3}{*}{Yes}	&\multirow{3}{*}{No}	&\multirow{3}{*}{NA}	&\multirow{3}{*}{NA}	&\multirow{3}{*}{SDE}	\\
& & &skip connection  & & & & & \\
& & &atrous convolution  & & & & & \\
\hline
TITS-18\cite{patil2018msfgnet}	&50 frames	&No	&CNN	&Yes	&No	&NA	&NA	&SDE	\\
\hline
\multirow{2}{*}{ICIP-18\cite{bakkay2018bscgan}}	&\multirow{2}{*}{2 frames} &temporal &\multirow{2}{*}{GAN}	&\multirow{2}{*}{No}	&\multirow{2}{*}{No}	&\multirow{2}{*}{GTX 1080}	&\multirow{2}{*}{400 fps}	&\multirow{2}{*}{SDE}	\\
& &median & & & & & & \\
\hline
\multirow{2}{*}{PRCM-18\cite{liao2018multiscale}}	&2 frames,	&\multirow{2}{*}{IUTIS-5}	&CNN,	&\multirow{2}{*}{No}	&\multirow{2}{*}{No}	&\multirow{2}{*}{NA}	&\multirow{2}{*}{NA}	&\multirow{2}{*}{SDE}	\\
&patch-based & &multi-scale & & & & & \\
\hline
\hline
    \end{tabular}
 }
    \label{tab_modeldesign1}
\end{table*}

\subsubsection{Auxiliary blocks and layers}
\label{sec_auxblock}
Several deep learning methods employ the statistical and hand-crafted background modelling techniques for temporal feature encoding. The reference background image is used along with the current frame for change detection. 
Similarly, some studies have proposed the addition of well-designed auxiliary blocks or layers to enhance the motion-related representation capability of the network. The following statistical auxiliary blocks have been used in the literature: SuBSENSE~\cite{liang2018deep,lin2018foreground,nguyen2018change,babaee2018deep,zeng2019combining}, IUTIS~\cite{braham2016deep,liao2018multiscale}, PAWCS~\cite{zeng2018background}, designed algorithm~\cite{lim2017background,zhao2018background,tao2019universal}, temporal median~\cite{akilan2018new,bakkay2018bscgan,akilan2019sendec,akilan2019video,minematsu2019,mandal20193dfr,zheng2019novel,tezcan2020bsuv}, temporal histogram and motion saliency map~\cite{patil2019fggan,patil2018msednet,patil2019motion}, optical flow~\cite{liang2019spatio,patil2020end,sultana2019unsupervised,vijayan2020universal}, and conditional random fields (CRF)~\cite{wang2017interactive,chen2017pixel}.\par

Nguyen et al.~\cite{nguyen2018change} generated the
background image using proven hand-crafted approaches SubSENSE. The SubSENSE background estimation (BE) block has also been used in ~\cite{liang2018deep,lin2018foreground,babaee2018deep,zeng2019combining}. Similarly, other proven methods such as IUTIS~\cite{braham2016deep,liao2018multiscale} and PAWCS~\cite{zeng2018background} have also been added as the BE block. Fig.~\ref{figlit_subsense} shows an existing method~\cite{lim2018foreground} using SuBSENSE as a BE block. Zhao et al.~\cite{zhao2018background} designed a RPoTP block to capture the random permutation of temporal feature in a particular pixel. The historical observations at each pixel are permuted at random. The RPoTP feature map is subtracted from the current frame for subsequent processing by the CNN model. Lim et al.~\cite{lim2017background} and Tao et al.~\cite{tao2019universal} have also augmented designed blocks to encode the historical patterns. The temporal median has been quite frequently used as a simple temporal feature encoder in~\cite{akilan2018new,bakkay2018bscgan,akilan2019sendec,akilan2019video,minematsu2019,mandal20193dfr,zheng2019novel,tezcan2020bsuv}. Fig.~\ref{figlit_with_median} depicts an existing method~\cite{tezcan2020bsuv} using temporal median as one of the inputs to the network. Whereas, Patil et al.~\cite{patil2019fggan,patil2018msednet,patil2019motion} have explored the temporal histogram and motion saliency maps. Some researchers have proposed to exploit the optical flow~\cite{liang2019spatio,patil2020end}, CRFs~\cite{wang2017interactive,chen2017pixel}, semantic segmentation~\cite{choo2018learning,tezcan2020bsuv} to refine the foreground segmentation map.

\begin{table*}[t]
\footnotesize
    \centering
    \caption{Network design based comparison of the existing deep learning approaches.\textit{Year 2019-2020}}
    \begin{tabular}{c|c c c c c c c c}
    \hline\hline
\multirow{2}{*}{\textbf{Pub-Yr}} &\textbf{Input}  &\textbf{Handcrafted}	&\textbf{Network}	&\multirow{2}{*}{\textbf{End-to-End}} &\textbf{Pretrained} &\textbf{System} &\multirow{2}{*}{\textbf{Speed}}	&\textbf{Evaluation} \\
&\textbf{frames} &\textbf{support} &\textbf{type} & &\textbf{weights} &\textbf{config.} & &\textbf{setup}\\
\hline\hline
\multirow{2}{*}{IA-19\cite{qiu2019fully}} &15 frames	&\multirow{2}{*}{No}	&\multirow{2}{*}{LSTM, CNN}	&\multirow{2}{*}{No}	&\multirow{2}{*}{No}	&\multirow{2}{*}{Titan Xp}	&\multirow{2}{*}{36 fps}	&\multirow{2}{*}{SDE}	\\
&patch-based & & & & & & & \\
\hline
\multirow{2}{*}{IA-19\cite{zhang2019x}}	&\multirow{2}{*}{2 frames}	&\multirow{2}{*}{No}	&CNN,	&\multirow{2}{*}{Yes}	&\multirow{2}{*}{VGG16}	&\multirow{2}{*}{GTX 1080Ti}	&\multirow{2}{*}{22 fps}	&\multirow{2}{*}{SDE}	\\
& & &multi-scale & & & & & \\
\hline
\multirow{2}{*}{IA-19\cite{yang2019end}}	&\multirow{2}{*}{14 frames}	&\multirow{2}{*}{No}	&ConvLSTM, CNN,	&\multirow{2}{*}{Yes}	&\multirow{2}{*}{VGG16}	&\multirow{2}{*}{GTX 1080Ti}	&\multirow{2}{*}{11 fps}	&\multirow{2}{*}{SDE}	\\
& & &multi-scale & & & & & \\
\hline
IA-19\cite{ou2019moving}	&single image	&No	&CNN, residual	&	&ResNet18	&NA	&NA	&SDE	\\
\hline
\multirow{2}{*}{JEI-19\cite{zeng2019combining}}	&\multirow{2}{*}{single image} &SuBSENSE,
FTSG, &CNN,	&\multirow{2}{*}{Yes}	&\multirow{2}{*}{VGG16}	&\multirow{2}{*}{NA}	&\multirow{2}{*}{NA}	&\multirow{2}{*}{SDE}	\\
& &CwisarDH &skip connections & & & & & \\
\hline

\multirow{2}{*}{TITS-19\cite{akilan2019sendec}}	&\multirow{2}{*}{2 frames}	&temporal &CNN,	&\multirow{2}{*}{Yes}	&\multirow{2}{*}{No}	&\multirow{2}{*}{NA}	&\multirow{2}{*}{NA}	&\multirow{2}{*}{SDE}	\\
& &median &skip connections & & & & & \\
\hline
\multirow{2}{*}{TVT-19\cite{akilan2019video}}	&\multirow{2}{*}{2 frames}	&temporal	&CNN,inception,	&\multirow{2}{*}{Yes}	&\multirow{2}{*}{No}	&\multirow{2}{*}{GTX 1080Ti}	&\multirow{2}{*}{42 fps}	&\multirow{2}{*}{SDE}	\\
& &median &residual  & & & & & \\
\hline
\multirow{3}{*}{Neuro-19\cite{han2019aerial}}	&\multirow{3}{*}{2 frames}	&\multirow{3}{*}{No}	&GAN, CNN	&\multirow{3}{*}{Yes}	&\multirow{3}{*}{Yes}	&\multirow{3}{*}{NA}	&\multirow{3}{*}{NA}	&\multirow{3}{*}{NA}	\\
& & &ResNet, & & & & & \\
& & &RoI Detection & & & & & \\
\hline
ISIE-19\cite{shahbaz2019deep}	&single image &No	&CNN	&Yes	&VGG16	&NA	&Na	&SDE	\\
\hline
\multirow{2}{*}{WACV-19\cite{patil2019fggan}}	&\multirow{2}{*}{single image}	&temporal 	&\multirow{2}{*}{GAN}	&\multirow{2}{*}{No}	&\multirow{2}{*}{No}	&\multirow{2}{*}{NA}	&\multirow{2}{*}{NA}	&\multirow{2}{*}{SDE}	\\
& &histogram & & & & & & \\
\hline
BMVC-19\cite{mondejar2019end}	&16 frames	&No	&CNN, U-Net	&Yes	&No	&NA	&NA	&SIE	\\
\hline
TITS-19\cite{akilan20193d}	&4 frames	&No	&3D CNN, LSTM	&Yes	&No	&GTX 1080Ti	&24 fps	&SDE	\\
\hline
\multirow{2}{*}{PAA-19\cite{lim2019learning}}	&\multirow{2}{*}{single image}	&\multirow{2}{*}{No}	&CNN,	&\multirow{2}{*}{Yes}	&\multirow{2}{*}{VGG16}	&\multirow{2}{*}{NA}	&\multirow{2}{*}{NA}	&\multirow{2}{*}{SDE}	\\
& & &Skip connections & & & & & \\
\hline
\multirow{2}{*}{AVSS-19\cite{patil2019motion}}	&\multirow{2}{*}{single image} &motion	&\multirow{2}{*}{GAN}	&\multirow{2}{*}{No}	&\multirow{2}{*}{No}	&\multirow{2}{*}{NA}	&\multirow{2}{*}{NA}	&\multirow{2}{*}{SDE}	\\
& &saliency map & & & & & & \\
\hline
\multirow{2}{*}{IA-19\cite{tao2019universal}}	&\multirow{2}{*}{3 frames}	&designed 	&\multirow{2}{*}{CNN}  &\multirow{2}{*}{No}	&\multirow{2}{*}{PSPNet}	&\multirow{2}{*}{NA}	&\multirow{2}{*}{NA}	&\multirow{2}{*}{SDE}	\\
& &algorithm & & & & & & \\
\hline
\multirow{2}{*}{AVSS-19\cite{minematsu2019}}	&\multirow{2}{*}{2 frames}	&temporal	&CNN,	&\multirow{2}{*}{No}	&\multirow{2}{*}{No}	&\multirow{2}{*}{GTX 1080Ti}	&\multirow{2}{*}{134 fps}	&\multirow{2}{*}{SDE}	\\
& &median &skip connection & & & & & \\
\hline
\multirow{2}{*}{AVSS-19\cite{jung2019cosine}}	&\multirow{2}{*}{single image}	&\multirow{2}{*}{No}	&CNN, multi-scale &\multirow{2}{*}{Yes}	&\multirow{2}{*}{No}	&\multirow{2}{*}{NA}	&\multirow{2}{*}{NA}	&\multirow{2}{*}{SDE}	\\
& & &cosine focal loss & & & & & \\
\hline
Sensors-19\cite{liang2019spatio}	&2 frames	&Optical Flow	&CNN, Attention	&Yes	&No	&RTX 2080Ti	&11 fps	&SDE	\\
\hline
\multirow{2}{*}{ISPL-19\cite{mandal20193dfr}}	&\multirow{2}{*}{50 frames}	&temporal	&3D-CNN, &\multirow{2}{*}{Yes}	&\multirow{2}{*}{No}	&\multirow{2}{*}{Titan Xp}	&\multirow{2}{*}{33 fps}	&\multirow{2}{*}{SIE}	\\
& &median &multi-scale & & & & & \\
\hline
MVA-19\cite{sultana2019unsupervised}	&single image &Optical Flow	&GAN	&No	&VGG19	&NA	&NA	&SIE	\\
\hline
\multirow{2}{*}{Neuro-19\cite{zheng2019novel}}	&\multirow{2}{*}{2 frames}	&temporal &\multirow{2}{*}{GAN}	&\multirow{2}{*}{No}	&\multirow{2}{*}{No}	&\multirow{2}{*}{GTX 970}	&\multirow{2}{*}{23 fps}	&\multirow{2}{*}{SDE}	\\
& &median & & & & & & \\
\hline
MTAP-19 \cite{gracewell2019dynamic}	&single image &No	&Autoencoder	&No	&No	&NA	&NA	&SIE	\\
\hline
\multirow{2}{*}{WACV-20\cite{tezcan2020bsuv}}	&\multirow{2}{*}{3 frames}	&temporal 	&CNN,	&\multirow{2}{*}{No}	&\multirow{2}{*}{DeepLabv3}	&\multirow{2}{*}{NA}	&\multirow{2}{*}{NA}	&\multirow{2}{*}{SIE}	\\
& &median &skip connections & & & & & \\
\hline
\multirow{2}{*}{WACV-20\cite{mandal2020motionrec}}	&10/20/30 	&\multirow{2}{*}{No}	&CNN, bounding
	&\multirow{2}{*}{Yes}	&\multirow{2}{*}{ResNet50}	&\multirow{2}{*}{Titan Xp}	&\multirow{2}{*}{5 fps}	&\multirow{2}{*}{SIE}	\\
&frames & &box regression & & & & & \\
\hline
    \end{tabular}
    \label{tab_modeldesign2}
\end{table*}

\subsubsection{Supervised methods}
The most commonly adopted framework for deep learning-based CD is the supervised setup. The methods use the manually labeled ground truths from the respective datasets for model training. The existing deep supervised CD methods can be categorized according to the following technical characteristics:

\begin{figure}[t]
\centering
\begin{subfigure}{1\linewidth}
    \centering
    \includegraphics[width=1\textwidth ]{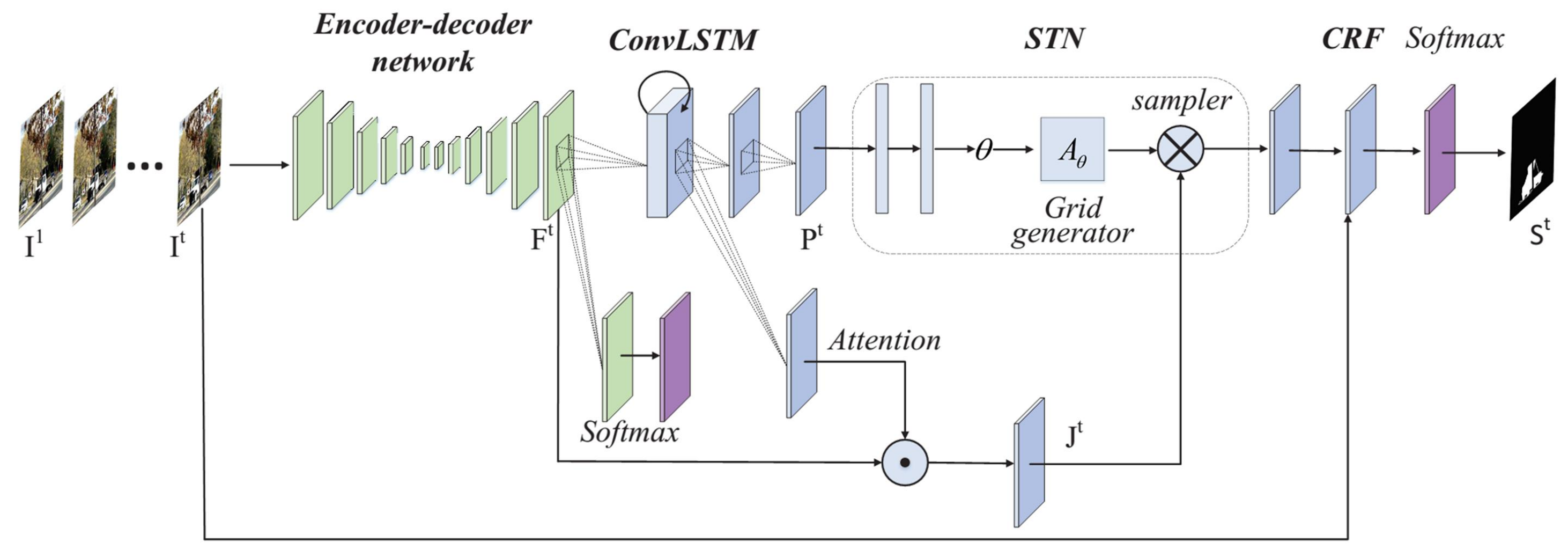}
    \caption{\cite{chen2017pixel}}\label{figlit_lstma}
\end{subfigure}

\begin{subfigure}{1\linewidth}
    \centering
    \includegraphics[width=1\textwidth ]{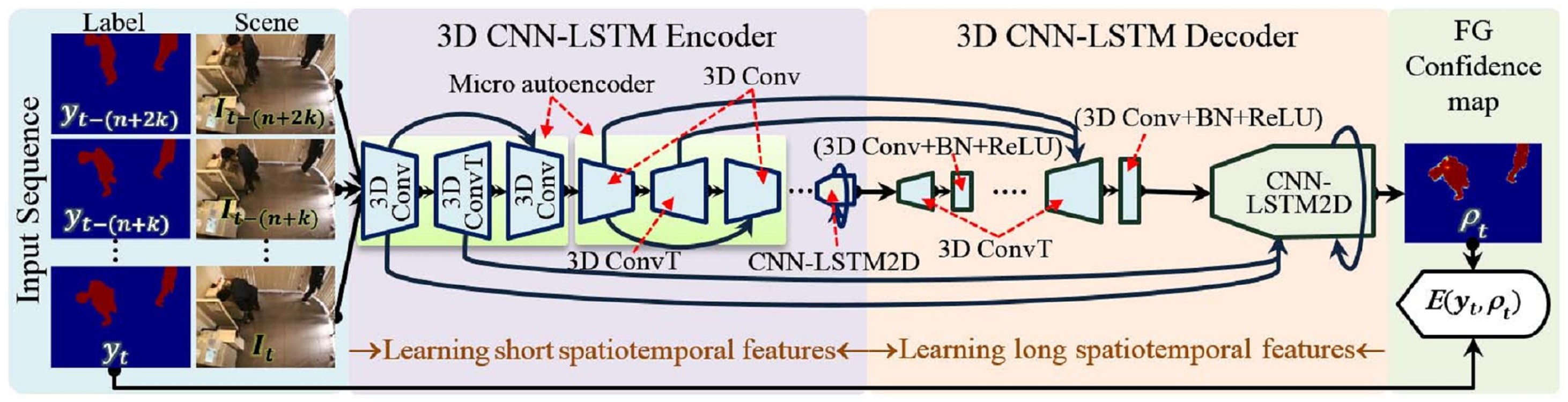}
    \caption{\cite{akilan20193d}}\label{figlit_lstmb}
\end{subfigure}
	\caption{ConvLSTM module based learning frameworks~\cite{chen2017pixel,akilan20193d}.}
	\label{figlit_lstm}
\end{figure}

\textbf{2D-CNN.} Most of the existing deep CD methods in the literature have designed 2D-CNN models~\cite{braham2016deep,zhao2017joint,wang2017interactive,chen2017pixel,lim2017background,zeng2018background,guo2018learning,chen2018mfcnet,liang2018deep,zhao2018background,varghese2018changenet,choo2018multi,choo2018learning,akilan2018new,lin2018foreground,lim2018foreground,zeng2018multiscale,nguyen2018change,babaee2018deep,patil2018msednet,yang2017deep,patil2018msfgnet,liao2018multiscale,qiu2019fully,zhang2019x,yang2019end,ou2019moving,zeng2019combining,akilan2019sendec,akilan2019video,shahbaz2019deep,mondejar2019end,lim2019learning,tao2019universal,minematsu2019,jung2019cosine,liang2019spatio,tezcan2020bsuv,mandal2020motionrec,cai2020background} to map the motion information. Learning the spatiotemporal features directly with 2D convolutional kernels is a non-trivial problem. Therefore, the researchers have designed additional blocks and layers (discussed in Section \ref{sec_auxblock}) to produce the contrasting effect for robust change detection. Patil et al.~\cite{patil2018msfgnet} have created average pooling layer in temporal direction (compatible in 2D-CNN network) to enable end-to-end operation without dependence on auxiliary modules. Using the 2D-CNN architecture also facilitate taking advantage of transfer learning with the pretrained weights (discussed in Section \ref{sec_pretrain}). The general design for all the existing methods resemble the encoder-decoder network structure as shown in Fig. \ref{cdsurvey_fig7}. Some existing methods using 2D-CNN are depicted in Fig.~\ref{figlit_manya}, Fig.~\ref{figlit_manyb}, Fig.~\ref{figlit_aux}, and Fig.~\ref{figlit_patch_based}.

\textbf{3D-CNN.} Studies in~\cite{hu20183d,sakkos2018end,wang2018foreground1,wang2018foreground2,akilan20193d,mandal20193dfr} show that performing 3D convolutions is a rewarding
approach to capture both spatial and temporal dimensional features in videos. The effectiveness of 3D-CNN for change detection was first presented in~\cite{hu20183d}. A more robust model validated over completely unseen videos was presented in~\cite{mandal20193dfr}. The authors designed 3DFR (refer Fig.~\ref{figlit_manyc}) to learn the spatiotemporal features through a swift feature reductionist approach in an end-to-end manner. Akilan et al.~\cite{akilan20193d} used the 3D convolutions (refer Fig.~\ref{figlit_lstmb}) to capture the short temporal motions while using LSTM to capture the long-short term temporal motions. Similarly, other 3D CNN network designs~\cite{sakkos2018end,wang2018foreground1,wang2018foreground2} have also been explored in the literature.  

\textbf{Multi-scale features.} Multi-scale feature representations have been successfully used in semantic segmentation applications to achieve robust performance~\cite{zhao2018icnet,zhang2018fully,chen2017deeplab,yang2018denseaspp}. Numerous CD approaches have also utilized multi-scale features (refer Fig.~\ref{figlit_many}) in the network~\cite{zeng2018background,liang2018deep,lim2018foreground,patil2018msednet,liao2018multiscale,zhang2019x,yang2019end,jung2019cosine,mandal20193dfr}. Lim et al.~\cite{lim2018foreground} designed a triplet CNN that operates in three different scales for feature encoding. Mandal et al.~\cite{mandal20193dfr} encode features from multiple 3D receptive fields $T\times1\times1$, $T\times3\times3$ and $T\times5\times5$ ($T$ denoting the kernel depth) and then take the average of the feature maps. Yang et al.~\cite{yang2019end} improve the robustness of background subtraction with an end-to-end multi-scale spatiotemporal (MS-ST) method. Likewise, the multi-scale features have been successfully employed in~\cite{zeng2018background,liang2018deep,patil2018msednet,liao2018multiscale,zhang2019x,jung2019cosine}.

\textbf{ConvLSTM module.} To exploit the pixel-level temporal context, the change detection can be considered as a sequence labeling problem. 
Chen et al.~\cite{chen2017pixel} proposed an attention ConvLSTM to model temporal changes over time. Similarly, Choo et al.~\cite{choo2018multi,choo2018learning} and Yang et al.~\cite{yang2019end} used a multi-scale ConvLSTM structure to model various types of spatial and temporal changes. Akilan et al.~\cite{akilan20193d} incorporate the cues from the LSTM module with the 3D-CNN network to robustly detect the moving objects. Some other notable works~\cite{hu20183d,qiu2019fully} also include such modules in their network. 

\textbf{Skip and Residual connections.} Usually, the first few layers consists of the low-level features as compared to the abstract high-level features at the deeper layers in the CNN network. The skip and residual connections help in preserving the detailed low-level features which contributes to more accurate pixel-wise binary segmentation at the final layer. Therefore several methods~\cite{lim2017background,akilan2018new,lin2018foreground,zeng2018multiscale,yang2017deep,ou2019moving,zeng2019combining,akilan2019sendec,akilan2019video,han2019aerial,lim2019learning,minematsu2019,tezcan2020bsuv,zhang2019x} have used skip and residual connections in the network to obtain better performance (refer Fig.~\ref{figlit_manyb}, Fig.~\ref{figlit_aux}, and Fig.~\ref{figlit_lstm}). In~\cite{tezcan2020bsuv,zhang2019x}, the U-net~\cite{ronneberger2015u} structure has been adopted to benefit from same-level connections between the encoder and decoder. 
Lin et al.~\cite{lin2018foreground} insert two residual connections at different scales between the encoder and decoder. Lim et al.~\cite{lim2017background} add one single skip connection to concatenate the encoder and decoder layers for a particular resolution level. Akilan et al.~\cite{akilan2019video} design the network MvRF-CNN by introducing a large number of skip and residual connections. Similarly, the residual features are added at multiple stages of the network in~\cite{akilan2019sendec,ou2019moving,zeng2018multiscale,zeng2019combining}. The atrous convolutions and attention based modules have also been implemented to improve the network capability for robust change detection~\cite{hu20183d,wang2018foreground1,yang2017deep,liang2019spatio}.

\begin{figure}[t]
\centering
\begin{subfigure}{1\linewidth}
    \centering
    \includegraphics[width=1\textwidth]{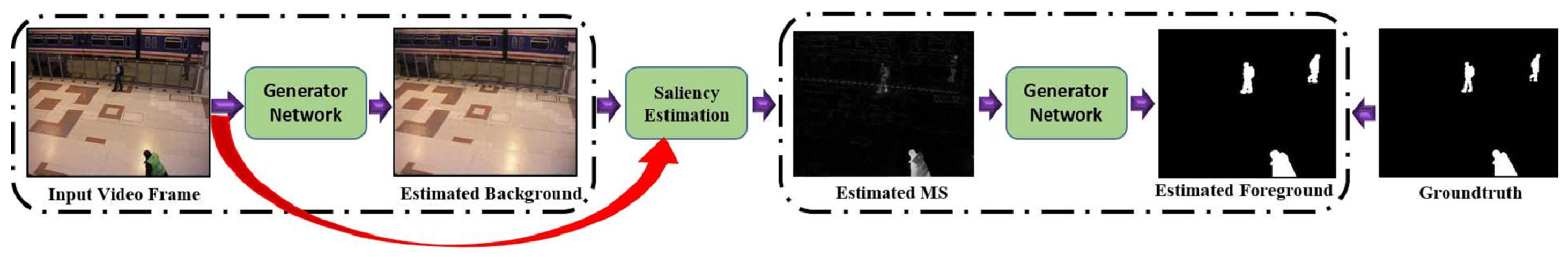}
    \caption{\cite{patil2019fggan}}\label{figlit_unsuperviseda}
\end{subfigure}

\begin{subfigure}{1\linewidth}
    \centering
    \includegraphics[width=1\textwidth ]{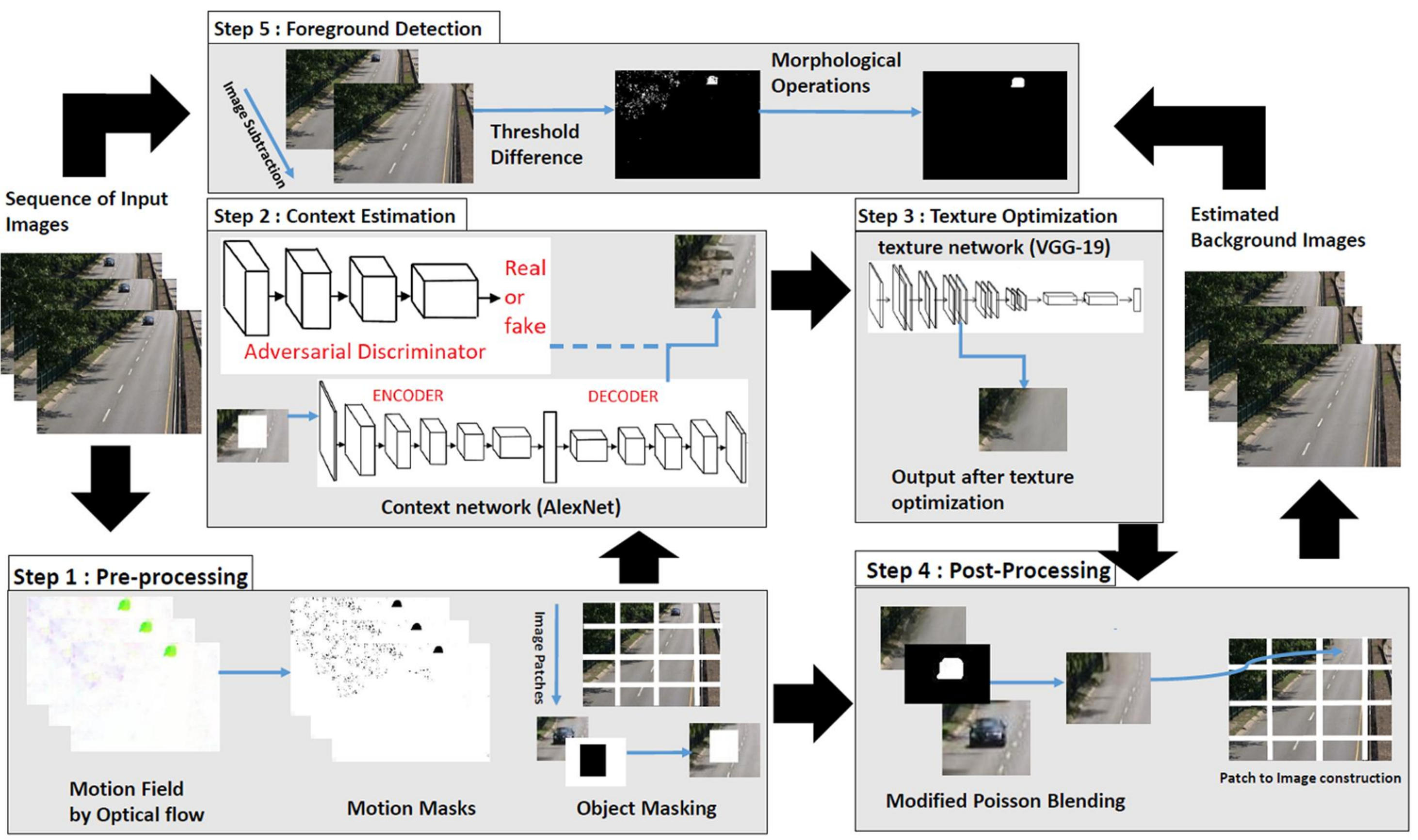}
    \caption{\cite{sultana2019unsupervised}}\label{figlit_unsupervisedb}
\end{subfigure}
	\caption{Representative GAN based semi-supervised methods~\cite{patil2019fggan,sultana2019unsupervised} designed for change detection}
	\label{figlit_unsupervised}
\end{figure}

\textbf{End-to-end vs Ensemble methods.} The change detection methods can also be categorized into end-to-end or ensemble methods. End-to-end implies that the CNN network takes the raw data as input and gives the final response without any support from external modules or blocks. The ensemble methods require a lot of intermediate computations leading to additional complexity. For example, Mandal et al.~\cite{mandal20193dfr} present 3DFR (Fig.~\ref{figlit_manyc}) that only use the previous 50 raw frames to compute the final segmentation map in an end-to-end manner. On the other hand, the Tezcan at al.~\cite{tezcan2020bsuv} aggregates a CNN, a semantic segmentation model, and temporal median to collectively perform change detection. The approaches in~\cite{chen2017pixel,lim2017background,guo2018learning,chen2018mfcnet,varghese2018changenet,choo2018multi,sakkos2018end,wang2018foreground1,lim2018foreground,zeng2018multiscale,yang2017deep,patil2018msfgnet,zhang2019x,yang2019end,zeng2019combining,akilan2019sendec,akilan2019video,han2019aerial,shahbaz2019deep,mondejar2019end,akilan20193d,lim2019learning,jung2019cosine,liang2019spatio,mandal20193dfr,mandal2020motionrec} offer end-to-end solutions. Whereas, the methods in~\cite{braham2016deep,zhao2017joint,wang2017interactive,zeng2018background,liang2018deep,zhao2018background,choo2018learning,akilan2018new,hu20183d,lin2018foreground,nguyen2018change,babaee2018deep,patil2018msednet,bakkay2018bscgan,liao2018multiscale,qiu2019fully,patil2019motion,tao2019universal,minematsu2019,sultana2019unsupervised,zheng2019novel,gracewell2019dynamic,tezcan2020bsuv} present a variety of assembled algorithms.

\subsubsection{Semi-supervised Methods}
With the advancements in generative adversarial networks (GAN) and autoencoders (AE), there has been a rise in the development of semi-supervised methods for change detection. The existing CD methods can be discussed in the following two groups:

\textbf{GAN Based Methods.} Bakkay et al.~\cite{bakkay2018bscgan} were the first to present a GAN based solution for background subtraction. They proposed a deep background subtraction model called BScGAN, using conditional GAN~\cite{isola2017image}. A simple U-net architecture with skip connections is used as a generator. The discriminator is composed of 4 convolutional and downsampling layers. Patil et al.~\cite{patil2019fggan} proposed an unpaired learning based approach FgGAN for background estimation and foreground segmentation. The FgGAN is inspired by the cycle-consistent adversarial networks (CycleGAN) \cite{zhu2017unpaired}. The block diagram for FgGAN is shown in Fig.~\ref{figlit_unsuperviseda}. Zheng et al.~\cite{zheng2019novel} designed a background subtraction method based on parallel vision and Bayesian GANs. Sultana et al.~\cite{sultana2019unsupervised} introduced an unsupervised algorithm by unifying an optical flow based pre-processor, GAN network for arbitrary region inpainting, Poisson blending based post-processor, and threshold based foreground detector. The detailed methodology is depicted in Fig. \ref{figlit_unsupervisedb}. Yu et al.~\cite{yu2020background} combined GAN with domain adaptation for background subtraction in remote sensing videos. 
Ammar et al.~\cite{ammar2020deep} utilize an anomaly discovery framework DeepSphere and GAN to segment and classify moving objects in video sequences.

\textbf{Autoencoders Based Methods.}
Gracewell and John~\cite{gracewell2019dynamic} designed an autoencoder network for background modelling. The network is first initialized with greedy layer wise pretraining approach and then fine tuned using conjugate gradient based back propagation algorithm. Garcia et al.~\cite{garcia2019background,garcia2019foreground} proposed a CD system with a stacked denoising autoencoder extracting the salient features for each patch of several shifted tilings of the video frame. For each patch of the frame, a probabilistic model is learned that are considered in pixel-level classification. The autoencoders have also been used for background estimation~\cite{xie2020autoencoder,jo2019regularized,xia2018stereoscopic} which can be used to further detect motion pixels.\par


\section{Training and Evaluation Frameworks}
\label{framework_compare}
The performance of any supervised method is affected primarily by two factors: 1) the model uncertainty i.e., the uncertainly about model architecture and parameters, 2) the evaluation setting i.e., the strictness in the data-division to validate the generalization strength of the model. The technical analysis of the model designs of different methods is already presented in the previous section. In this section, we focus on the training and evaluation setups adopted by the existing methods. \par

Based on the extensive literature study, we report that the existing supervised CD methods has been evaluated with a variety of different training-testing set selections. The traditional methods~\cite{st2014subsense,jiang2017wesambe,braham2017semantic,mandal2018antic,mandal2018candid} for change detection usually do not require labeled training data.
Thus, there is no need to define train-test splits. However, it is a crucial decision in supervised (deep learning) change detection techniques. Most of the benchmark datasets like CDnet 2014~\cite{wang2014cdnet}, LASIESTA~\cite{cuevas2016labeled}, SBI2015~\cite{maddalena2015towards}, Fish4Knowledge~\cite{kavasidis2014innovative}, GTFD~\cite{li2016weighted}, UCSD~\cite{mahadevan2009spatiotemporal}, and PITS~\cite{li2004statistical} do not define the train-test division. Thus, researchers have used different data division strategies for network training and evaluation. This makes the results claimed in different papers incomparable to each other as well as with previous unsupervised (traditional) approaches. We categorize the evaluation strategies into scene dependent evaluation (SDE), scene independent evaluation (SIE), and cross-dataset evaluation (CDE) settings (refer Fig.~\ref{cdsurvey_fig3}). A comparative analysis of different evaluation schemes is tabulated in Table~\ref{tab_framework1} and Table~\ref{tab_framework2}. 

\subsection{SDE}
\begin{figure}[]
\centering
	\includegraphics[width=0.9\textwidth ]{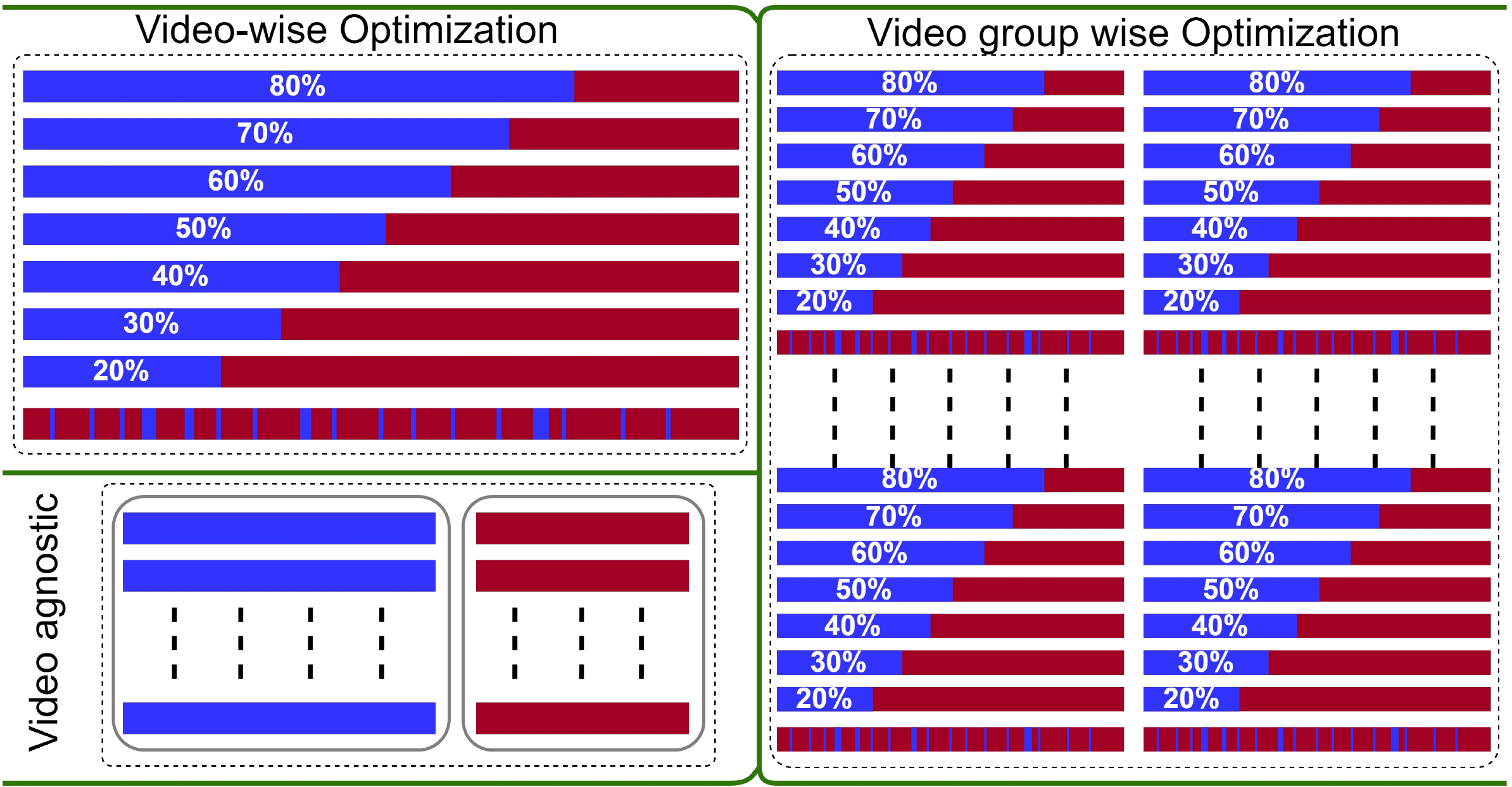}
	\caption{The different training-testing data-division strategies adopted in the literature for deep CD methods. (a) video wise optimization, (b) video-group wise optimization, (c) video agnostic}
	\label{cdsurvey_fig8}
\end{figure}

In SDE setup, usually a certain percentage of frames from a video is put into a training and the rest is used for testing as shown in Fig.~\ref{cdsurvey_fig8}. Most of the existing deep learning methods~\cite{lin2018foreground,lim2018foreground,zeng2018multiscale,lim2017background,nguyen2018change,babaee2018deep,braham2016deep,wang2017interactive,chen2017pixel,yang2017deep,patil2018msfgnet,akilan20193d,bakkay2018bscgan} follow SDE scheme for evaluation. In~\cite{braham2016deep,chen2017pixel,bakkay2018bscgan}, 50\% frames from each video are selected for training. Similarly, 90\% and 70\% frames were used to train CNN models in~\cite{yang2017deep} and~\cite{akilan20193d}, respectively. Few researchers~\cite{lim2018foreground,zeng2018multiscale,nguyen2018change,babaee2018deep,patil2018msednet,wang2017interactive,patil2018msfgnet,patil2019fggan} have selectively or randomly chosen the training frames. In~\cite{babaee2018deep}, 5\% of the frames were randomly chosen for training. Likewise, 50/200 frames are manually selected in~\cite{lim2018foreground,wang2017interactive,patil2019fggan}. Moreover, 150 and 100 training frames are randomly in~\cite{zeng2018multiscale} and~\cite{nguyen2018change}, respectively. In the literature, two types of SDE setups have been adopted for training: video-wise optimization and video-group wise optimization. In video-wise optimization, the network is optimized for \textit{each video separately}. Whereas, in video-group wise optimization, the network is optimized over a group of videos (usually, the complete dataset) and \textit{a single} model is obtained. The difference between these two types of training strategies is depicted in Fig.~\ref{cdsurvey_fig8}(a) and Fig.~\ref{cdsurvey_fig8}(b). A detailed description of the scene dependent settings in the existing methods in tabulated in Table~\ref{tab_framework1} and Table~\ref{tab_framework2}.\par

Since all frames of a video sequence have the same background representation which is learned during training. Such schemes clearly favor the CNN model while testing. Therefore, even though high performance has been claimed in the literature, it is very difficult to evaluate their robustness in unseen videos. The problem is with the adopted experimental framework and not necessarily with model design. The stricter evaluation is possible when completely unseen videos are tested as done in the SIE setup. 

\begin{table*}[]
\footnotesize
    \centering
    \caption{Experimental Setting based comparison of the existing deep learning approaches.V-Opt: Video Optimized, SI: Scene Independence, VFs: Video frames, vids: Videos CD14: CDNet 2014 \textit{Year 2016-2018} }
    \begin{tabular}{c|c c c c c}
    \hline\hline
\multirow{2}{*}{\textbf{Pub-Yr}} &\textbf{Training data} 	&\textbf{Testing data}  &\multirow{2}{*}{\textbf{V-Opt}} &\multirow{2}{*}{\textbf{SI}} &\multirow{2}{*}{\textbf{Dataset}}	\\
&\textbf{selection} &\textbf{selection} & & & \\
\hline\hline
ICSSIP-16\cite{braham2016deep}	&50\% of VFs	&Remaining 50\% of VFs	&Yes	&No &CD14 (23 vids)	\\
\hline
ICME-17\cite{zhao2017joint}	&50\% of VFs	&Remaining 50\% of VFs 	&Yes	&No	&CD14(all)	\\
\hline
\multirow{2}{*}{PRL-17\cite{wang2017interactive}}	&\multirow{2}{*}{Selective 50/200 VFs}	&\multirow{2}{*}{Complete dataset}	&\multirow{2}{*}{Yes}	&\multirow{2}{*}{No}	&CD14(all), \\
& & & & & SBMI2015 (14 vids)\\
\hline
\multirow{2}{*}{TCSVT-17\cite{chen2017pixel}}	&20\%/30\%/40\%/50\% 	&Remaining 80\%/70\%/	&\multirow{2}{*}{Yes}	&\multirow{2}{*}{No}	&CD14 (all),\\
&of VFs & 60\%/50\% of VFs & & & LASIESTA (all)\\
\hline
AVSS-17\cite{lim2017background}	&39 vids	&10 unseen vids	&Yes	&Yes	&CD14 (49 vids)\\
\hline
\multirow{2}{*}{IA-18\cite{zeng2018background}}	&Video-wise, &\multirow{2}{*}{Complete dataset}	&\multirow{2}{*}{Yes} &\multirow{2}{*}{No} &CD14 (all),\\
&Selective 200 frames & & & & SBM-RGBD\\
\hline
\multirow{2}{*}{Arxiv-18 \cite{guo2018learning}}	&Image-pairs  	&Self defined 	&\multirow{2}{*}{No}	&\multirow{2}{*}{Yes}	&CD14, PCD2015,	\\
&selection &test set & & & VL-CMU-CD\\
\hline
\multirow{2}{*}{ICIP-18\cite{chen2018mfcnet}} &Video-wise, &\multirow{2}{*}{Complete dataset}	&\multirow{2}{*}{Yes}	&\multirow{2}{*}{No}	&\multirow{2}{*}{CD14 (all)}\\
&selective 500 frames & & & & \\
\hline
\multirow{2}{*}{ICME-18 \cite{liang2018deep}} &Video-wise, &\multirow{2}{*}{Complete dataset}	&\multirow{2}{*}{Yes}	&\multirow{2}{*}{No}	&\multirow{2}{*}{CD14(all)}	\\
&selective 500 frames & & & & \\
\hline
\multirow{2}{*}{ICME-18 \cite{zhao2018background}} &Video-wise, &\multirow{2}{*}{Complete dataset}	&\multirow{2}{*}{Yes}	&\multirow{2}{*}{No}	&\multirow{2}{*}{CD14 (all)}\\
&selective 1/20/40 frames & & & & \\
\hline
\multirow{2}{*}{ECCVW-18\cite{varghese2018changenet}}	&Image-pairs  	&Self defined 	&\multirow{2}{*}{No}	&\multirow{2}{*}{Yes}	&PCD2015, \\
&selection &test set & & & VL-CMU-CD\\
\hline
\multirow{2}{*}{ICPR-18\cite{choo2018multi}}	&Video-wise, &\multirow{2}{*}{Complete dataset}	&\multirow{2}{*}{Yes}	&\multirow{2}{*}{No}	&\multirow{2}{*}{CD14 (all)}\\
&selective 200 frames & & & &\\
\hline
\multirow{2}{*}{ACCV-18\cite{choo2018learning}}	&Image synthesis, 	&\multirow{2}{*}{Complete dataset}	&\multirow{2}{*}{Yes}	&\multirow{2}{*}{No}	&\multirow{2}{*}{CD14 (49 vids)}\\
&background frame & & & &\\
\hline
MWSCAS-18\cite{akilan2018new}	&70\% of VFs	&Complete dataset	&Yes	&No	&CD14 (7 vids)\\
\hline
IA-18\cite{hu20183d}	&80\% of VFs	&20\% of VFs	&Yes	&No	&CD14 (all)\\
\hline
MTAP-18\cite{sakkos2018end}	&70\% of VFs	&Remaining 30\% of VFs	&No	&No	&CD14 (all), ESI (5 vids)\\
\hline
Sensors-18\cite{wang2018foreground1}	&50\% of VFs	&Remaining 50\% of VFs	&Yes	&No	&CD14 (all)\\
\hline
IGRSL-18 \cite{wang2018foreground2}	&50\% of VFs	&Remaining 50\% of VFs	&Yes	&No	&CD14 (5 vids)\\
\hline
ICIP-18\cite{lin2018foreground}	&20 vids	&6 unseen vids	&No	&Yes &CD14 (26 vids)\\
\hline
PRL-18\cite{lim2018foreground}	&Selective 50/200 VFs	&Complete dataset	&Yes	&No	&CD14 (all)\\
\hline
\multirow{2}{*}{IGRSL-18\cite{zeng2018multiscale}}	&Randomly selected 	&\multirow{2}{*}{Complete dataset}	&\multirow{2}{*}{Yes}	&\multirow{2}{*}{No}	&\multirow{2}{*}{CD14 (5 vids)}\\
&150 VFs & & & &\\
\hline
\multirow{2}{*}{TCSVT-18\cite{nguyen2018change}}	&Randomly selected 	&\multirow{2}{*}{Complete dataset}	&\multirow{2}{*}{Yes}	&\multirow{2}{*}{No}	&CD14 (6 vids), BMC (5 vids), 	\\
&100 VFs & & & &Wallflower (4 vids)\\
\hline
\multirow{2}{*}{PR-18\cite{babaee2018deep}}	&Randomly selected &Remaining 95\% 	&\multirow{2}{*}{Yes}	&\multirow{2}{*}{No}	&CD14 (all)	\\
&5\% of VFs &of VFs & & &Wallflower (7 vids)\\
\hline
\multirow{2}{*}{SMC-18\cite{patil2018msednet}}	&Randomly selected 	&\multirow{2}{*}{Remaining VFs}	&\multirow{2}{*}{No}	&\multirow{2}{*}{No}	&CD14 (35 vids),\\
&5,500 frames of VFs & & & &Wallflower (6 vids)\\
\hline
\multirow{2}{*}{TITS-18\cite{yang2017deep}}	&\multirow{2}{*}{90\% of VFs}	&Remaining 10\%	&\multirow{2}{*}{Yes}	&\multirow{2}{*}{No}	&\multirow{2}{*}{CD14 (6 vids)}	\\
& &of VFs & & &\\
\hline
\multirow{2}{*}{TITS-18\cite{patil2018msfgnet}}	&Randomly selected 	&\multirow{2}{*}{Remaining VFs}	&\multirow{2}{*}{Yes}	&\multirow{2}{*}{No}	&CD14 (all), PTIS(all),	\\
&9,800 frames & & & &LASIESTA (all)\\
\hline
\multirow{2}{*}{ICIP-18\cite{bakkay2018bscgan}}	&\multirow{2}{*}{50\% of VFs}	&Remaining 50\% 	&\multirow{2}{*}{Yes}	&\multirow{2}{*}{No}	&CD14 (49 vids),\\
& &of VFs & & &BMC (10 vids)\\
\hline
PRCM-18\cite{liao2018multiscale}	&5/10/20 VFs	&Remaining VFs	&Yes	&No	&CD14 (49 vids)\\
\hline\hline

    \end{tabular}
    \label{tab_framework1}
\end{table*}

\subsection{SIE}
\label{sec_SIE}
In SIE setup, the train and test sets consist of completely different videos. It ensures evaluation over videos with unseen background.
In Fig.~\ref{cdsurvey_fig8}(c), a sample video-agnostic setting is shown for scene independent evaluation. To ensure complete scene independence, Mandal et al.~\cite{mandal20193dfr} adopted a leave-one-video-out (LOVO) strategy. They leave out 1 video from each category of CDnet 2014 for testing and use the remaining videos to train the model. For example, if a category such as ‘bad weather’ contains 4 videos, then 3 videos are selected for training and the remaining one video is used for testing. Such an experimental setup makes the model design much more challenging as compared to the SDE setup. Tezcan et al.~\cite{tezcan2020bsuv} first collect all the videos across the categories into a single pool. Thereafter, they create 18 different subdivisions to ensure that the none of the training videos overlap with testing videos. This way, all the videos were at some point tested in completely unseen manner. Similarly, Mondezar et al.~\cite{mondejar2019end} conducted scene-wise 3-fold cross-validation in order to evaluate the capability of the proposed architecture to extrapolate to unseen scenes. The authors in~\cite{lin2018foreground} have also attempted to conduct SIE with only a selected set of videos from CDnet 2014.

\subsection{Cross-dataset}
Another kind of SIE can be ensured by using the models trained over a particular dataset and test on a completely different public dataset. Patil and Murala~\cite{patil2018msfgnet} used the model trained over the CDnet 2014+LASIESTA to test the videos in PTIS. Similarly, in~\cite{patil2020end}, the trained model on CDnet 2014 thermal videos are evaluated over GTFD. Other types of video selection schemes could be explored in future along with solving the network design challenges for the same. 

\subsection{SIE Vs SDE}
\label{sec_sievssde}
\begin{figure}[]
\centering
	\includegraphics[width=0.7\textwidth ]{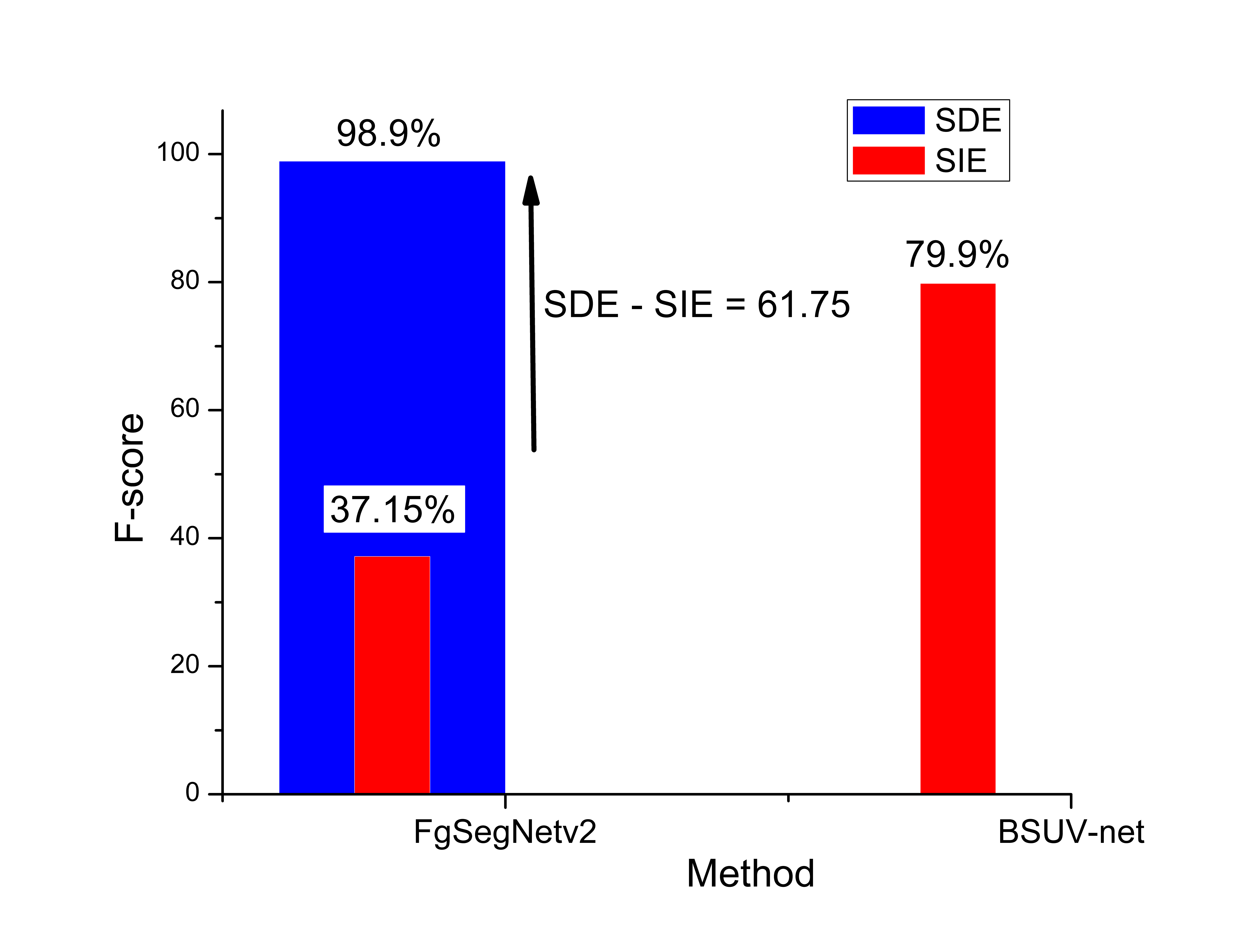}
	\caption{The FgSegNetv2 model is evaluated in both SDE and SIE setup on the CDnet 2014 dataset. It obtains 98.9\% and 37.15\% F-score in SDE and SIE setup, respectively. It clearly highlights the challenges in designing a CNN based solution for robust performance in completely unseen videos. It also shows that the SIE/CDE setup is much more strict over the SDE setup. The SIE and CDE setting also ensures generalized model over the over-fitted models obtained in SDE evaluation. The recent methods in~\cite{mandal20193dfr,tezcan2020bsuv,mondejar2019end} have attempted to present such generalized solution by validating the models in SIE setting.}
	\label{cdsurvey_graph8}
\end{figure}

As discussed in Section~\ref{sec_SIE}, the SDE setup leads to model optimization only for the same set of videos used in training. This is due to the fact that some frames from the test videos are used for training. Therefore, it is essential to evaluate the model over unseen or scene independent videos. This also makes the process of model design much more challenging in order to ensure robust performance even in real-world scenarios. Such SIE scheme ensures proper evaluation of the designed model as compared to SDE. Therefore, the recent works have opted for more challenging SIE setups model evaluation. Tezcun et al.~\cite{tezcan2020bsuv} have shown that some of the models~\cite{lim2018foreground,lim2019learning} claiming very high performance in SDE setup performed poorly when evaluated with in SIE setup. More specifically, the original paper for FgSegNetv2 model~\cite{lim2018foreground} reports an overall F-score of 98.9 over CDnet 2014. However, when trained and evaluated in the SIE setup as in~\cite{tezcan2020bsuv}, it could only obtain 37.15 F-score. Therefore, the SDE based results are unreliable to validate the actual robustness of the deep learning models. More recent benchmark datasets for other video-based applications~\cite{milan2016mot16,sultani2018real} already ensure such scene independency in their evaluation schemes. Based on all these observations, our proposition is to give more importance to SIE or CDE over SDE for change detection model evaluation. We demonstrate the performance degradation of FgSegNetv2 in SIE setup in comparison to SDE in Fig.~\ref{cdsurvey_graph8}.

\begin{table*}[]
\footnotesize
    \centering
    \caption{Experimental Setting based comparison of the existing deep learning approaches. V-Opt: Video Optimized, SI: Scene Independence, VFs: Video frames, vids: Videos CD14: CDNet 2014 \textit{Year 2019-2020} }
    \begin{tabular}{c|c c c c c}
    \hline\hline
\multirow{2}{*}{\textbf{Pub-Yr}} &\textbf{Training data} 	&\textbf{Testing data}  &\multirow{2}{*}{\textbf{V-Opt}} &\multirow{2}{*}{\textbf{SI}} &\multirow{2}{*}{\textbf{Dataset}}	\\
&\textbf{selection} &\textbf{selection} & & & \\
\hline\hline
\multirow{2}{*}{IA-19\cite{qiu2019fully}}	&Video-wise, &\multirow{2}{*}{Complete dataset}	&\multirow{2}{*}{Yes}	&\multirow{2}{*}{No}	&\multirow{2}{*}{CD14 (all)}\\
&selective frames & & & & \\
\hline

\multirow{2}{*}{IA-19\cite{zhang2019x}}	&Video-wise, 
	&\multirow{2}{*}{Complete dataset}	&\multirow{2}{*}{Yes}	&\multirow{2}{*}{No}	&\multirow{2}{*}{CD14 (all)}	\\
&selective 50/200 frames & & & & \\
\hline

\multirow{2}{*}{IA-19\cite{yang2019end}}	&\multirow{2}{*}{Video-wise, 70\% of VFs}	&\multirow{2}{*}{30\% of VFs} 	&\multirow{2}{*}{Yes}	&\multirow{2}{*}{No}	&CD14 (all),\\
& & & & & LASIESTA\\
\hline

\multirow{2}{*}{IA-19\cite{ou2019moving}}	&Video-wise, &\multirow{2}{*}{Complete dataset}	&\multirow{2}{*}{Yes}	&\multirow{2}{*}{No}	&CD14 (all),\\
&random 20\% selected frames & & & & PTIS\\
\hline

\multirow{2}{*}{JEI-19\cite{zeng2019combining}}	&Video-wise, 	&\multirow{2}{*}{Complete dataset}	&\multirow{2}{*}{Yes}	&\multirow{2}{*}{No}	&\multirow{2}{*}{CD14 (all)} \\
&randomly selected frames & & & & \\
\hline

TITS-19\cite{akilan2019sendec}	&Video-wise, 70\% of VFs	&30\% of VFs 	&Yes	&No &CD14 (16 vids) \\
\hline

TVT-19\cite{akilan2019video}	&Video-wise, 70\% of VFs	&30\% of VFs 	&Yes	&No &CD14 (37 vids) \\
\hline

\multirow{2}{*}{Neuro-19\cite{han2019aerial}}	&\multirow{2}{*}{NA}	&\multirow{2}{*}{NA}	&\multirow{2}{*}{NA}	&\multirow{2}{*}{NA}	&CD14 (10 vids), AICD 2012,  \\
& & & & & aerial dataset\\
\hline

\multirow{2}{*}{ISIE-19\cite{shahbaz2019deep}} &Video-wise,	&\multirow{2}{*}{Complete dataset}	&\multirow{2}{*}{Yes}	&\multirow{2}{*}{No} &\multirow{2}{*}{CD14 (all)}\\
&selective 50/200 frames & & & & \\
\hline

WACV-19\cite{patil2019fggan}	&200 VFs	&Remaining VFs	&Yes	&No	&CD14 (35 vids)\\
\hline

\multirow{2}{*}{BMVC-19\cite{mondejar2019end}}	&Scene-wise 3-fold 	&Scene-wise 3-fold 	&\multirow{2}{*}{No}	&\multirow{2}{*}{Yes} &\multirow{2}{*}{CD14 (49 vids)}\\
&cross-validation &cross-validation & & &\\
\hline

TITS-19\cite{akilan20193d}	&70\% of VFs	&Remaining 30\% of VFs	&Yes	&No	&CD14 (16 vids) \\
\hline

\multirow{2}{*}{PAA-19\cite{lim2019learning}} &25/200 VFs and &Remaining VFs,
 &\multirow{2}{*}{Yes}	&\multirow{2}{*}{No}	&CD14 (all), UCSD (18 vids) \\
&20\%/50\% of VFs &Remaining 80\%/50\% of VFs & & &  SBI2015 (14 vids)\\
\hline

\multirow{2}{*}{AVSS-19\cite{patil2019motion}}	&\multirow{2}{*}{1100 VFs}	&\multirow{2}{*}{Remaining VFs}	&\multirow{2}{*}{No}	&\multirow{2}{*}{No} &, CD14 (10 vids),	\\
& & & & & Fish4Knowledge (7 vids)\\
\hline

\multirow{2}{*}{IA-19\cite{tao2019universal}}	&\multirow{2}{*}{80\% of VFs}	&\multirow{2}{*}{Remaining 20\% of VFs}	&\multirow{2}{*}{Yes}	&\multirow{2}{*}{No}	&CD14 
(all),\\
& & & & & LASIESTA\\
\hline

AVSS-19\cite{minematsu2019}	&N\% of VFs	&N\% of VFs	&Yes	&No	&CD14 (28 vids)	\\
\hline

AVSS-19\cite{jung2019cosine} &160 VFs	&40 VFs	&Yes	&No	&CD14 (all)\\
\hline

\multirow{2}{*}{Sensors-19\cite{liang2019spatio}}	&Randomly selected 	&Remaining 95\% of 
VFs	&\multirow{2}{*}{No}	&\multirow{2}{*}{No}	&CD14(all), PETS2009,\\
&5\% of VFs & & & &  Wallflower (6 vids)\\
\hline

ISPL-19\cite{mandal20193dfr}	&38 vids	&10 unseen vids	&No	&Yes &CD14 (48 vids)\\
\hline

\multirow{2}{*}{MVA-19\cite{sultana2019unsupervised}} &Unsupervised approach,	&\multirow{2}{*}{Complete dataset}	&\multirow{2}{*}{No}	&\multirow{2}{*}{Yes}	&\multirow{2}{*}{CD14 (35 vids)}	\\
&labeled data not required & & & &\\
\hline

\multirow{2}{*}{Neuro-19\cite{zheng2019novel}}	&Randomly selected 	&\multirow{2}{*}{Complete dataset}	&\multirow{2}{*}{Yes}	&\multirow{2}{*}{No}	&CD14 (all), UCSD (18 vids),\\
&100 VFs & & & &SBMI2015 (14 vids)\\
\hline

\multirow{2}{*}{MTAP-19 \cite{gracewell2019dynamic}} &Unsupervised approach, &\multirow{2}{*}{Complete dataset}	&\multirow{2}{*}{No}	&\multirow{2}{*}{Yes} &CD14 (49 vids),	\\
&labeled data not required & & & & AVSS, CAVIAR\\
\hline

\multirow{2}{*}{WACV-20\cite{tezcan2020bsuv}}	&18 split-combinations 	&18 split-combinations &\multirow{2}{*}{No}	&\multirow{2}{*}{Yes} &\multirow{2}{*}{CD14 (all)}\\
&of all the vids &of all the vids & & &\\
\hline
WACV-20\cite{mandal2020motionrec}	&16 vids	&3 unseen vids	&No	&Yes &CD14 (19 vids)	\\
\hline\hline

    \end{tabular}
    \label{tab_framework2}
\end{table*}

\subsection{Discussion}
\label{sec_discussion_framework}
As discussed in Section~\ref{sec_sievssde}, the SIE/CDE setup is much more challenging than the popularly used SDE setup. However, even for SDE evaluation, we notice a clear inconsistency among the schemes adopted in the literature~\cite{lin2018foreground,lim2018foreground,zeng2018multiscale,lim2017background,nguyen2018change,babaee2018deep,braham2016deep,wang2017interactive,chen2017pixel,yang2017deep,patil2018msfgnet,akilan20193d,bakkay2018bscgan} to report the results over CD datasets. For traditional unsupervised methods, the experimental setup is clearly defined which makes these results comparable without any bias issues. However, in deep learning methods, it is essential to maintain nonoverlapping (SIE/CDE) between the training and testing sets. Moreover, it is highly desired that the training and testing data should not be similar as in SDE. These factors have not been carefully considered in the existing literature. For example, although temporal information forms the basis of change detection, the highest results claimed in~\cite{lim2018foreground,lim2019learning} do not even consider it in their respective CNN models. They use a carefully selected set of frames (50/200 frames) from each video to train the model and achieve more than 98\% F-score over CDnet 2014 dataset. It can be argued that such evaluation is clearly overfitted as the training and testing data is almost the same and no spatiotemporal feature is learned by the networks to identify the change. Such models are not suitable to handle the challenges in change detection in unseen videos as demonstrated through Fig.~\ref{cdsurvey_graph8}. Moreover, even for SDE setup, researchers have apparently adopted different SDE schemes for training and evaluation. These documented results cannot be fairly compared. Therefore, benchmarking the performances of different CNN, GAN models in a standard evaluation setup (SDE/SIE/CDE) is an important scope in change detection research.

\section{Datasets and Evaluation Metrics}
The current research needs across all the computer vision applications are motivated by the success of deep learning algorithms. The success of the deep learning algorithms depends on the availability of sufficient labeled training data that include as many variations of the populations and environments as possible. Higher the diversity in the captured video scenarios for training, the more robustly one can estimate the model parameters. \par

\begin{figure*}[]
\centering
	\includegraphics[width=0.9\textwidth ]{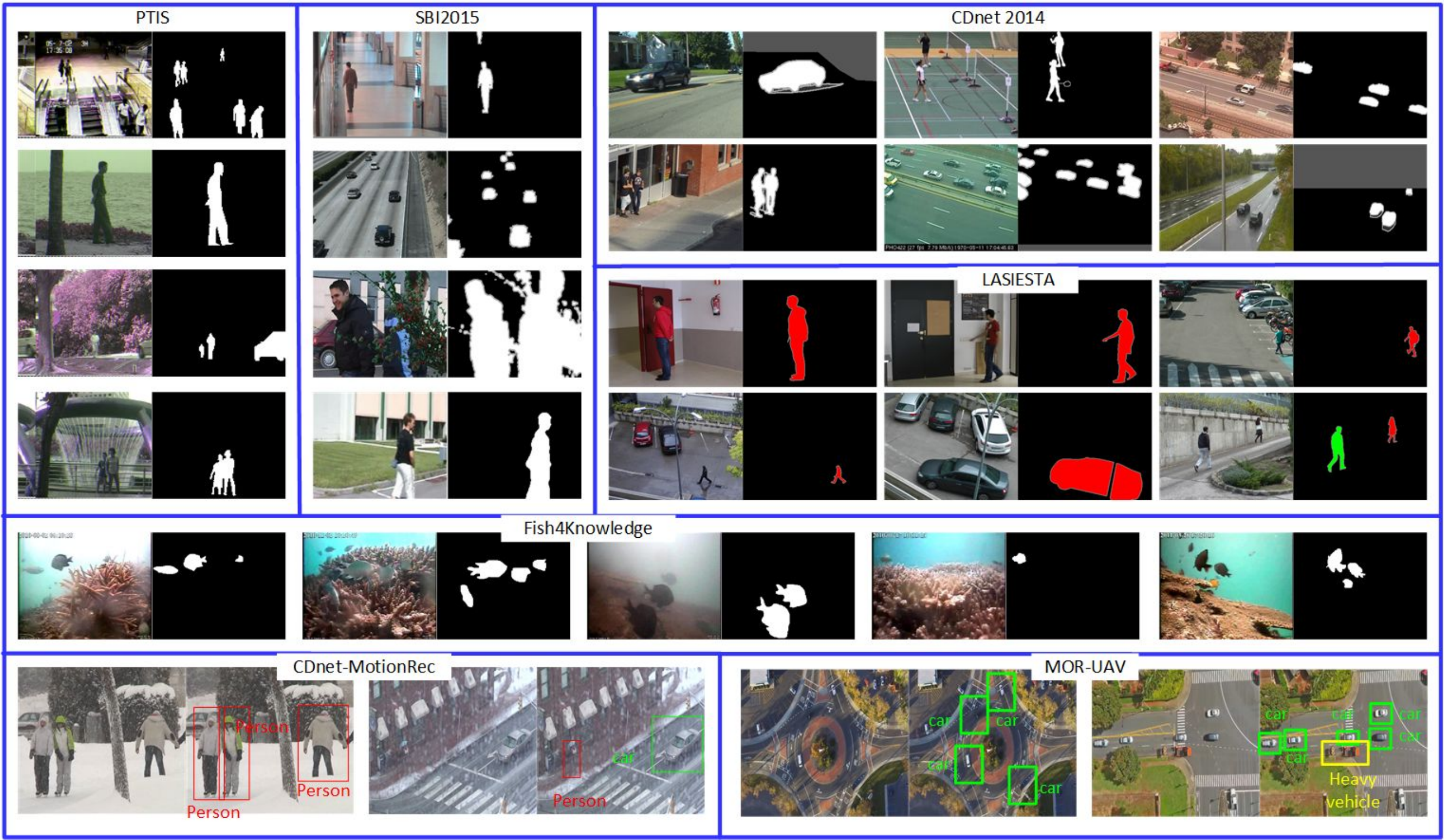}
	\caption{Sample frames from the change detection datasets: CDnet 2014, PTIS, SBI2015, Fish4Knowledge, LASIESTA, CDnet-MotionRec, and MOR-UAV}
	\label{cdsurvey_fig10}
\end{figure*}

In this section, we primarily discuss about the publicly available change detection datasets that has been used for evaluating the deep learning methods in the literature. A more detailed review of all the existing datasets for CD is presented in~\cite{kalsotra2019comprehensive}. We list out the most popular datasets used in the deep learning methods for evaluation in Table~\ref{dataset_tab}. It provides the main reference, number of videos, number of frames, number of labels, and the access details. These are the most relevant information needed while working with deep learning methods. Sample video frames from some of these datasets are shown in Fig.~\ref{cdsurvey_fig10}. Below, we discuss these change detection datasets based on the type of captured video frames i.e. conventional, thermal, underwater, and aerial.\par

\textbf{Conventional video datasets:} The eight conventional CD datasets: CDnet 2014~\cite{wang2014cdnet}, LASIESTA~\cite{cuevas2016labeled}, PTIS/I2R~\cite{li2004statistical}, SBI2015~\cite{maddalena2015towards}, UCSD~\cite{mahadevan2009spatiotemporal}, SegTrack-v2~\cite{FliICCV2013}, DAVIS-2016~\cite{Perazzi2016}, FBMS~\cite{ochs2013segmentation}, have been most prominently used for evaluation in the literature. Among them, CDnet 2014~\cite{wang2014cdnet} offers the most diverse set of scenarios along with the highest number of labeled frames (~90,000). The video sequences are quite large with an average video length of about 3005 frames. Whereas, SegTrack-v2~\cite{FliICCV2013}, DAVIS-2016~\cite{Perazzi2016}, FBMS~\cite{ochs2013segmentation} provide small videos with average video lengths 70, 69, 234, respectively. The LASIESTA~\cite{cuevas2016labeled} provides the object category labels along with the foreground pixel labels. However, the dataset contains only three type of objects. The PTIS~\cite{li2004statistical}, SBI2015~\cite{maddalena2015towards} and UCSD~\cite{mahadevan2009spatiotemporal} provides further challenging set of videos for change detection.\par

\textbf{Underwater datasets:} For moving object detection in underwater videos, the Fish4Knowledge~\cite{kavasidis2014innovative} and UnderwaterCD~\cite{underwaterchangedetection} datasets have been introduced. The Fish4Knowledge~\cite{kavasidis2014innovative} contains 17 videos each containing about 2777 frames. However, the labels for only a combined 869 frames are available for all the videos. The UnderwaterCD~\cite{underwaterchangedetection} has only 5 videos. Each video has 100 frames and 100 labels for evaluation.\par 

\textbf{Thermal video datasets:} One of the categories in CDnet 2014~\cite{wang2014cdnet} consist of 5 thermal videos. Overall, there are 21,100 frames and 6,137 labels. Recently, more datasets are created for thermal video-based change detection. The GTFD~\cite{li2016weighted} dataset introduced 25 new videos captured with infra-red devices. A total of 1,067 frames are available to analysis. The TU-VDN~\cite{singha2019tu,singha2019salient} captured longer thermal video sequences (19 videos, 43,247 frames) from challenging scenarios for change detection.\par

\textbf{Aerial video datasets:} More recently, Mandal et al.~\cite{m2020moruav} introduced a new large-scale aerial video dataset named MOR-UAV. The dataset consists of aerial videos captured from UAVs in numerous scenarios. The moving objects are labeled with axis-aligned bounding boxes along with corresponding object class. The bounding box labelling requires less computational resources than producing pixel-level estimates. Moreover, the associated foreground labels are useful in many applications.

\subsection{Evaluation Metrics}
Researchers have used a variety of metrics to measure the effectiveness of the change detection methods. However, identifying the best metric to accurately measure the potency of a change detection method is a non-trivial decision. The designed CD method must be able to report minimal false positive (FP) and false negative (FN). Similarly, the count for true positive (TP) and true negative (TN) should ideally be on higher side. The precision favors methods with low FP whereas, the recall favors methods with low FN. Thus, the F-score presents a balanced metric which demands lower values of both FP and FN. Similarly, the Percentage of Wrong Classifications (PWC) computes the ratio between all the wrong classifications (FP+FN) by the total predictions (TP+TN+FN+FP). The mean absolute error (MAE), specificity, false positive rate (FPR), false negative rate (FNR), also offers useful insights about the ability of the CD methods. We define the performance metrics: precision, recall, specificity (Sp), FPR, FNR, F-score (FS), and PWC for pixel-wise change detection in Eq.~\ref{survey_eq1} - Eq.~\ref{survey_eq7}.

\noindent\begin{minipage}{.5\linewidth}
\begin{equation}
  Precision = \frac{TP}{TP+FP}
  \label{survey_eq1}
 \end{equation}
\end{minipage}%
\begin{minipage}{.5\linewidth}
\begin{equation}
  Recall = \frac{TP}{TP+FN}
  \label{survey_eq2}
\end{equation}
\end{minipage}

\noindent\begin{minipage}{.5\linewidth}
\begin{equation}
  Sp = \frac{TN}{TN+FP}
  \label{survey_eq3}
 \end{equation}
\end{minipage}%
\begin{minipage}{.5\linewidth}
\begin{equation}
  FPR = \frac{FP}{FP+TN}
  \label{survey_eq4}
\end{equation}
\end{minipage}

\noindent\begin{minipage}{.5\linewidth}
\begin{equation}
  FNR = \frac{FN}{TN+FP}
  \label{survey_eq5}
 \end{equation}
\end{minipage}%
\begin{minipage}{.5\linewidth}
\begin{equation}
  FS = 2\times\frac{Pr\times Re}{Pr+Re}
  \label{survey_eq6}
\end{equation}
\end{minipage}
\begin{equation}
  PWC = 100\times \frac{FN+FP}{FN+FP+TP+TN}
  \label{survey_eq7}
\end{equation}


\begin{table*}[]
\footnotesize
    \centering
    \caption{An overview of the commonly used change detection datasets in the literature}
    \resizebox{\columnwidth}{!}{
    \begin{tabular}{c c c c c c}
    \hline\hline
\textbf{Database}	&\textbf{\#Frames}	&\textbf{\#Videos}	&\textbf{\#Labels}	&\textbf{Frame Type}	&\textbf{Access}\\
\hline\hline
CDnet 2014 \cite{wang2014cdnet}	&159,279	&53	&~90,000	&Conventional	&changedetection.net \\

LASIESTA \cite{cuevas2016labeled}	&18,425	&48	&18,425	&Conventional	&https://www.gti.ssr.upm.es/data/lasiesta\_database.html \\

PTIS/I2R \cite{li2004statistical}	&18,449	&9	&180	&Conventional	&http://vis-www.cs.umass.edu/$\sim$ narayana/castanza/I2Rdataset/ \\

SBI2015 \cite{maddalena2015towards}	&5,029	&14	&5,029	&Conventional	&https://github.com/lim-anggun/FgSegNet \\

UCSD \cite{mahadevan2009spatiotemporal}	&885	&18	&885	&Conventional	&https://github.com/lim-anggun/FgSegNet \\

SegTrack-v2 \cite{FliICCV2013}	&976	&14	&976	&Conventional	&https://web.engr.oregonstate.edu/$\sim$ lif/SegTrack2/dataset.html \\

DAVIS-2016 \cite{Perazzi2016}	&3,455	&50	&3,455	&Conventional	&https://davischallenge.org/davis2016/code.html \\

FBMS \cite{ochs2013segmentation}	&13806	&59	&720	&Conventional	&https://lmb.informatik.uni-freiburg.de/resources/datasets/ \\

CDnet-MotionRec~\cite{Mandal_2020_WACV}	&24,923	&19	&24,923	&Conventional	&https://github.com/murari023/MotionRec\\

Fish4Knowledge \cite{kavasidis2014innovative}	&47,200	&17	&869	&Underwater	&http://www.perceivelab.com/dataset/Fish\%20Detection\%20and\%20Tracking \\

UnderwaterCD \cite{underwaterchangedetection}	&500	&5	&500	&Underwater	&http://underwaterchangedetection.eu/index.html \\

GTFD \cite{li2016weighted}	&1067	&25	&1067	&Thermal	&Request based access \\

TU-VDN \cite{singha2019tu}	&43,247	&19	&4,313	&Thermal	&http://www.mkbhowmik.in/tuvdn.aspx\\

MOR-UAV~\cite{m2020moruav}	&10,948	&30	&10,948	&Aerial	&https://visionintelligence.github.io/Datasets.html\\
\hline
\end{tabular}
}
\label{dataset_tab}
\end{table*}

\section{Research Needs and Future Directions}
\label{research_needs}
\subsection{Research Needs}
\subsubsection{Bench-marking the deep learning results}
As discussed in Section \ref{sec_discussion_framework}, there is a problem of non-comparability among the recent deep learning-based change detection methods. Different CNN networks have been trained over different training-testing data-divisions. This is due to unavailability of earmarked divisions for train and test set in the large-scale datasets such as CDnet 2014, LASIESTA, and SBI2015. Some recent methods~\cite{tezcan2020bsuv,mondejar2019end} have evaluated the model with cross-data validation by testing over completely unseen videos. However, the ad-hoc nature of the cross-validation data divisions raise the question of data division bias in order to boost the performance of the proposed models. Similarly, the authors in~\cite{mandal20193dfr} evaluate over selected unseen videos to show robustness of the models. There is a need for benchmark data division to ensure uniform comparative analysis of the presented deep learning models.

\subsubsection{Change detection datasets}
As discussed in the previous section, the labeled CD datasets are available for natural, thermal and underwater scenes. All these videos are captured with fixed cameras. Few datasets~\cite{wang2014cdnet,cuevas2016labeled} contain limited set of videos captured with moving cameras. However, the camera motion is very miniscule and under controlled settings. There is a need for labeled CD datasets captured with unconstrained camera movements. An example of unconstrained scenario is the videos captured with the dash camera inside a moving vehicle. Such a dataset could potentially pave the way for a new set of challenges for deep learning algorithms.\par

The aerial view-based CD is another important research area that requires specific attention. There is a need for labeled CD dataset for aerial videos captured from UAV mounted cameras. Some recent datasets~\cite{zhu2018vision,zhu2019visdrone,mueller2016benchmark} have presented annotations for object tracking and video object detection. Similarly, the CD labels could be produced for evaluation of change detection models in aerial scenes. 

\subsubsection{Robustness and Real-time Challenges}
To attain robust performance in real-world scenarios, it is important to validate the models in SIE and CDE settings. 
Furthermore, the current reported model speeds (GPU or CPU) are not comparable due difference in the hardware configurations. There is a need to benchmark the process of measuring the speed of the deep learning models~\cite{murari-tits}.

\subsubsection{Bounding-box based motion detection}
Most of the CD datasets provide pixel-wise labels which require algorithms to estimate pixel-wise binary classifications. The object class (car, person, truck, etc.) for the foreground pixels are not identified. Separate object detectors~\cite{lin2017focal} could be used for obtaining the relevant object classes as post-processing. However, this will lead to poor latency, inefficient resource utilizations and memory overhead. The recent work in~\cite{Mandal_2020_WACV} advertise the utility of a single stage localization and classification algorithm for moving object recognition (MOR) in end-to-end manner. This work takes advantage of the state-of-the-art object detectors~\cite{linfocalpami,lin2017focal} to achieves robust MOR performance. The authors annotate the CDnet 2014 for training the bounding-box based MOR algorithm. Further investigation in this direction should be done to benefit from the rapid advancement in deep learning-based object detectors.

\subsection{Future Directions}
\subsubsection{CD in Aerial Videos}
Aerial vision-based data is being generated in abundance in both consumer and industrial market space. The aerial video data captured with UAV-mounted cameras facilitate numerous applications such as aerial surveillance, search and rescue, event recognition, urban and rural scene understanding. Thus, change detection in aerial videos is an important domain of research. Some of the most challenging scenarios is to detect object motion when the camera mounted UAV is also moving either is same direction or in other direction of the moving objects~\cite{m2020moruav,chen2020high}. Similarly, the variable speed of objects and the UAVs lead to even more challenging cases. The future works should focus to address these challenges by both collecting labeled data and designing novel algorithms for aerial videos.

\subsubsection{CD in Moving Cameras}
There has been some important research in the field of CD from moving cameras~\cite{yazdi2018new,jiang2021moving}. However, limited labeled datasets and lack of deep learning-based solutions make it a very important research direction to explore. Most of the existing CD work primarily focused on the fixed camera-based videos. The future work requires more attention towards both data collection and algorithm development in the direction of CD in moving cameras. The autonomous vehicles and aerial video analytics are two prominent applications for such solutions.   

\subsubsection{Vision-based Forecasting in Videos}
The deep learning-based methods have bridged the gap between research and real-world applications for vision-based detection, recognition and segmentation. However, beyond these well studied problems, vision-based forecasting will likely be one of the next big research topics. The ability to make prediction the future in videos will test the ability of the deep learning algorithms. The existing CD algorithms can be extended to predict the location of the moving objects in future frames. This will have utility in traffic monitoring~\cite{datondji2016survey,ahmed2019query}, anomaly detection~\cite{roy2020detection,wan2020intelligent}, trajectory prediction~\cite{cai2021pedestrian,song2020pedestrian}, etc.
\par

One of the applications of such algorithms can be autonomous driving~\cite{feng2020deep,rasouli2019autonomous}, where precognition of certain events, object positions, through vision-based understanding of the temporal video data can resolve some of the safety issues. The research could be dedicated towards future object localization and anticipation of trajectories. The existing CD datasets can be re-purposed to serve as benchmark to evaluate models to predict the future location of objects after few frames (10/20/30/40… frames).

\subsubsection{Self-supervised Learning in Videos}
The change detection methods can provide the free labels to develop self-supervised systems for several applications including object detection~\cite{fang2020move}, video order prediction~\cite{xu2019self}, and Video Representation Learning~\cite{han2019video}. The future works include creating self-supervised algorithms by leveraging the existing CD methods. Such algorithms could also be extended to autonomous driving, intelligent surveillance, and anomaly detection.

\section{Conclusion}
\label{sec_conclusion}
This paper presents an empirical review of the deep learning frameworks for change detection. Particularly, the existing CD methods are analyzed in terms of model design and evaluation frameworks. The variety of existing deep learning architectures are examined for their effectiveness to change detection. The CD methods are divided into broad categories and respective subcategories to provides a comprehensive review. The deep CD networks are analyzed based on the architecture design, network input, pretrained modules, finetuning, and auxiliary blocks. We notice that the supervised methods obtain superior performance than the unsupervised methods. Several methods combine multiple heavy CNN modules to obtain higher performance which is matched by much smaller networks. It is observed that the pretrained models have limited effect on the performance and carefully designed lightweight networks also obtain good results on the benchmark datasets. Thus, the research challenge is to design the CNN network with minimal set of operations for motion analysis. Another key open challenge is to design resource efficient deep networks that can run at high speed over the CPU devices for real-time deployment.\par

The important categorization of the evaluation frameworks is presented as SIE, SDE, and CDE setups. Although, most of the existing works use the SDE to present their results, the SIE and CDE setups enforce a more stringent setting to test the generalization capability of the models. Thus, we encourage the researchers to report the SIE results for a robust evaluation of their deep CD methods. Similarly, other important challenges in robust CD, the current research needs, and future directions are discussed. We also described the related video datasets and the evaluation metrics of pixel-wise mask techniques. We believe this review will benefit the researchers in this field and provide useful insights into this important research topic. We hope to encourage more future work to develop in this direction.

\bibliographystyle{IEEEtran}
\bibliography{IEEEfull}

\begin{thebibliography}{100}
\providecommand{\url}[1]{#1}
\csname url@samestyle\endcsname
\providecommand{\newblock}{\relax}
\providecommand{\bibinfo}[2]{#2}
\providecommand{\BIBentrySTDinterwordspacing}{\spaceskip=0pt\relax}
\providecommand{\BIBentryALTinterwordstretchfactor}{4}
\providecommand{\BIBentryALTinterwordspacing}{\spaceskip=\fontdimen2\font plus
\BIBentryALTinterwordstretchfactor\fontdimen3\font minus
  \fontdimen4\font\relax}
\providecommand{\BIBforeignlanguage}[2]{{%
\expandafter\ifx\csname l@#1\endcsname\relax
\typeout{** WARNING: IEEEtran.bst: No hyphenation pattern has been}%
\typeout{** loaded for the language `#1'. Using the pattern for}%
\typeout{** the default language instead.}%
\else
\language=\csname l@#1\endcsname
\fi
#2}}
\providecommand{\BIBdecl}{\relax}
\BIBdecl

\bibitem{li2017general}
X.~Li, B.~Zhao, and X.~Lu, ``A general framework for edited video and raw video
  summarization,'' \emph{IEEE Transactions on Image Processing}, vol.~26,
  no.~8, pp. 3652--3664, 2017.

\bibitem{nie2019collision}
Y.~Nie, Z.~Li, Z.~Zhang, Q.~Zhang, T.~Ma, and H.~Sun, ``Collision-free video
  synopsis incorporating object speed and size changes,'' \emph{IEEE
  Transactions on Image Processing}, vol.~29, pp. 1465--1478, 2019.

\bibitem{marotirao2019challenges}
K.~Marotirao~Biradar, A.~Gupta, M.~Mandal, and S.~Kumar~Vipparthi, ``Challenges
  in time-stamp aware anomaly detection in traffic videos,'' in
  \emph{Proceedings of the IEEE Conference on Computer Vision and Pattern
  Recognition Workshops}, 2019, pp. 13--20.

\bibitem{dong2019quadruplet}
X.~Dong, J.~Shen, D.~Wu, K.~Guo, X.~Jin, and F.~Porikli, ``Quadruplet network
  with one-shot learning for fast visual object tracking,'' \emph{IEEE
  Transactions on Image Processing}, vol.~28, no.~7, pp. 3516--3527, 2019.

\bibitem{xu2019learning}
T.~Xu, Z.-H. Feng, X.-J. Wu, and J.~Kittler, ``Learning adaptive discriminative
  correlation filters via temporal consistency preserving spatial feature
  selection for robust visual object tracking,'' \emph{IEEE Transactions on
  Image Processing}, vol.~28, no.~11, pp. 5596--5609, 2019.

\bibitem{zhang2020rapnet}
P.~Zhang, W.~Liu, Y.~Lei, H.~Wang, and H.~Lu, ``Rapnet: Residual atrous pyramid
  network for importance-aware street scene parsing,'' \emph{IEEE Transactions
  on Image Processing}, vol.~29, pp. 5010--5021, 2020.

\bibitem{motiian2016online}
S.~Motiian, F.~Siyahjani, R.~Almohsen, and G.~Doretto, ``Online human
  interaction detection and recognition with multiple cameras,'' \emph{IEEE
  transactions on Circuits and Systems for Video Technology}, vol.~27, no.~3,
  pp. 649--663, 2016.

\bibitem{li2020quantifying}
X.~Li, M.~Chen, and Q.~Wang, ``Quantifying and detecting collective motion in
  crowd scenes,'' \emph{IEEE Transactions on Image Processing}, vol.~29, pp.
  5571--5583, 2020.

\bibitem{zhang2018real}
B.~Zhang, L.~Wang, Z.~Wang, Y.~Qiao, and H.~Wang, ``Real-time action
  recognition with deeply transferred motion vector cnns,'' \emph{IEEE
  Transactions on Image Processing}, vol.~27, no.~5, pp. 2326--2339, 2018.

\bibitem{yi2016pedestrian}
S.~Yi, H.~Li, and X.~Wang, ``Pedestrian behavior modeling from stationary
  crowds with applications to intelligent surveillance,'' \emph{IEEE
  transactions on image processing}, vol.~25, no.~9, pp. 4354--4368, 2016.

\bibitem{fu2019foreground}
Z.~Fu, Y.~Chen, H.~Yong, R.~Jiang, L.~Zhang, and X.-S. Hua, ``Foreground gating
  and background refining network for surveillance object detection,''
  \emph{IEEE Transactions on Image Processing}, vol.~28, no.~12, pp.
  6077--6090, 2019.

\bibitem{Mandal_2020_WACV}
M.~Mandal, L.~K. Kumar, M.~S. Saran, and S.~K. vipparthi, ``Motionrec: A
  unified deep framework for moving object recognition,'' in \emph{The IEEE
  Winter Conference on Applications of Computer Vision (WACV)}, March 2020.

\bibitem{rasouli2019autonomous}
A.~Rasouli and J.~K. Tsotsos, ``Autonomous vehicles that interact with
  pedestrians: A survey of theory and practice,'' \emph{IEEE Transactions on
  Intelligent Transportation Systems}, vol.~21, no.~3, pp. 900--918, 2019.

\bibitem{kim2020anomaly}
H.~Kim, J.~Park, K.~Min, and K.~Huh, ``Anomaly monitoring framework in lane
  detection with a generative adversarial network,'' \emph{IEEE Transactions on
  Intelligent Transportation Systems}, 2020.

\bibitem{roy2020detection}
D.~Roy, T.~Ishizaka, C.~K. Mohan, and A.~Fukuda, ``Detection of collision-prone
  vehicle behavior at intersections using siamese interaction lstm,''
  \emph{IEEE Transactions on Intelligent Transportation Systems}, 2020.

\bibitem{tian2019online}
W.~Tian, M.~Lauer, and L.~Chen, ``Online multi-object tracking using joint
  domain information in traffic scenarios,'' \emph{IEEE Transactions on
  Intelligent Transportation Systems}, vol.~21, no.~1, pp. 374--384, 2019.

\bibitem{wan2020intelligent}
S.~Wan, X.~Xu, T.~Wang, and Z.~Gu, ``An intelligent video analysis method for
  abnormal event detection in intelligent transportation systems,'' \emph{IEEE
  Transactions on Intelligent Transportation Systems}, 2020.

\bibitem{ahmed2019query}
S.~A. Ahmed, D.~P. Dogra, S.~Kar, R.~Patnaik, S.-C. Lee, H.~Choi, G.~P. Nam,
  and I.-J. Kim, ``Query-based video synopsis for intelligent traffic
  monitoring applications,'' \emph{IEEE Transactions on Intelligent
  Transportation Systems}, 2019.

\bibitem{patil2018msfgnet}
P.~W. Patil and S.~Murala, ``Msfgnet: A novel compact end-to-end deep network
  for moving object detection,'' \emph{IEEE Transactions on Intelligent
  Transportation Systems}, vol.~20, no.~11, pp. 4066--4077, 2018.

\bibitem{akilan20193d}
T.~Akilan, Q.~J. Wu, A.~Safaei, J.~Huo, and Y.~Yang, ``A 3d cnn-lstm-based
  image-to-image foreground segmentation,'' \emph{IEEE Transactions on
  Intelligent Transportation Systems}, 2019.

\bibitem{mandal20203dcd}
M.~{Mandal}, V.~{Dhar}, A.~{Mishra}, S.~K. {Vipparthi}, and
  M.~{Abdel-Mottaleb}, ``3dcd: Scene independent end-to-end spatiotemporal
  feature learning framework for change detection in unseen videos,''
  \emph{IEEE Transactions on Image Processing}, vol.~30, pp. 546--558, 2021.

\bibitem{murari-tits}
M.~Mandal and S.~K. Vipparthi, ``Scene independency matters: An empirical study
  of scene dependent and scene independent evaluation for cnn-based change
  detection,'' \emph{IEEE Transactions on Intelligent Transportation Systems},
  pp. 1--14, 2020.

\bibitem{roy2017real}
S.~M. Roy and A.~Ghosh, ``Real-time adaptive histogram min-max bucket (hmmb)
  model for background subtraction,'' \emph{IEEE Transactions on Circuits and
  Systems for Video Technology}, vol.~28, no.~7, pp. 1513--1525, 2017.

\bibitem{zhao2016real}
L.~Zhao, Z.~He, W.~Cao, and D.~Zhao, ``Real-time moving object segmentation and
  classification from hevc compressed surveillance video,'' \emph{IEEE
  Transactions on Circuits and Systems for Video Technology}, vol.~28, no.~6,
  pp. 1346--1357, 2016.

\bibitem{zivkovic2004improved}
Z.~Zivkovic, ``Improved adaptive gaussian mixture model for background
  subtraction,'' in \emph{Proceedings of the 17th International Conference on
  Pattern Recognition, 2004. ICPR 2004.}, vol.~2.\hskip 1em plus 0.5em minus
  0.4em\relax IEEE, 2004, pp. 28--31.

\bibitem{stauffer1999adaptive}
C.~Stauffer and W.~E.~L. Grimson, ``Adaptive background mixture models for
  real-time tracking,'' in \emph{Proceedings. 1999 IEEE Computer Society
  Conference on Computer Vision and Pattern Recognition (Cat. No PR00149)},
  vol.~2.\hskip 1em plus 0.5em minus 0.4em\relax IEEE, 1999, pp. 246--252.

\bibitem{varadarajan2013spatial}
S.~Varadarajan, P.~Miller, and H.~Zhou, ``Spatial mixture of gaussians for
  dynamic background modelling,'' in \emph{2013 10th IEEE International
  Conference on Advanced Video and Signal Based Surveillance}.\hskip 1em plus
  0.5em minus 0.4em\relax IEEE, 2013, pp. 63--68.

\bibitem{jiang2017wesambe}
S.~Jiang and X.~Lu, ``Wesambe: A weight-sample-based method for background
  subtraction,'' \emph{IEEE Transactions on Circuits and Systems for Video
  Technology}, vol.~28, no.~9, pp. 2105--2115, 2017.

\bibitem{wang2014fast}
B.~Wang and P.~Dudek, ``A fast self-tuning background subtraction algorithm,''
  in \emph{Proceedings of the IEEE Conference on Computer Vision and Pattern
  Recognition Workshops}, 2014, pp. 395--398.

\bibitem{ramirez2016auto}
G.~Ram{\'\i}rez-Alonso and M.~I. Chac{\'o}n-Murgu{\'\i}a, ``Auto-adaptive
  parallel som architecture with a modular analysis for dynamic object
  segmentation in videos,'' \emph{Neurocomputing}, vol. 175, pp. 990--1000,
  2016.

\bibitem{mandal2018antic}
M.~Mandal, M.~Chaudhary, S.~K. Vipparthi, S.~Murala, A.~B. Gonde, and S.~K.
  Nagar, ``Antic: Antithetic isomeric cluster patterns for medical image
  retrieval and change detection,'' \emph{IET Computer Vision}, vol.~13, no.~1,
  pp. 31--43, 2018.

\bibitem{st2014subsense}
P.-L. St-Charles, G.-A. Bilodeau, and R.~Bergevin, ``Subsense: A universal
  change detection method with local adaptive sensitivity,'' \emph{IEEE
  Transactions on Image Processing}, vol.~24, no.~1, pp. 359--373, 2014.

\bibitem{hofmann2012background}
M.~Hofmann, P.~Tiefenbacher, and G.~Rigoll, ``Background segmentation with
  feedback: The pixel-based adaptive segmenter,'' in \emph{2012 IEEE computer
  society conference on computer vision and pattern recognition
  workshops}.\hskip 1em plus 0.5em minus 0.4em\relax IEEE, 2012, pp. 38--43.

\bibitem{wang2014cdnet}
Y.~Wang, P.-M. Jodoin, F.~Porikli, J.~Konrad, Y.~Benezeth, and P.~Ishwar,
  ``Cdnet 2014: An expanded change detection benchmark dataset,'' in
  \emph{Proceedings of the IEEE conference on computer vision and pattern
  recognition workshops}, 2014, pp. 387--394.

\bibitem{li2004statistical}
L.~Li, W.~Huang, I.~Y.-H. Gu, and Q.~Tian, ``Statistical modeling of complex
  backgrounds for foreground object detection,'' \emph{IEEE Transactions on
  Image Processing}, vol.~13, no.~11, pp. 1459--1472, 2004.

\bibitem{bouwmans2014traditional}
T.~Bouwmans, ``Traditional and recent approaches in background modeling for
  foreground detection: An overview,'' \emph{Computer science review}, vol.~11,
  pp. 31--66, 2014.

\bibitem{sobral2014comprehensive}
A.~Sobral and A.~Vacavant, ``A comprehensive review of background subtraction
  algorithms evaluated with synthetic and real videos,'' \emph{Computer Vision
  and Image Understanding}, vol. 122, pp. 4--21, 2014.

\bibitem{datondji2016survey}
S.~R.~E. Datondji, Y.~Dupuis, P.~Subirats, and P.~Vasseur, ``A survey of
  vision-based traffic monitoring of road intersections,'' \emph{IEEE
  transactions on intelligent transportation systems}, vol.~17, no.~10, pp.
  2681--2698, 2016.

\bibitem{sommer2016survey}
L.~W. Sommer, M.~Teutsch, T.~Schuchert, and J.~Beyerer, ``A survey on moving
  object detection for wide area motion imagery,'' in \emph{2016 IEEE Winter
  Conference on Applications of Computer Vision (WACV)}.\hskip 1em plus 0.5em
  minus 0.4em\relax IEEE, 2016, pp. 1--9.

\bibitem{cuevas2016detection}
C.~Cuevas, R.~Mart{\'\i}nez, and N.~Garc{\'\i}a, ``Detection of stationary
  foreground objects: A survey,'' \emph{Computer Vision and Image
  Understanding}, vol. 152, pp. 41--57, 2016.

\bibitem{prasad2017video}
D.~K. Prasad, D.~Rajan, L.~Rachmawati, E.~Rajabally, and C.~Quek, ``Video
  processing from electro-optical sensors for object detection and tracking in
  a maritime environment: a survey,'' \emph{IEEE Transactions on Intelligent
  Transportation Systems}, vol.~18, no.~8, pp. 1993--2016, 2017.

\bibitem{goyal2018review}
K.~Goyal and J.~Singhai, ``Review of background subtraction methods using
  gaussian mixture model for video surveillance systems,'' \emph{Artificial
  Intelligence Review}, vol.~50, no.~2, pp. 241--259, 2018.

\bibitem{bouwmans2017scene}
T.~Bouwmans, L.~Maddalena, and A.~Petrosino, ``Scene background initialization:
  A taxonomy,'' \emph{Pattern Recognition Letters}, vol.~96, pp. 3--11, 2017.

\bibitem{prasad2018object}
D.~K. Prasad, C.~K. Prasath, D.~Rajan, L.~Rachmawati, E.~Rajabally, and
  C.~Quek, ``Object detection in a maritime environment: Performance evaluation
  of background subtraction methods,'' \emph{IEEE Transactions on Intelligent
  Transportation Systems}, vol.~20, no.~5, pp. 1787--1802, 2018.

\bibitem{maddalena2018background}
L.~Maddalena and A.~Petrosino, ``Background subtraction for moving object
  detection in rgbd data: A survey,'' \emph{Journal of Imaging}, vol.~4, no.~5,
  p.~71, 2018.

\bibitem{yazdi2018new}
M.~Yazdi and T.~Bouwmans, ``New trends on moving object detection in video
  images captured by a moving camera: A survey,'' \emph{Computer Science
  Review}, vol.~28, pp. 157--177, 2018.

\bibitem{kalsotra2019comprehensive}
R.~Kalsotra and S.~Arora, ``A comprehensive survey of video datasets for
  background subtraction,'' \emph{IEEE Access}, vol.~7, pp. 59\,143--59\,171,
  2019.

\bibitem{bouwmans2019deep}
T.~Bouwmans, S.~Javed, M.~Sultana, and S.~K. Jung, ``Deep neural network
  concepts for background subtraction: A systematic review and comparative
  evaluation,'' \emph{Neural Networks}, vol. 117, pp. 8--66, 2019.

\bibitem{chapel2020moving}
M.-N. Chapel and T.~Bouwmans, ``Moving objects detection with a moving camera:
  A comprehensive review,'' \emph{arXiv preprint arXiv:2001.05238}, 2020.

\bibitem{espinosa2020detection}
J.~E. Espinosa, S.~A. Velast{\'\i}n, and J.~W. Branch, ``Detection of
  motorcycles in urban traffic using video analysis: A review,'' \emph{IEEE
  Transactions on Intelligent Transportation Systems}, 2020.

\bibitem{wang2007consensus}
H.~Wang and D.~Suter, ``A consensus-based method for tracking: Modelling
  background scenario and foreground appearance,'' \emph{Pattern recognition},
  vol.~40, no.~3, pp. 1091--1105, 2007.

\bibitem{heikkila2006texture}
M.~Heikkila and M.~Pietikainen, ``A texture-based method for modeling the
  background and detecting moving objects,'' \emph{IEEE transactions on pattern
  analysis and machine intelligence}, vol.~28, no.~4, pp. 657--662, 2006.

\bibitem{el2008fuzzy}
F.~El~Baf, T.~Bouwmans, and B.~Vachon, ``Fuzzy integral for moving object
  detection,'' in \emph{2008 IEEE International Conference on Fuzzy Systems
  (IEEE World Congress on Computational Intelligence)}.\hskip 1em plus 0.5em
  minus 0.4em\relax IEEE, 2008, pp. 1729--1736.

\bibitem{maddalena2008self}
L.~Maddalena and A.~Petrosino, ``A self-organizing approach to background
  subtraction for visual surveillance applications,'' \emph{IEEE Transactions
  on Image Processing}, vol.~17, no.~7, pp. 1168--1177, 2008.

\bibitem{sheikh2009background}
Y.~Sheikh, O.~Javed, and T.~Kanade, ``Background subtraction for freely moving
  cameras,'' in \emph{2009 IEEE 12th International Conference on Computer
  Vision}.\hskip 1em plus 0.5em minus 0.4em\relax IEEE, 2009, pp. 1219--1225.

\bibitem{liao2010modeling}
S.~Liao, G.~Zhao, V.~Kellokumpu, M.~Pietik{\"a}inen, and S.~Z. Li, ``Modeling
  pixel process with scale invariant local patterns for background subtraction
  in complex scenes,'' in \emph{2010 IEEE Computer Society Conference on
  Computer Vision and Pattern Recognition}.\hskip 1em plus 0.5em minus
  0.4em\relax IEEE, 2010, pp. 1301--1306.

\bibitem{barnich2010vibe}
O.~Barnich and M.~Van~Droogenbroeck, ``Vibe: A universal background subtraction
  algorithm for video sequences,'' \emph{IEEE Transactions on Image
  processing}, vol.~20, no.~6, pp. 1709--1724, 2010.

\bibitem{he2011online}
J.~He, L.~Balzano, and J.~Lui, ``Online robust subspace tracking from partial
  information,'' \emph{arXiv preprint arXiv:1109.3827}, 2011.

\bibitem{wang2014static}
R.~Wang, F.~Bunyak, G.~Seetharaman, and K.~Palaniappan, ``Static and moving
  object detection using flux tensor with split gaussian models,'' in
  \emph{Proceedings of the IEEE Conference on Computer Vision and Pattern
  Recognition Workshops}, 2014, pp. 414--418.

\bibitem{javed2014or}
S.~Javed, S.~H. Oh, A.~Sobral, T.~Bouwmans, and S.~K. Jung, ``Or-pca with mrf
  for robust foreground detection in highly dynamic backgrounds,'' in
  \emph{Asian Conference on Computer Vision}.\hskip 1em plus 0.5em minus
  0.4em\relax Springer, 2014, pp. 284--299.

\bibitem{st2016universal}
P.-L. St-Charles, G.-A. Bilodeau, and R.~Bergevin, ``Universal background
  subtraction using word consensus models,'' \emph{IEEE Transactions on Image
  Processing}, vol.~25, no.~10, pp. 4768--4781, 2016.

\bibitem{bianco2017combination}
S.~Bianco, G.~Ciocca, and R.~Schettini, ``Combination of video change detection
  algorithms by genetic programming,'' \emph{IEEE Transactions on Evolutionary
  Computation}, vol.~21, no.~6, pp. 914--928, 2017.

\bibitem{braham2017semantic}
M.~Braham, S.~Pi{\'e}rard, and M.~Van~Droogenbroeck, ``Semantic background
  subtraction,'' in \emph{2017 IEEE International Conference on Image
  Processing (ICIP)}.\hskip 1em plus 0.5em minus 0.4em\relax IEEE, 2017, pp.
  4552--4556.

\bibitem{mandal2018candid}
M.~Mandal, P.~Saxena, S.~K. Vipparthi, and S.~Murala, ``Candid: Robust change
  dynamics and deterministic update policy for dynamic background
  subtraction,'' in \emph{2018 24th International Conference on Pattern
  Recognition (ICPR)}.\hskip 1em plus 0.5em minus 0.4em\relax IEEE, 2018, pp.
  2468--2473.

\bibitem{javed2018moving}
S.~Javed, A.~Mahmood, S.~Al-Maadeed, T.~Bouwmans, and S.~K. Jung, ``Moving
  object detection in complex scene using spatiotemporal structured-sparse
  rpca,'' \emph{IEEE Transactions on Image Processing}, vol.~28, no.~2, pp.
  1007--1022, 2018.

\bibitem{braham2016deep}
M.~Braham and M.~Van~Droogenbroeck, ``Deep background subtraction with
  scene-specific convolutional neural networks,'' in \emph{2016 international
  conference on systems, signals and image processing (IWSSIP)}.\hskip 1em plus
  0.5em minus 0.4em\relax IEEE, 2016, pp. 1--4.

\bibitem{wang2017interactive}
Y.~Wang, Z.~Luo, and P.-M. Jodoin, ``Interactive deep learning method for
  segmenting moving objects,'' \emph{Pattern Recognition Letters}, vol.~96, pp.
  66--75, 2017.

\bibitem{babaee2018deep}
M.~Babaee, D.~T. Dinh, and G.~Rigoll, ``A deep convolutional neural network for
  video sequence background subtraction,'' \emph{Pattern Recognition}, vol.~76,
  pp. 635--649, 2018.

\bibitem{chen2017pixel}
Y.~Chen, J.~Wang, B.~Zhu, M.~Tang, and H.~Lu, ``Pixel-wise deep sequence
  learning for moving object detection,'' \emph{IEEE Transactions on Circuits
  and Systems for Video Technology}, 2017.

\bibitem{bakkay2018bscgan}
M.~C. Bakkay, H.~A. Rashwan, H.~Salmane, L.~Khoudour, D.~Puigtt, and
  Y.~Ruichek, ``Bscgan: deep background subtraction with conditional generative
  adversarial networks,'' in \emph{2018 25th IEEE International Conference on
  Image Processing (ICIP)}.\hskip 1em plus 0.5em minus 0.4em\relax IEEE, 2018,
  pp. 4018--4022.

\bibitem{lim2018foreground}
L.~A. Lim and H.~Y. Keles, ``Foreground segmentation using convolutional neural
  networks for multiscale feature encoding,'' \emph{Pattern Recognition
  Letters}, vol. 112, pp. 256--262, 2018.

\bibitem{mandal20193dfr}
M.~Mandal, V.~Dhar, A.~Mishra, and S.~K. Vipparthi, ``3dfr: A swift 3d feature
  reductionist framework for scene independent change detection,'' \emph{IEEE
  Signal Processing Letters}, vol.~26, no.~12, pp. 1882--1886, 2019.

\bibitem{patil2019fggan}
P.~Patil and S.~Murala, ``Fggan: A cascaded unpaired learning for background
  estimation and foreground segmentation,'' in \emph{2019 IEEE Winter
  Conference on Applications of Computer Vision (WACV)}.\hskip 1em plus 0.5em
  minus 0.4em\relax IEEE, 2019, pp. 1770--1778.

\bibitem{tezcan2020bsuv}
O.~Tezcan, P.~Ishwar, and J.~Konrad, ``Bsuv-net: a fully-convolutional neural
  network for background subtraction of unseen videos,'' in \emph{The IEEE
  Winter Conference on Applications of Computer Vision}, 2020, pp. 2774--2783.

\bibitem{krizhevsky2012imagenet}
A.~Krizhevsky, I.~Sutskever, and G.~E. Hinton, ``Imagenet classification with
  deep convolutional neural networks,'' in \emph{Advances in neural information
  processing systems}, 2012, pp. 1097--1105.

\bibitem{szegedy2015going}
C.~Szegedy, W.~Liu, Y.~Jia, P.~Sermanet, S.~Reed, D.~Anguelov, D.~Erhan,
  V.~Vanhoucke, and A.~Rabinovich, ``Going deeper with convolutions,'' in
  \emph{Proceedings of the IEEE conference on computer vision and pattern
  recognition}, 2015, pp. 1--9.

\bibitem{simonyan2014very}
K.~Simonyan and A.~Zisserman, ``Very deep convolutional networks for
  large-scale image recognition,'' \emph{arXiv preprint arXiv:1409.1556}, 2014.

\bibitem{maddalena2015towards}
L.~Maddalena and A.~Petrosino, ``Towards benchmarking scene background
  initialization,'' in \emph{International conference on image analysis and
  processing}.\hskip 1em plus 0.5em minus 0.4em\relax Springer, 2015, pp.
  469--476.

\bibitem{cuevas2016labeled}
C.~Cuevas, E.~M. Y{\'a}{\~n}ez, and N.~Garc{\'\i}a, ``Labeled dataset for
  integral evaluation of moving object detection algorithms: Lasiesta,''
  \emph{Computer Vision and Image Understanding}, vol. 152, pp. 103--117, 2016.

\bibitem{yang2017deep}
L.~Yang, J.~Li, Y.~Luo, Y.~Zhao, H.~Cheng, and J.~Li, ``Deep background
  modeling using fully convolutional network,'' \emph{IEEE Transactions on
  Intelligent Transportation Systems}, vol.~19, no.~1, pp. 254--262, 2017.

\bibitem{choo2018learning}
S.~Choo, W.~Seo, D.-j. Jeong, and N.~I. Cho, ``Learning background subtraction
  by video synthesis and multi-scale recurrent networks,'' in \emph{Asian
  Conference on Computer Vision}.\hskip 1em plus 0.5em minus 0.4em\relax
  Springer, 2018, pp. 357--372.

\bibitem{choo2018multi}
S.~{Choo}, W.~{Seo}, D.~{Jeong}, and N.~I. {Cho}, ``Multi-scale recurrent
  encoder-decoder network for dense temporal classification,'' in \emph{2018
  24th International Conference on Pattern Recognition (ICPR)}, 2018, pp.
  103--108.

\bibitem{zeng2018background}
D.~Zeng and M.~Zhu, ``Background subtraction using multiscale fully
  convolutional network,'' \emph{IEEE Access}, vol.~6, pp. 16\,010--16\,021,
  2018.

\bibitem{lim2019learning}
L.~A. Lim and H.~Y. Keles, ``Learning multi-scale features for foreground
  segmentation,'' \emph{Pattern Analysis and Applications}, pp. 1--12, 2019.

\bibitem{minematsu2019}
T.~Minematsu, A.~Shimada, and R.-i. Taniguchi, ``Simple background subtraction
  constraint for weakly supervised background subtraction network,'' in
  \emph{2019 16th IEEE International Conference on Advanced Video and Signal
  Based Surveillance (AVSS)}.\hskip 1em plus 0.5em minus 0.4em\relax IEEE, Sep
  2019, p. 1–8.

\bibitem{mondejar2019end}
V.~Mond{\'e}jar-Guerra, J.~Rouco, J.~Novo, and M.~Ortega, ``An end-to-end deep
  learning approach for simultaneous background modeling and subtraction,'' in
  \emph{British Machine Vision Conference (BMVC), Cardiff}, 2019.

\bibitem{akilan2019sendec}
T.~Akilan and Q.~J. Wu, ``sendec: An improved image to image cnn for foreground
  localization,'' \emph{IEEE Transactions on Intelligent Transportation
  Systems}, 2019.

\bibitem{st2014improving}
P.-L. St-Charles and G.-A. Bilodeau, ``Improving background subtraction using
  local binary similarity patterns,'' in \emph{IEEE Winter Conference on
  Applications of Computer Vision}.\hskip 1em plus 0.5em minus 0.4em\relax
  IEEE, 2014, pp. 509--515.

\bibitem{lin2018foreground}
C.~Lin, B.~Yan, and W.~Tan, ``Foreground detection in surveillance video with
  fully convolutional semantic network,'' in \emph{2018 25th IEEE International
  Conference on Image Processing (ICIP)}.\hskip 1em plus 0.5em minus
  0.4em\relax IEEE, 2018, pp. 4118--4122.

\bibitem{zeng2018multiscale}
D.~Zeng and M.~Zhu, ``Multiscale fully convolutional network for foreground
  object detection in infrared videos,'' \emph{IEEE Geoscience and Remote
  Sensing Letters}, vol.~15, no.~4, pp. 617--621, 2018.

\bibitem{lim2017background}
K.~Lim, W.-D. Jang, and C.-S. Kim, ``Background subtraction using
  encoder-decoder structured convolutional neural network,'' in \emph{2017 14th
  IEEE International Conference on Advanced Video and Signal Based Surveillance
  (AVSS)}.\hskip 1em plus 0.5em minus 0.4em\relax IEEE, 2017, pp. 1--6.

\bibitem{nguyen2018change}
T.~P. Nguyen, C.~C. Pham, S.~V.-U. Ha, and J.~W. Jeon, ``Change detection by
  training a triplet network for motion feature extraction,'' \emph{IEEE
  Transactions on Circuits and Systems for Video Technology}, vol.~29, no.~2,
  pp. 433--446, 2018.

\bibitem{romero2017background}
J.~D. Romero, M.~J. Lado, and A.~J. Mendez, ``A background modeling and
  foreground detection algorithm using scaling coefficients defined with a
  color model called lightness-red-green-blue,'' \emph{IEEE Transactions on
  Image Processing}, vol.~27, no.~3, pp. 1243--1258, 2017.

\bibitem{sajid2017universal}
H.~Sajid and S.-C.~S. Cheung, ``Universal multimode background subtraction,''
  \emph{IEEE Transactions on Image Processing}, vol.~26, no.~7, pp. 3249--3260,
  2017.

\bibitem{roy2017adaptive}
K.~Roy, J.~Kim, M.~T.~B. Iqbal, F.~Makhmudkhujaev, B.~Ryu, and O.~Chae, ``An
  adaptive fusion scheme of color and edge features for background
  subtraction,'' in \emph{2017 14th IEEE International Conference on Advanced
  Video and Signal Based Surveillance (AVSS)}.\hskip 1em plus 0.5em minus
  0.4em\relax IEEE, 2017, pp. 1--6.

\bibitem{rivera2013background}
A.~R. Rivera, M.~Murshed, J.~Kim, and O.~Chae, ``Background modeling through
  statistical edge-segment distributions,'' \emph{IEEE Transactions on Circuits
  and Systems for Video Technology}, vol.~23, no.~8, pp. 1375--1387, 2013.

\bibitem{chen2017spatiotemporal}
M.~Chen, X.~Wei, Q.~Yang, Q.~Li, G.~Wang, and M.-H. Yang, ``Spatiotemporal gmm
  for background subtraction with superpixel hierarchy,'' \emph{IEEE
  transactions on pattern analysis and machine intelligence}, vol.~40, no.~6,
  pp. 1518--1525, 2017.

\bibitem{chen2019effective}
Y.-Q. Chen, Z.-L. Sun, and K.-M. Lam, ``An effective subsuperpixel-based
  approach for background subtraction,'' \emph{IEEE Transactions on Industrial
  Electronics}, vol.~67, no.~1, pp. 601--609, 2019.

\bibitem{javed2015background}
S.~Javed, S.~Ho~Oh, A.~Sobral, T.~Bouwmans, and S.~Ki~Jung, ``Background
  subtraction via superpixel-based online matrix decomposition with structured
  foreground constraints,'' in \emph{Proceedings of the IEEE International
  Conference on Computer Vision Workshops}, 2015, pp. 90--98.

\bibitem{haines2013background}
T.~S. Haines and T.~Xiang, ``Background subtraction with dirichletprocess
  mixture models,'' \emph{IEEE transactions on pattern analysis and machine
  intelligence}, vol.~36, no.~4, pp. 670--683, 2013.

\bibitem{lee2005effective}
D.-S. Lee, ``Effective gaussian mixture learning for video background
  subtraction,'' \emph{IEEE transactions on pattern analysis and machine
  intelligence}, vol.~27, no.~5, pp. 827--832, 2005.

\bibitem{lin2010regularized}
H.-H. Lin, J.-H. Chuang, and T.-L. Liu, ``Regularized background adaptation: a
  novel learning rate control scheme for gaussian mixture modeling,''
  \emph{IEEE Transactions on Image Processing}, vol.~20, no.~3, pp. 822--836,
  2010.

\bibitem{lanza2010accurate}
A.~Lanza, F.~Tombari, and L.~Di~Stefano, ``Accurate and efficient background
  subtraction by monotonic second-degree polynomial fitting,'' in \emph{2010
  7th IEEE International Conference on Advanced Video and Signal Based
  Surveillance}.\hskip 1em plus 0.5em minus 0.4em\relax IEEE, 2010, pp.
  376--383.

\bibitem{maddalena2007self}
L.~Maddalena and A.~Petrosino, ``A self-organizing approach to detection of
  moving patterns for real-time applications,'' in \emph{International
  Symposium on Brain, Vision, and Artificial Intelligence}.\hskip 1em plus
  0.5em minus 0.4em\relax Springer, 2007, pp. 181--190.

\bibitem{maddalena2008self2}
L.~{Maddalena} and A.~{Petrosino}, ``A self-organizing approach to background
  subtraction for visual surveillance applications,'' \emph{IEEE Transactions
  on Image Processing}, vol.~17, no.~7, pp. 1168--1177, 2008.

\bibitem{maddalena2012sobs}
L.~Maddalena and A.~Petrosino, ``The sobs algorithm: What are the limits?''
  \emph{2012 IEEE Computer Society Conference on Computer Vision and Pattern
  Recognition Workshops}, pp. 21--26, 2012.

\bibitem{chacon2013improvement}
M.~I. Chacon-Murgu{\'\i}a, G.~Ramirez-Alonso, and S.~Gonzalez-Duarte,
  ``Improvement of a neural-fuzzy motion detection vision model for complex
  scenario conditions,'' in \emph{The 2013 International Joint Conference on
  Neural Networks (IJCNN)}.\hskip 1em plus 0.5em minus 0.4em\relax IEEE, 2013,
  pp. 1--8.

\bibitem{gemignani2015novel}
G.~Gemignani and A.~Rozza, ``A novel background subtraction approach based on
  multi layered self-organizing maps,'' in \emph{2015 IEEE International
  Conference on Image Processing (ICIP)}.\hskip 1em plus 0.5em minus
  0.4em\relax IEEE, 2015, pp. 462--466.

\bibitem{cheng2009realtime}
L.~Cheng and M.~Gong, ``Realtime background subtraction from dynamic scenes,''
  in \emph{2009 IEEE 12th International Conference on Computer Vision}.\hskip
  1em plus 0.5em minus 0.4em\relax IEEE, 2009, pp. 2066--2073.

\bibitem{han2011density}
B.~Han and L.~S. Davis, ``Density-based multifeature background subtraction
  with support vector machine,'' \emph{IEEE Transactions on Pattern Analysis
  and Machine Intelligence}, vol.~34, no.~5, pp. 1017--1023, 2011.

\bibitem{yu2013online}
M.~Yu, Y.~Yu, A.~Rhuma, S.~M.~R. Naqvi, L.~Wang, and J.~A. Chambers, ``An
  online one class support vector machine-based person-specific fall detection
  system for monitoring an elderly individual in a room environment,''
  \emph{IEEE journal of biomedical and health informatics}, vol.~17, no.~6, pp.
  1002--1014, 2013.

\bibitem{farcas2012background}
D.~Farcas, C.~Marghes, and T.~Bouwmans, ``Background subtraction via
  incremental maximum margin criterion: a discriminative subspace approach,''
  \emph{Machine Vision and Applications}, vol.~23, no.~6, pp. 1083--1101, 2012.

\bibitem{marghes2010background}
C.~Marghes and T.~Bouwman, ``Background modeling via incremental maximum margin
  criterion,'' in \emph{Asian Conference on Computer Vision}.\hskip 1em plus
  0.5em minus 0.4em\relax Springer, 2010, pp. 394--403.

\bibitem{marghes2012background}
C.~Marghes, T.~Bouwmans, and R.~Vasiu, ``Background modeling and foreground
  detection via a reconstructive and discriminative subspace learning
  approach,'' in \emph{International Conference on Image Processing, Computer
  Vision, and Pattern Recognition, IPCV 2012}, 2012.

\bibitem{rodriguez2016incremental}
P.~Rodriguez and B.~Wohlberg, ``Incremental principal component pursuit for
  video background modeling,'' \emph{Journal of Mathematical Imaging and
  Vision}, vol.~55, no.~1, pp. 1--18, 2016.

\bibitem{narayanamurthy2018fast}
P.~Narayanamurthy and N.~Vaswani, ``A fast and memory-efficient algorithm for
  robust pca (merop),'' in \emph{2018 IEEE International Conference on
  Acoustics, Speech and Signal Processing (ICASSP)}.\hskip 1em plus 0.5em minus
  0.4em\relax IEEE, 2018, pp. 4684--4688.

\bibitem{cheng2015hybrid}
F.-C. Cheng, B.-H. Chen, and S.-C. Huang, ``A hybrid background subtraction
  method with background and foreground candidates detection,'' \emph{ACM
  Transactions on Intelligent Systems and Technology (TIST)}, vol.~7, no.~1,
  pp. 1--14, 2015.

\bibitem{st2015self}
P.-L. St-Charles, G.-A. Bilodeau, and R.~Bergevin, ``A self-adjusting approach
  to change detection based on background word consensus,'' in \emph{2015 IEEE
  winter conference on applications of computer vision}.\hskip 1em plus 0.5em
  minus 0.4em\relax IEEE, 2015, pp. 990--997.

\bibitem{sedky2014spectral}
M.~Sedky, M.~Moniri, and C.~C. Chibelushi, ``Spectral-360: A physics-based
  technique for change detection,'' in \emph{Proceedings of the IEEE Conference
  on Computer Vision and Pattern Recognition Workshops}, 2014, pp. 399--402.

\bibitem{chiranjeevi2016interval}
P.~Chiranjeevi and S.~Sengupta, ``Interval-valued model level fuzzy
  aggregation-based background subtraction,'' \emph{IEEE transactions on
  cybernetics}, vol.~47, no.~9, pp. 2544--2555, 2016.

\bibitem{faro2011adaptive}
A.~Faro, D.~Giordano, and C.~Spampinato, ``Adaptive background modeling
  integrated with luminosity sensors and occlusion processing for reliable
  vehicle detection,'' \emph{IEEE Transactions on Intelligent Transportation
  Systems}, vol.~12, no.~4, pp. 1398--1412, 2011.

\bibitem{javed2016spatiotemporal}
S.~Javed, A.~Mahmood, T.~Bouwmans, and S.~K. Jung, ``Spatiotemporal low-rank
  modeling for complex scene background initialization,'' \emph{IEEE
  Transactions on Circuits and Systems for Video Technology}, vol.~28, no.~6,
  pp. 1315--1329, 2016.

\bibitem{javed2017background}
S.~{Javed}, A.~{Mahmood}, T.~{Bouwmans}, and S.~K. {Jung},
  ``Background–foreground modeling based on spatiotemporal sparse subspace
  clustering,'' \emph{IEEE Transactions on Image Processing}, vol.~26, no.~12,
  pp. 5840--5854, 2017.

\bibitem{lam2020statistical}
B.~S. Lam, A.~M. Chu, and H.~Yan, ``Statistical bootstrap-based principal mode
  component analysis for dynamic background subtraction,'' \emph{Pattern
  Recognition}, vol. 100, p. 107153, 2020.

\bibitem{miron2015change}
A.~Miron and A.~Badii, ``Change detection based on graph cuts,'' in \emph{2015
  International Conference on Systems, Signals and Image Processing
  (IWSSIP)}.\hskip 1em plus 0.5em minus 0.4em\relax IEEE, 2015, pp. 273--276.

\bibitem{allebosch2015efic}
G.~Allebosch, F.~Deboeverie, P.~Veelaert, and W.~Philips, ``Efic: edge based
  foreground background segmentation and interior classification for dynamic
  camera viewpoints,'' in \emph{International Conference on Advanced Concepts
  for Intelligent Vision Systems}.\hskip 1em plus 0.5em minus 0.4em\relax
  Springer, 2015, pp. 130--141.

\bibitem{patil2020end}
P.~W. Patil, K.~M. Biradar, A.~Dudhane, and S.~Murala, ``An end-to-end edge
  aggregation network for moving object segmentation,'' in \emph{Proceedings of
  the IEEE/CVF Conference on Computer Vision and Pattern Recognition}, 2020,
  pp. 8149--8158.

\bibitem{wei2019enhanced}
J.~Wei, J.~He, Y.~Zhou, K.~Chen, Z.~Tang, and Z.~Xiong, ``Enhanced object
  detection with deep convolutional neural networks for advanced driving
  assistance,'' \emph{IEEE Transactions on Intelligent Transportation Systems},
  vol.~21, no.~4, pp. 1572--1583, 2019.

\bibitem{mandal2019avdnet}
M.~Mandal, M.~Shah, P.~Meena, S.~Devi, and S.~K. Vipparthi, ``Avdnet: A
  small-sized vehicle detection network for aerial visual data,'' \emph{IEEE
  Geoscience and Remote Sensing Letters}, 2019.

\bibitem{redmon2017yolo9000}
J.~Redmon and A.~Farhadi, ``Yolo9000: better, faster, stronger,'' in
  \emph{Proceedings of the IEEE conference on computer vision and pattern
  recognition}, 2017, pp. 7263--7271.

\bibitem{mandal2019sssdet}
M.~Mandal, M.~Shah, P.~Meena, and S.~K. Vipparthi, ``Sssdet: Simple short and
  shallow network for resource efficient vehicle detection in aerial scenes,''
  in \emph{2019 IEEE International Conference on Image Processing
  (ICIP)}.\hskip 1em plus 0.5em minus 0.4em\relax IEEE, 2019, pp. 3098--3102.

\bibitem{feng2020deep}
D.~Feng, C.~Haase-Sch{\"u}tz, L.~Rosenbaum, H.~Hertlein, C.~Glaeser, F.~Timm,
  W.~Wiesbeck, and K.~Dietmayer, ``Deep multi-modal object detection and
  semantic segmentation for autonomous driving: Datasets, methods, and
  challenges,'' \emph{IEEE Transactions on Intelligent Transportation Systems},
  2020.

\bibitem{chen2017deeplab}
L.-C. Chen, G.~Papandreou, I.~Kokkinos, K.~Murphy, and A.~L. Yuille, ``Deeplab:
  Semantic image segmentation with deep convolutional nets, atrous convolution,
  and fully connected crfs,'' \emph{IEEE transactions on pattern analysis and
  machine intelligence}, vol.~40, no.~4, pp. 834--848, 2017.

\bibitem{ji2019context}
Y.~Ji, Y.~Zhan, Y.~Yang, X.~Xu, F.~Shen, and H.~T. Shen, ``A context knowledge
  map guided coarse-to-fine action recognition,'' \emph{IEEE Transactions on
  Image Processing}, vol.~29, pp. 2742--2752, 2019.

\bibitem{li2020learning}
Z.~Li, K.~Nai, G.~Li, and S.~Jiang, ``Learning a dynamic feature fusion tracker
  for object tracking,'' \emph{IEEE Transactions on Intelligent Transportation
  Systems}, 2020.

\bibitem{zhang2019x}
J.~Zhang, Y.~Li, F.~Chen, Z.~Pan, X.~Zhou, Y.~Li, and S.~Jiao, ``X-net: A
  binocular summation network for foreground segmentation,'' \emph{IEEE
  Access}, vol.~7, pp. 71\,412--71\,422, 2019.

\bibitem{yang2019end}
Y.~Yang, T.~Zhang, J.~Hu, D.~Xu, and G.~Xie, ``End-to-end background
  subtraction via a multi-scale spatio-temporal model,'' \emph{IEEE Access},
  vol.~7, pp. 97\,949--97\,958, 2019.

\bibitem{ou2019moving}
X.~Ou, P.~Yan, Y.~Zhang, B.~Tu, G.~Zhang, J.~Wu, and W.~Li, ``Moving object
  detection method via resnet-18 with encoder--decoder structure in complex
  scenes,'' \emph{IEEE Access}, vol.~7, pp. 108\,152--108\,160, 2019.

\bibitem{zeng2019combining}
D.~Zeng, M.~Zhu, and A.~Kuijper, ``Combining background subtraction algorithms
  with convolutional neural network,'' \emph{Journal of Electronic Imaging},
  vol.~28, no.~1, p. 013011, 2019.

\bibitem{shahbaz2019deep}
A.~Shahbaz and K.-H. Jo, ``Deep foreground segmentation using convolutional
  neural network,'' in \emph{2019 IEEE 28th International Symposium on
  Industrial Electronics (ISIE)}.\hskip 1em plus 0.5em minus 0.4em\relax IEEE,
  2019, pp. 1397--1400.

\bibitem{tao2019universal}
Y.~Tao, Z.~Ling, and I.~Patras, ``Universal foreground segmentation based on
  deep feature fusion network for multi-scene videos,'' \emph{IEEE Access},
  vol.~7, pp. 158\,326--158\,337, 2019.

\bibitem{mandal2020motionrec}
M.~Mandal, L.~K. Kumar, M.~S. Saran \emph{et~al.}, ``Motionrec: A unified deep
  framework for moving object recognition,'' in \emph{The IEEE Winter
  Conference on Applications of Computer Vision}, 2020, pp. 2734--2743.

\bibitem{zhao2017joint}
X.~Zhao, Y.~Chen, M.~Tang, and J.~Wang, ``Joint background reconstruction and
  foreground segmentation via a two-stage convolutional neural network,'' in
  \emph{2017 IEEE International Conference on Multimedia and Expo
  (ICME)}.\hskip 1em plus 0.5em minus 0.4em\relax IEEE, 2017, pp. 343--348.

\bibitem{liang2018deep}
X.~Liang, S.~Liao, X.~Wang, W.~Liu, Y.~Chen, and S.~Z. Li, ``Deep background
  subtraction with guided learning,'' in \emph{2018 IEEE International
  Conference on Multimedia and Expo (ICME)}.\hskip 1em plus 0.5em minus
  0.4em\relax IEEE, 2018, pp. 1--6.

\bibitem{zhao2018background}
C.~Zhao, T.-L. Cham, X.~Ren, J.~Cai, and H.~Zhu, ``Background subtraction based
  on deep pixel distribution learning,'' in \emph{2018 IEEE International
  Conference on Multimedia and Expo (ICME)}.\hskip 1em plus 0.5em minus
  0.4em\relax IEEE, 2018, pp. 1--6.

\bibitem{patil2018msednet}
P.~Patil, S.~Murala, A.~Dhall, and S.~Chaudhary, ``Msednet: multi-scale deep
  saliency learning for moving object detection,'' in \emph{2018 IEEE
  International Conference on Systems, Man, and Cybernetics (SMC)}.\hskip 1em
  plus 0.5em minus 0.4em\relax IEEE, 2018, pp. 1670--1675.

\bibitem{jung2019cosine}
J.-I. Jung, J.~Jang, and J.~Hong, ``Cosine focal loss-based change detection
  for video surveillance systems,'' in \emph{2019 16th IEEE International
  Conference on Advanced Video and Signal Based Surveillance (AVSS)}.\hskip 1em
  plus 0.5em minus 0.4em\relax IEEE, 2019, pp. 1--6.

\bibitem{gracewell2019dynamic}
J.~Gracewell and M.~John, ``Dynamic background modeling using deep learning
  autoencoder network,'' \emph{Multimedia Tools and Applications}, pp. 1--21,
  2019.

\bibitem{guo2018learning}
E.~Guo, X.~Fu, J.~Zhu, M.~Deng, Y.~Liu, Q.~Zhu, and H.~Li, ``Learning to
  measure change: Fully convolutional siamese metric networks for scene change
  detection,'' \emph{arXiv preprint arXiv:1810.09111}, 2018.

\bibitem{chen2018mfcnet}
Y.~Chen, X.~Ouyang, and G.~Agam, ``Mfcnet: End-to-end approach for change
  detection in images,'' in \emph{2018 25th IEEE International Conference on
  Image Processing (ICIP)}.\hskip 1em plus 0.5em minus 0.4em\relax IEEE, 2018,
  pp. 4008--4012.

\bibitem{varghese2018changenet}
A.~Varghese, J.~Gubbi, A.~Ramaswamy, and P.~Balamuralidhar, ``Changenet: a deep
  learning architecture for visual change detection,'' in \emph{Proceedings of
  the European Conference on Computer Vision (ECCV)}, 2018, pp. 0--0.

\bibitem{liao2018multiscale}
J.~Liao, G.~Guo, Y.~Yan, and H.~Wang, ``Multiscale cascaded scene-specific
  convolutional neural networks for background subtraction,'' in \emph{Pacific
  Rim Conference on Multimedia}.\hskip 1em plus 0.5em minus 0.4em\relax
  Springer, 2018, pp. 524--533.

\bibitem{akilan2019video}
T.~Akilan, Q.~J. Wu, and W.~Zhang, ``Video foreground extraction using
  multi-view receptive field and encoder--decoder dcnn for traffic and
  surveillance applications,'' \emph{IEEE Transactions on Vehicular
  Technology}, vol.~68, no.~10, pp. 9478--9493, 2019.

\bibitem{han2019aerial}
P.~Han, C.~Ma, Q.~Li, P.~Leng, S.~Bu, and K.~Li, ``Aerial image change
  detection using dual regions of interest networks,'' \emph{Neurocomputing},
  vol. 349, pp. 190--201, 2019.

\bibitem{liang2019spatio}
D.~Liang, J.~Pan, H.~Sun, and H.~Zhou, ``Spatio-temporal attention model for
  foreground detection in cross-scene surveillance videos,'' \emph{Sensors},
  vol.~19, no.~23, p. 5142, 2019.

\bibitem{zheng2019novel}
W.~Zheng, K.~Wang, and F.-Y. Wang, ``A novel background subtraction algorithm
  based on parallel vision and bayesian gans,'' \emph{Neurocomputing}, 2019.

\bibitem{akilan2018new}
T.~Akilan, Q.~J. Wu, W.~Jiang, A.~Safaei, and J.~Huo, ``New trend in video
  foreground detection using deep learning,'' in \emph{2018 IEEE 61st
  International Midwest Symposium on Circuits and Systems (MWSCAS)}.\hskip 1em
  plus 0.5em minus 0.4em\relax IEEE, 2018, pp. 889--892.

\bibitem{sakkos2018end}
D.~Sakkos, H.~Liu, J.~Han, and L.~Shao, ``End-to-end video background
  subtraction with 3d convolutional neural networks,'' \emph{Multimedia Tools
  and Applications}, vol.~77, no.~17, pp. 23\,023--23\,041, 2018.

\bibitem{hu20183d}
Z.~Hu, T.~Turki, N.~Phan, and J.~T. Wang, ``A 3d atrous convolutional long
  short-term memory network for background subtraction,'' \emph{IEEE Access},
  vol.~6, pp. 43\,450--43\,459, 2018.

\bibitem{wang2018foreground1}
Y.~Wang, Z.~Yu, and L.~Zhu, ``Foreground detection with deeply learned
  multi-scale spatial-temporal features,'' \emph{Sensors}, vol.~18, no.~12, p.
  4269, 2018.

\bibitem{wang2018foreground2}
Y.~Wang, L.~Zhu, and Z.~Yu, ``Foreground detection for infrared videos with
  multiscale 3-d fully convolutional network,'' \emph{IEEE Geoscience and
  Remote Sensing Letters}, vol.~16, no.~5, pp. 712--716, 2018.

\bibitem{qiu2019fully}
M.~Qiu and X.~Li, ``A fully convolutional encoder--decoder spatial--temporal
  network for real-time background subtraction,'' \emph{IEEE Access}, vol.~7,
  pp. 85\,949--85\,958, 2019.

\bibitem{patil2019motion}
P.~W. Patil, O.~Thawakar, A.~Dudhane, and S.~Murala, ``Motion saliency based
  generative adversarial network for underwater moving object segmentation,''
  in \emph{2019 IEEE International Conference on Image Processing
  (ICIP)}.\hskip 1em plus 0.5em minus 0.4em\relax IEEE, 2019, pp. 1565--1569.

\bibitem{sultana2019unsupervised}
M.~Sultana, A.~Mahmood, S.~Javed, and S.~K. Jung, ``Unsupervised deep context
  prediction for background estimation and foreground segmentation,''
  \emph{Machine Vision and Applications}, vol.~30, no.~3, pp. 375--395, 2019.

\bibitem{vijayan2020universal}
M.~Vijayan and R.~Mohan, ``A universal foreground segmentation technique using
  deep-neural network,'' \emph{MULTIMEDIA TOOLS AND APPLICATIONS}, 2020.

\bibitem{cai2020background}
X.~Cai and G.~Han, ``Background subtraction based on modified pulse coupled
  neural network in compressive domain,'' \emph{IEEE Access}, vol.~8, pp.
  114\,422--114\,432, 2020.

\bibitem{zhao2018icnet}
H.~Zhao, X.~Qi, X.~Shen, J.~Shi, and J.~Jia, ``Icnet for real-time semantic
  segmentation on high-resolution images,'' in \emph{Proceedings of the
  European Conference on Computer Vision (ECCV)}, 2018, pp. 405--420.

\bibitem{zhang2018fully}
Y.~Zhang, Z.~Qiu, T.~Yao, D.~Liu, and T.~Mei, ``Fully convolutional adaptation
  networks for semantic segmentation,'' in \emph{Proceedings of the IEEE
  Conference on Computer Vision and Pattern Recognition}, 2018, pp. 6810--6818.

\bibitem{yang2018denseaspp}
M.~Yang, K.~Yu, C.~Zhang, Z.~Li, and K.~Yang, ``Denseaspp for semantic
  segmentation in street scenes,'' in \emph{Proceedings of the IEEE Conference
  on Computer Vision and Pattern Recognition}, 2018, pp. 3684--3692.

\bibitem{ronneberger2015u}
O.~Ronneberger, P.~Fischer, and T.~Brox, ``U-net: Convolutional networks for
  biomedical image segmentation,'' in \emph{International Conference on Medical
  image computing and computer-assisted intervention}.\hskip 1em plus 0.5em
  minus 0.4em\relax Springer, 2015, pp. 234--241.

\bibitem{isola2017image}
P.~Isola, J.-Y. Zhu, T.~Zhou, and A.~A. Efros, ``Image-to-image translation
  with conditional adversarial networks,'' in \emph{Proceedings of the IEEE
  conference on computer vision and pattern recognition}, 2017, pp. 1125--1134.

\bibitem{zhu2017unpaired}
J.-Y. Zhu, T.~Park, P.~Isola, and A.~A. Efros, ``Unpaired image-to-image
  translation using cycle-consistent adversarial networks,'' in
  \emph{Proceedings of the IEEE international conference on computer vision},
  2017, pp. 2223--2232.

\bibitem{yu2020background}
W.~Yu, J.~Bai, and L.~Jiao, ``Background subtraction based on gan and domain
  adaptation for vhr optical remote sensing videos,'' \emph{IEEE Access}, 2020.

\bibitem{ammar2020deep}
S.~Ammar, T.~Bouwmans, N.~Zaghden, and M.~Neji, ``Deep detector classifier
  (deepdc) for moving objects segmentation and classification in video
  surveillance,'' \emph{IET Image Processing}, 2020.

\bibitem{garcia2019background}
J.~Garc{\'\i}a-Gonz{\'a}lez, J.~M. Ortiz-de Lazcano-Lobato, R.~M. Luque-Baena,
  and E.~L{\'o}pez-Rubio, ``Background modeling by shifted tilings of stacked
  denoising autoencoders,'' in \emph{International Work-Conference on the
  Interplay Between Natural and Artificial Computation}.\hskip 1em plus 0.5em
  minus 0.4em\relax Springer, 2019, pp. 307--316.

\bibitem{garcia2019foreground}
J.~Garc{\'\i}a-Gonz{\'a}lez, J.~M. Ortiz-de Lazcano-Lobato, R.~M. Luque-Baena,
  M.~A. Molina-Cabello, and E.~L{\'o}pez-Rubio, ``Foreground detection by
  probabilistic modeling of the features discovered by stacked denoising
  autoencoders in noisy video sequences,'' \emph{Pattern Recognition Letters},
  vol. 125, pp. 481--487, 2019.

\bibitem{xie2020autoencoder}
W.~Xie, B.~Liu, Y.~Li, J.~Lei, and Q.~Du, ``Autoencoder and
  adversarial-learning-based semisupervised background estimation for
  hyperspectral anomaly detection,'' \emph{IEEE Transactions on Geoscience and
  Remote Sensing}, 2020.

\bibitem{jo2019regularized}
H.~Jo and J.~Kim, ``Regularized auto-encoder-based separation of defects from
  backgrounds for inspecting display devices,'' \emph{Electronics}, vol.~8,
  no.~5, p. 533, 2019.

\bibitem{xia2018stereoscopic}
C.~Xia, F.~Qi, G.~Shi, and C.~Lin, ``Stereoscopic saliency estimation with
  background priors based deep reconstruction,'' \emph{Neurocomputing}, vol.
  321, pp. 126--138, 2018.

\bibitem{kavasidis2014innovative}
I.~Kavasidis, S.~Palazzo, R.~Di~Salvo, D.~Giordano, and C.~Spampinato, ``An
  innovative web-based collaborative platform for video annotation,''
  \emph{Multimedia Tools and Applications}, vol.~70, no.~1, pp. 413--432, 2014.

\bibitem{li2016weighted}
C.~Li, X.~Wang, L.~Zhang, J.~Tang, H.~Wu, and L.~Lin, ``Weighted low-rank
  decomposition for robust grayscale-thermal foreground detection,'' \emph{IEEE
  Transactions on Circuits and Systems for Video Technology}, vol.~27, no.~4,
  pp. 725--738, 2016.

\bibitem{mahadevan2009spatiotemporal}
V.~Mahadevan and N.~Vasconcelos, ``Spatiotemporal saliency in dynamic scenes,''
  \emph{IEEE transactions on pattern analysis and machine intelligence},
  vol.~32, no.~1, pp. 171--177, 2009.

\bibitem{milan2016mot16}
A.~Milan, L.~Leal-Taix{\'e}, I.~Reid, S.~Roth, and K.~Schindler, ``Mot16: A
  benchmark for multi-object tracking,'' \emph{arXiv preprint
  arXiv:1603.00831}, 2016.

\bibitem{sultani2018real}
W.~Sultani, C.~Chen, and M.~Shah, ``Real-world anomaly detection in
  surveillance videos,'' in \emph{Proceedings of the IEEE Conference on
  Computer Vision and Pattern Recognition}, 2018, pp. 6479--6488.

\bibitem{FliICCV2013}
F.~Li, T.~Kim, A.~Humayun, D.~Tsai, and J.~M. Rehg, ``Video segmentation by
  tracking many figure-ground segments,'' in \emph{ICCV}, 2013.

\bibitem{Perazzi2016}
F.~Perazzi, J.~Pont-Tuset, B.~McWilliams, L.~{Van Gool}, M.~Gross, and
  A.~Sorkine-Hornung, ``A benchmark dataset and evaluation methodology for
  video object segmentation,'' in \emph{Computer Vision and Pattern
  Recognition}, 2016.

\bibitem{ochs2013segmentation}
P.~Ochs, J.~Malik, and T.~Brox, ``Segmentation of moving objects by long term
  video analysis,'' \emph{IEEE transactions on pattern analysis and machine
  intelligence}, vol.~36, no.~6, pp. 1187--1200, 2013.

\bibitem{underwaterchangedetection}
\BIBentryALTinterwordspacing
\emph{underwater change detection dataset}, 2019 (accessed June 25, 2020).
  [Online]. Available: \url{http://underwaterchangedetection.eu/index.html}
\BIBentrySTDinterwordspacing

\bibitem{singha2019tu}
A.~Singha and M.~K. Bhowmik, ``Tu-vdn: Tripura university video dataset at
  night time in degraded atmospheric outdoor conditions for moving object
  detection,'' in \emph{2019 IEEE International Conference on Image Processing
  (ICIP)}.\hskip 1em plus 0.5em minus 0.4em\relax IEEE, 2019, pp. 2936--2940.

\bibitem{singha2019salient}
A.~{Singha} and M.~K. {Bhowmik}, ``Salient features for moving object detection
  in adverse weather conditions during night time,'' \emph{IEEE Transactions on
  Circuits and Systems for Video Technology}, pp. 1--1, 2019.

\bibitem{m2020moruav}
M.~Mandal, L.~K. Kumar, and S.~K. Vipparthi, ``Mor-uav: A benchmark dataset and
  baselines for moving object recognition in uav videos,'' in \emph{Proceedings
  of the 28th ACM International Conference on Multimedia}, 2020, pp.
  2626--2635.

\bibitem{zhu2018vision}
P.~Zhu, L.~Wen, X.~Bian, H.~Ling, and Q.~Hu, ``Vision meets drones: A
  challenge,'' \emph{arXiv preprint arXiv:1804.07437}, 2018.

\bibitem{zhu2019visdrone}
P.~Zhu, D.~Du, L.~Wen, X.~Bian, H.~Ling, Q.~Hu, T.~Peng, J.~Zheng, X.~Wang,
  Y.~Zhang \emph{et~al.}, ``Visdrone-vid2019: The vision meets drone object
  detection in video challenge results,'' in \emph{Proceedings of the IEEE
  International Conference on Computer Vision Workshops}, 2019, pp. 0--0.

\bibitem{mueller2016benchmark}
M.~Mueller, N.~Smith, and B.~Ghanem, ``A benchmark and simulator for uav
  tracking,'' in \emph{European conference on computer vision}.\hskip 1em plus
  0.5em minus 0.4em\relax Springer, 2016, pp. 445--461.

\bibitem{lin2017focal}
T.-Y. Lin, P.~Goyal, R.~Girshick, K.~He, and P.~Doll{\'a}r, ``Focal loss for
  dense object detection,'' in \emph{Proceedings of the IEEE international
  conference on computer vision}, 2017, pp. 2980--2988.

\bibitem{linfocalpami}
T.~{Lin}, P.~{Goyal}, R.~{Girshick}, K.~{He}, and P.~{Dollár}, ``Focal loss
  for dense object detection,'' \emph{IEEE Transactions on Pattern Analysis and
  Machine Intelligence}, vol.~42, no.~2, pp. 318--327, Feb 2020.

\bibitem{chen2020high}
X.~Chen, Z.~Li, Y.~Yang, L.~Qi, and R.~Ke, ``High-resolution vehicle trajectory
  extraction and denoising from aerial videos,'' \emph{IEEE Transactions on
  Intelligent Transportation Systems}, 2020.

\bibitem{jiang2021moving}
C.~Jiang, D.~P. Paudel, D.~Fofi, Y.~Fougerolle, and C.~Demonceaux, ``Moving
  object detection by 3d flow field analysis,'' \emph{IEEE Transactions on
  Intelligent Transportation Systems}, 2021.

\bibitem{cai2021pedestrian}
Y.~Cai, L.~Dai, H.~Wang, L.~Chen, Y.~Li, M.~A. Sotelo, and Z.~Li, ``Pedestrian
  motion trajectory prediction in intelligent driving from far shot
  first-person perspective video,'' \emph{IEEE Transactions on Intelligent
  Transportation Systems}, 2021.

\bibitem{song2020pedestrian}
X.~Song, K.~Chen, X.~Li, J.~Sun, B.~Hou, Y.~Cui, B.~Zhang, G.~Xiong, and
  Z.~Wang, ``Pedestrian trajectory prediction based on deep convolutional lstm
  network,'' \emph{IEEE Transactions on Intelligent Transportation Systems},
  2020.

\bibitem{fang2020move}
Z.~Fang, A.~Jain, G.~Sarch, A.~W. Harley, and K.~Fragkiadaki, ``Move to see
  better: Towards self-supervised amodal object detection,'' \emph{arXiv
  preprint arXiv:2012.00057}, 2020.

\bibitem{xu2019self}
D.~Xu, J.~Xiao, Z.~Zhao, J.~Shao, D.~Xie, and Y.~Zhuang, ``Self-supervised
  spatiotemporal learning via video clip order prediction,'' in
  \emph{Proceedings of the IEEE Conference on Computer Vision and Pattern
  Recognition}, 2019, pp. 10\,334--10\,343.

\bibitem{han2019video}
T.~Han, W.~Xie, and A.~Zisserman, ``Video representation learning by dense
  predictive coding,'' in \emph{Proceedings of the IEEE International
  Conference on Computer Vision Workshops}, 2019, pp. 0--0.

\end{thebibliography}

\end{document}